\documentclass[sigconf]{acmart}
\AtBeginDocument{%
  }

\setcopyright{cc}
\setcctype{by-nc-nd}
\copyrightyear{2026}
\acmYear{2026}
\acmDOI{10.1145/3772318.3790920}
\acmConference[CHI '26]{Proceedings of the 2026 CHI Conference on Human Factors in Computing Systems}{April 13--17, 2026}{Barcelona, Spain}
\acmBooktitle{Proceedings of the 2026 CHI Conference on Human Factors in Computing Systems (CHI '26), April 13--17, 2026, Barcelona, Spain}
\acmISBN{979-8-4007-2278-3/2026/04}

\begin{document}

\title{How Do We Research Human-Robot Interaction in the Age of Large Language Models? A Systematic Review}


\author{Yufeng Wang}
\authornote{Both authors contributed equally to this research.}
\orcid{0009-0002-4824-6168}
\affiliation{%
  \institution{The Hong Kong University of Science and Technology (Guangzhou)}
  \city{Guangzhou}
  \country{China}
}
\affiliation{%
  \institution{Zhejiang University}
  \city{Hangzhou}
  \country{China}
}
\email{wyufeng@zju.edu.cn}

\author{Yuan Xu}
\authornotemark[1]
\orcid{0009-0004-0811-9505}
\affiliation{%
  \institution{The Hong Kong University of Science and Technology (Guangzhou)}
  \city{Guangzhou}
  \country{China}
}
\email{yxu712@connect.hkust-gz.edu.cn}

\author{Anastasia Nikolova}
\orcid{0009-0001-7476-1596}
\affiliation{%
  \institution{The Hong Kong University of Science and Technology (Guangzhou)}
  \city{Guangzhou}
  \country{China}
}
\email{anikolova721@connect.hkust-gz.edu.cn}

\author{Yuxuan Wang}
\orcid{0009-0001-2638-1847} 
\affiliation{%
  \institution{The Hong Kong University of Science and Technology (Guangzhou)}
  \city{Guangzhou}
  \country{China}
}
\affiliation{%
  \institution{Savannah College of Art and Design}
  \city{Savannah}
  \country{USA}
}
\email{yuwang85@student.scad.edu}

\author{Jianyu Wang}
\orcid{0009-0000-3731-6250}
\affiliation{%
  \institution{The Hong Kong University of Science and Technology (Guangzhou)}
  \city{Guangzhou}
  \country{China}
}
\affiliation{%
  \institution{Zhejiang University}
  \city{Hangzhou}
  \country{China}
}
\email{3220100890@zju.edu.cn}

\author{Chongyang Wang}
\authornote{Corresponding authors}
\orcid{0000-0002-9819-088X} 
\affiliation{%
  \institution{West China Hospital, Sichuan University}
  \city{Chengdu}
  \country{China}
}
\email{wangchongyang@scu.edu.cn}

\author{Xin Tong}
\authornotemark[2]
\orcid{0000-0002-8037-6301} 
\affiliation{%
  \institution{The Hong Kong University of Science and Technology (Guangzhou)}
  \city{Guangzhou}
  \country{China}
}
\affiliation{%
  \institution{The Hong Kong University of Science and Technology}
  \city{Hong Kong}
  \country{China}
}
\email{xint@hkust-gz.edu.cn}

\renewcommand{\shortauthors}{Wang and Xu et al.}

\begin{abstract}
Advances in large language models (LLMs) are profoundly reshaping the field of human–robot interaction (HRI). While prior work has highlighted the technical potential of LLMs, few studies have systematically examined their human-centered impact (e.g., human-oriented understanding, user modeling, and levels of autonomy), making it difficult to consolidate emerging challenges in LLM-driven HRI systems. Therefore, we conducted a systematic literature search following the PRISMA guideline, identifying 86 articles that met our inclusion criteria. Our findings reveal that: (1) LLMs are transforming the fundamentals of HRI by reshaping how robots sense context, generate socially grounded interactions, and maintain continuous alignment with human needs in embodied settings; and (2) current research is largely exploratory, with different studies focusing on different facets of LLM-driven HRI, resulting in wide-ranging choices of experimental setups, study methods, and evaluation metrics. Finally, we identify key design considerations and challenges, offering a coherent overview and guidelines for future research at the intersection of LLMs and HRI.
\end{abstract}



\begin{teaserfigure}
\centering
  \includegraphics[width= 1\linewidth]{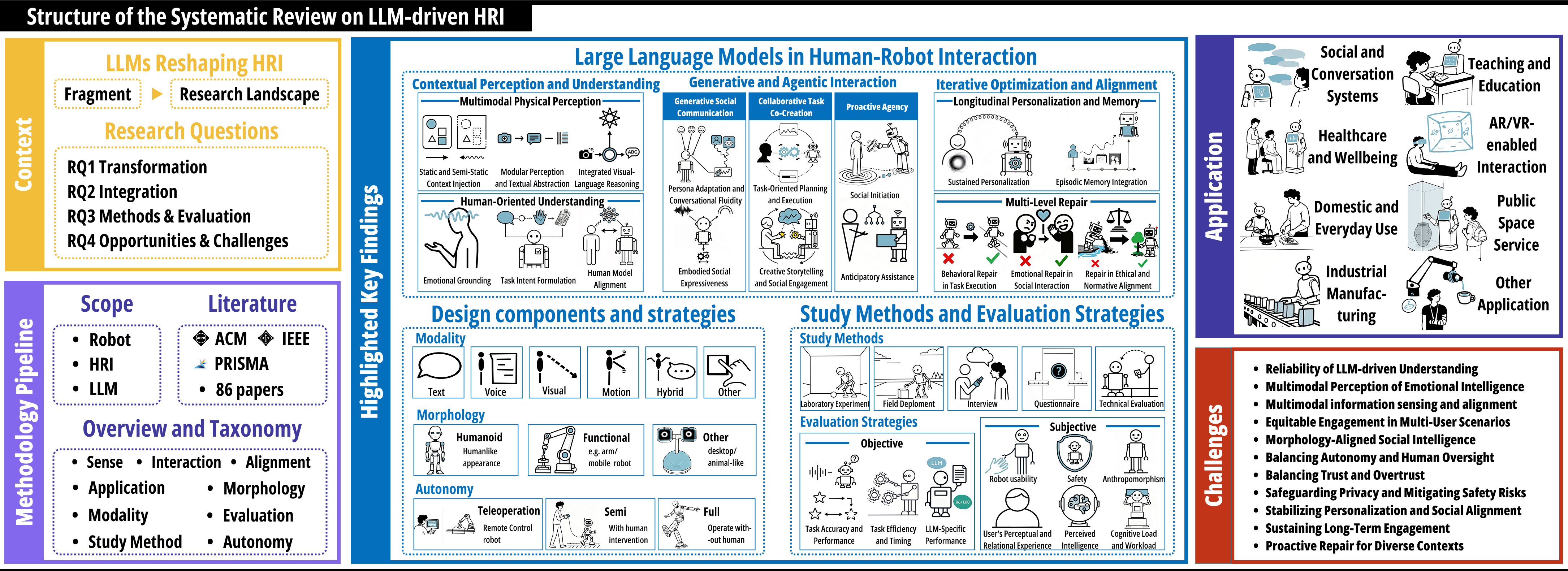}
  \caption{Visual abstract of our systematic review of LLM-driven HRI, summarizing the overall structure of our work.} 
  \Description{This teaser diagram illustrates the holistic structure of the systematic review on Large Language Model (LLM)-driven Human-Robot Interaction (HRI), organized into interconnected thematic sections: (1) Context (top-left, yellow section): Outlines how LLMs reshape the fragmented HRI research landscape, alongside four core research questions (RQ1: Transformation, RQ2: Integration, RQ3: Methods & Evaluation, RQ4: Opportunities & Challenges). (2) Methodology Pipeline (bottom-left, purple section): Details the review’s scope (focused on Robot, HRI, LLM), literature retrieval (sources: ACM, IEEE; following PRISMA guidelines; 86 papers included), and the review’s taxonomy (key dimensions: Sense-Interaction-Alignment, Application, Modality, Morphology, Evaluation, Study Method, Autonomy). (3) Highlighted Key Findings (central blue section): The core section, centered on the Sense-Interaction-Alignment framework for LLM-driven HRI: Contextual Perception and Understanding (multimodal physical perception, human-oriented understanding like emotional grounding); Generative and Agentic Interaction (generative social communication, collaborative task co-creation, proactive agency); Iterative Optimization and Alignment (longitudinal personalization/memory, multi-level repair); It also integrates Design Components and Strategies (Modality, Morphology, Autonomy) and Study Methods and Evaluation Strategies (research methods e.g., laboratory experiments; evaluation metrics split into Objective/Subjective). (4) Application (top-right, purple section): Enumerates 8 application domains (e.g., Social and Conversation Systems, Healthcare and Wellbeing, Industrial Manufacturing). (5)Challenges (bottom-right, red section): Summarizes core 11 research challenges (e.g., LLM-driven understanding reliability, multi-modal emotional perception, trust calibration, long-term engagement).} 
  \label{fig:teaser}
\end{teaserfigure}


\maketitle

\section{Introduction}

Human--robot interaction (HRI) is a multidisciplinary field devoted to creating efficient, safe, and comfortable ways for people to collaborate with robots~\cite{ahmadExploringHumanRobotInteraction2024, bartneckHumanrobotInteractionIntroduction2020}. Its enduring goal is to facilitate natural and effective interactions between humans and robots~\cite{goodrichHumanRobotInteractionSurvey2007a, rodriguez-guerraHumanRobotInteractionReview2021}. However, the field has persistently faced significant challenges in achieving this, particularly in enabling robots to adapt to unexpected situations or unpredictable human behaviors in real-world, synchronous environments~\cite{delucaIntegratedControlPHRI2012, schreiterEvaluatingEfficiencyEngagement2025}.

The rapid advancement of large language models (LLMs) has introduced a transformative potential to address these challenges. To elaborate, LLMs enable robots to acquire, reason, and apply knowledge in physically grounded and socially environments~\cite{paolo2024embodiedai} by enhancing their in-context learning~\cite{BrownLanguagemodels2020}, commonsense reasoning~\cite{talmor2022commonsenseqa}, and chain-of-thought capabilities~\cite{WeiChainofthought2022}. Consequently, LLMs are reshaping the HRI landscape, powering innovations in emotional responses~\cite{mishra2023real}, adaptive task planning~\cite{zhaxizhuoma2024alignbot}, context-aware tutoring~\cite{sievers2025humanoidsocialrobotteaching, sun2025integratingemotionalintelligencememory}, and personalized care~\cite{app14219922, LimEnhancinghuman2024, GKOURNELOS20249}. This integration constitutes a promising research frontier to advance embodied intelligence and enable more seamless collaboration~\cite{linEmbodiedAILarge2024b}.

\begin{figure*}[t]
    \centering
    \includegraphics[width= 1\textwidth]{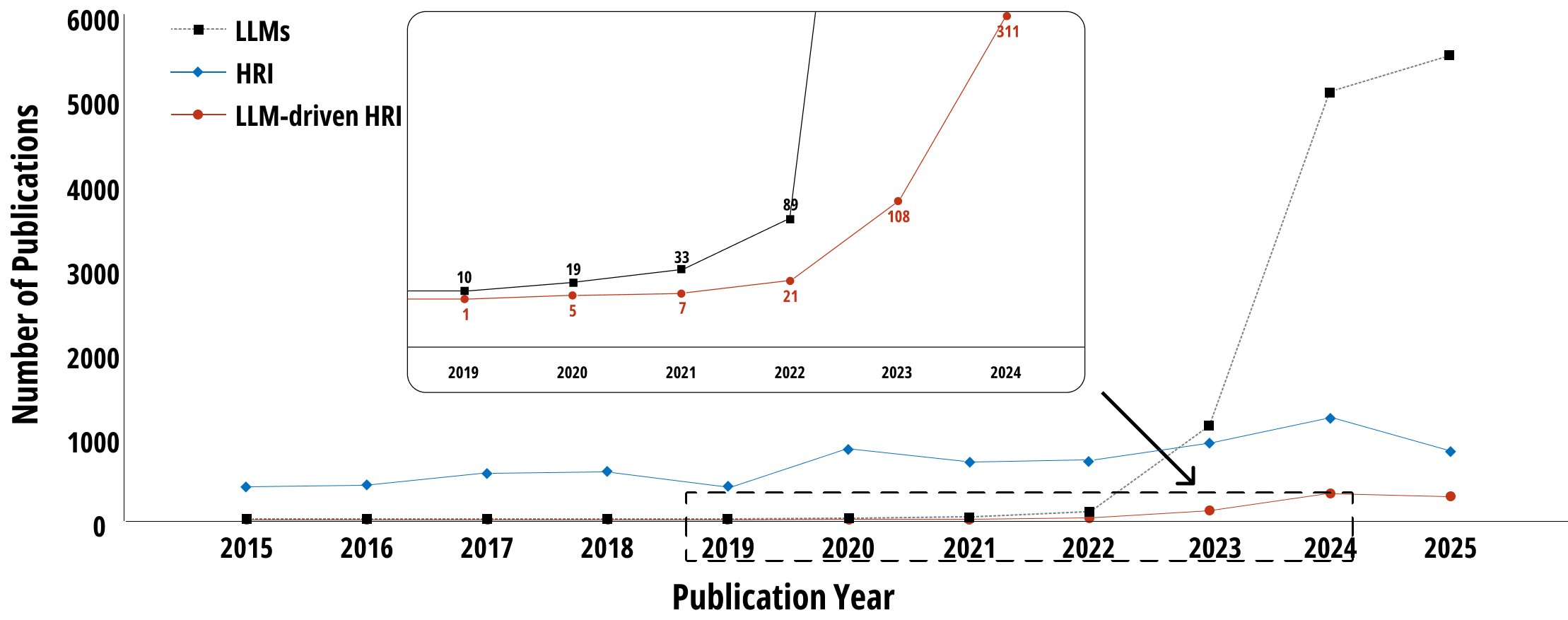}
    \caption{Publication trends of three research domains (LLMs, HRI, LLM-driven HRI) in the ACM digital library from 2015 to 2025 (details of search keywords are provided in Appendix~\ref{section:appendix_A}).}
    \Description{Publication Trend Chart: Temporal Distribution of Publications in LLMs, HRI, and LLM-driven HRI (2015–2025). The chart presents three time-series curves: (1) LLMs (dashed black line): Shows minimal publication volume (near 0) before 2019, followed by gradual growth (10 in 2019, 33 in 2021, 89 in 2022) and a sharp surge post-2023 (exceeding 1000 by 2024, with a continued upward trend into 2025). (2) HRI (solid blue line): Maintains a stable, moderate volume (consistently ~500–1000 publications) across 2015–2025, with minor fluctuations. (3) LLM-driven HRI (solid red line): Emerges in 2019 (1 publication), experiences slow initial growth (5 in 2020, 7 in 2021, 21 in 2022), then accelerates rapidly (108 in 2023, 311 in 2024) — aligning with the expansion of LLMs research, while remaining distinct from the established HRI literature base. An inset box (2019–2024) magnifies the early growth phase of LLM-driven HRI and the initial expansion of LLMs, highlighting the temporal coupling between the two fields. The chart illustrates the recent, explosive growth of LLM-related research (including LLM-driven HRI) against the stable background of traditional HRI scholarship.}
    \label{fig:trend}
\end{figure*}

Given the transformative impact of LLMs, we posit that it is a critical juncture to systematically review and synthesize this rapidly evolving landscape. The urgency of this synthesis is underscored by our preliminary analysis of publication trends in the ACM digital library (DL) (2015–2025), which reveals that the integration of LLMs into HRI has become a burgeoning and vital research domain since 2021 (see Figure~\ref{fig:trend}). Furthermore, interdisciplinary human–computer interaction (HCI) methodologies facilitate the design and evaluation of technologies from a human-centered perspective, adhering to guidelines like ISO 9241-210:2019~\cite{firminodesouzaTrustTrustworthinessHumanCentered2025, ISO9241}. However, our survey of the current literature highlights a critical gap: while existing reviews predominantly focus on technical implementation, such as model robustness and architectural advancements~\cite{zhangLargeLanguageModels2023}, they often overlook the human-centered considerations that are foundational to the HRI field, including human-oriented understanding~\cite{fongSurveySociallyInteractive2003a}, user modeling~\cite{schaalNewRoboticsHumancentered2007}, and levels of autonomy~\cite{goodrichHumanRobotInteractionSurvey2007a}. Moreover, discussions in current HRI studies remain fragmented, lacking a cohesive synthesis that aligns these technological leaps with established human-centered perspectives~\cite{pramanick2024transformingscholarlylandscapesinfluence}.

To address these gaps and provide a structured overview of this vital field, we conducted a systematic review following the PRISMA guidelines~\cite{page2021prisma}. Our review specifically focuses on LLM-driven HRI studies from the past five years that involve practical interaction scenarios and examine the evolving role of LLMs in shaping the interaction lifecycle, resulting in 86 papers for in-depth analysis. Our investigation is structured around the following research questions:
(1) \textbf{RQ1:} How do LLMs transform the foundational capabilities of HRI? (Section ~\ref{section:LLMs-HRI} proposes the Sense–Interaction–Alignment framework); 
(2) \textbf{RQ2:} How are LLMs integrated into HRI system design? (Section ~\ref{sec: design} discusses design components and strategies); 
(3) \textbf{RQ3:} How can LLM-driven HRI systems be evaluated? (Section ~\ref{sec: method} explores study methods and evaluation strategies); and 
(4) \textbf{RQ4:} What are the opportunities and challenges for future research? (Section ~\ref{sec:application} and ~\ref{sec:challenges} discuss applications and challenges respectively).

As LLMs inject new vitality into HRI, we systematically synthesize the current landscape of LLM-driven robotic systems to offer researchers a holistic overview connecting foundational capabilities, system design, and evaluation, while highlighting emergent challenges to inform future directions. Our analysis reveals that existing research has increasingly organized around a Sense-Interaction-Alignment paradigm, marking a key shift from rigid, task-specific pipelines toward adaptive, socially aware, and iteratively optimizable embodied intelligence. We further identify substantial heterogeneity in how LLMs are integrated across robot autonomy, interaction modalities, and physical embodiments, as well as a dual-focus evaluation trend that jointly considers objective task performance and subjective human experience. In summary, this paper presents the following contributions:

\begin{itemize}
    \item A synthesis of LLM-driven HRI studies, demonstrating how LLMs enable contextual sensing, generative interaction, and adaptive alignment in embodied settings, and providing insights to support researchers in navigating the evolving landscape of LLM-HRI integration.
    \item  A proposed systematic taxonomy, identifying the core areas emphasized in LLM-era HRI research and offering a structured categorization of studies across nine key dimensions.
    \item Identification of emerging challenges for future research directions, highlighting key design considerations, such as ensuring the reliability of LLM-driven understanding, maintaining appropriate levels of user trust, and achieving robust multimodal grounding in dynamic environments.
    \item An open-access, searchable online database containing the 86 included studies\footnote{\url{https://llms-hri.github.io/}}. The platform allows users to browse the literature, perform retrieval, and interact with visualized charts, thereby supporting transparency and reproducibility.
\end{itemize}

\section{Scope and Related Work}

\subsection{Scope} 
\label{subsec:Scope}
This section defines the key terms used in this paper to establish a common ground for our discussion and clarify the scope of this study.
\subsubsection{Robot} 
\label{subsubsec:Robot}
A robot is conventionally defined as a physical agent that autonomously perceives its environment via sensors and acts through effectors~\cite{sicilianoSpringerHandbookRobotics2008}, generating behaviors to accomplish one or more tasks~\cite{IEEEStandardOntologies2015}. According to ISO 8373:2021~\cite{ISO8373}, this is further specified as a ``programmed actuated mechanism with a degree of autonomy to perform locomotion, manipulation, or positioning.'' Complementing this technical framing, HCI and HRI work often conceives robots as an umbrella term describing a miscellaneous collection of (semi-)automated devices with various capabilities and appearances~\cite{goodrichHumanRobotInteractionSurvey2007a}, ranging from traditional industrial robots to simulated agents and actuated user interfaces~\cite{suzukiAugmentedRealityRobotics2022, lawrenceRoleSocialNorms2025, nigroSocialGroupHumanRobot2025}. Despite this inclusive conceptualization, several studies further emphasize that a robot's physical presence and embodied action are its most critical properties, as they enable the system to act in, sense, and reshape human environments~\cite{weisswangeDesignImplicationsRobots2025, pascherHowCommunicateRobot2023, weiHumanRobotInteraction2025}.
In this paper, we adopt a precise definition of robots as entities possessing a physical embodiment or a simulated physical presence (e.g., in VR or AR environments)~\cite{fongSurveySociallyInteractive2003a}. This includes systems such as robotic arms~\cite{karliAlchemistLLMAidedEndUser2024a, ikedaMARCERMultimodalAugmented2025a, mahadevanImageInThatManipulatingImages2025} and humanoid robots~\cite{pereiraMultimodalUserEnjoyment2024a, grassiStrategiesControllingConversation2025} that interact with the physical environment (pHRI) through sensors (e.g., cameras, LiDAR) and actuators (e.g., motors, joints).
While we acknowledge that purely disembodied agents, such as chatbots, play a pivotal role in natural language processing research~\cite{shrivastavaNaturalLanguageProcessing2025, westerchatbot2024}, the primary focus of this work is to explore the unique implications of embodiment. Therefore, to maintain this research focus and better investigate what we consider the essential characteristics of HRI~\cite{dautenhahnSociallyIntelligentRobots2007, dautenhahnMethodologyThemesHumanRobot2007}, we deliberately exclude non-embodied agents. 

\subsubsection{Human-Robot Interaction} \label{subsubsec:HRI}
\begin{figure}[t]
    \centering
    \includegraphics[width=1\linewidth]{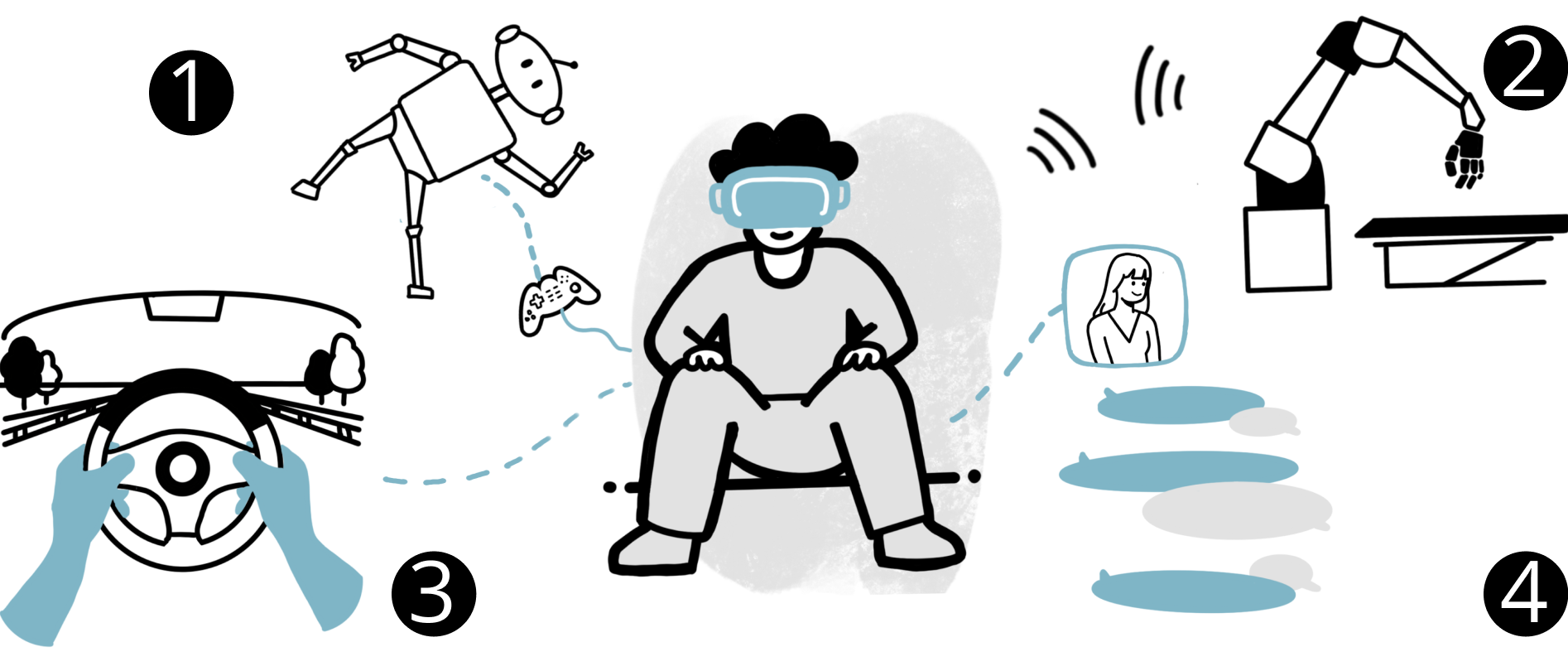}
    \caption{Illustration of the four core HRI types in Sheridan’s classification framework: (1) supervisory control; (2) teleoperation; (3) automated vehicles; (4) social interaction.}
    \Description{Illustration of the four core Human-Robot Interaction (HRI) types in Sheridan’s classification framework (delineating the scope of HRI): (1) Supervisory Control (depicted by a robot paired with a remote controller, representing human oversight of robotic systems); (2) Teleoperation (showcased via a robotic arm operated through VR interfaces, enabling remote manual control); (3) Automated Vehicles (illustrated by a vehicle steering system, corresponding to human-automation interaction in autonomous mobility); (4) Social Interaction (represented by conversational dialogue bubbles alongside human-robot engagement, capturing social-oriented HRI).}
    \label{fig:Cato_HRI}
\end{figure}

As a multidisciplinary research domain, HRI integrates perspectives from HCI, psychology, and sociology~\cite{dautenhahnMethodologyThemesHumanRobot2007, stock-homburgSurveyEmotionsHuman2022, jostHumanRobotInteractionEvaluation2020a}, with a primary emphasis on Human-Centered Design (HCD)~\cite{mortezapour2025humancenteredaifocushumanrobot, electronics12010167, xieEmbodiedGenerativeAI2025, zhangHumancenteredAITechnologies2024}. To delineate the scope of HRI, we adopt the classification framework proposed by Sheridan~\cite{SheridanHRI2016}, which identifies four major interaction types: (1) supervisory control; (2) teleoperation; (3) automated vehicles; and (4) social interaction, as illustrated in Figure \ref{fig:Cato_HRI}. To ensure empirical grounding, HRI research relies on user studies, in which you measure how users respond to variations of the robot, the interaction itself, or the context of the interaction~\cite{bartneckHumanrobotInteractionIntroduction2020, leichtmannCrisisAheadWhy2022}. Typical approaches include Wizard-of-Oz studies~\cite{riek2012wizard, dahlbackWizardOzStudies1993}, structured interviews~\cite{velingQualitativeResearchHRI2021}, questionnaires~\cite{rueben2020introduction}, and field deployments~\cite{bufieldnotes2024}. 

\subsubsection{Large Language Model} 
\label{subsubsec:LLM}
Large language models are natural language models that refer to Transformer-based architectures comprising billions of parameters~\cite{zhao2025surveylargelanguagemodels}, with notable examples including GPT-3~\cite{floridiGPT3ItsNature2020}, Grok 3~\cite{Grok3xAI2025}, and LLaMA~\cite{touvron2023llamaopenefficientfoundation}, as well as recent multi-modal LLMs (MLLMs) such as GPT-4~\cite{openai2024gpt4technicalreport}, GPT-5~\cite{OpenAI2025GPT5}, and Gemini~\cite{geminiroboticsteam2025geminiroboticsbringingai}. Although these architectures were originally designed for Natural Language Processing (NLP) tasks~\cite{kojima2022large}, the field has rapidly expanded into the multimodal domain. This evolution has driven the emergence of vision-language models (VLMs) and  MLLMs~\cite{Yin_2024}, which integrate textual, visual, and sometimes auditory modalities to enable richer forms of perception, reasoning, and interaction. Given this rapid expansion and the need to ensure our review remains sufficiently forward-looking, we adopt a broad definition of LLMs that encompasses pre-trained language models (PLMs)~\cite{han2021pretrainedmodelspastpresent, Qiu_2020, zhou2023comprehensivesurveypretrainedfoundation}, as well as cutting-edge VLMs~\cite{zhang2024vision, shao2025large}, and MLLMs~\cite{li2025surveyingmllmlandscapemetareview, zhu2024multilingual}.

\subsection{Related Work}

To provide a structured overview of related studies and their contributions, we summarize key prior works in Appendix~\ref{section:appendix_B}, categorizing them by type, focus, and main contributions. 

With the rapid development of LLMs, an increasing number of surveys and reviews have emerged in this domain, addressing different aspects of robotics, such as robotic systems~\cite{zengLargeLanguageModels2023, wangLargeLanguageModels2025, firooziFoundationModelsRobotics2023}, robotic intelligence~\cite{jeongSurveyRobotIntelligence2024, kimSurveyIntegrationLarge2024}, robot autonomy~\cite{wangSurveyLargeLanguage2024, liuIntegratingLargeLanguage2025}, multi-agents~\cite{guoLargeLanguageModel2024, liLargeLanguageModels2025, xiRisePotentialLarge2025}, and embodied AI~\cite{linEmbodiedAILarge2024b, salimpourEmbodiedAgenticAI2025}. For example, Zeng et al.~\cite{zengLargeLanguageModels2023} survey mainstream LLMs and analyze interaction as related technologies, with a focus on game-based and language-based HRI. Wang et al.~\cite{wangLargeLanguageModels2025} and Liu et al.\cite{liuIntegratingLargeLanguage2025} situate HRI in the robot task category, offering analysis in areas such as natural language interaction, task planning, and interaction experience. Xi et al.~\cite{xiRisePotentialLarge2025} broadly categorize the interactions of LLM-based agents into two types: cooperative interaction and adversarial interaction. Collectively, these works provide domain-specific analyses that enrich the broader understanding of LLM-driven robotics and offer conceptual perspectives that help situate and contextualize our focus on HRI.
There are some studies initially focusing on LLMs in a HCI perspective. Among these works, Zhang et al.\cite{zhangLargeLanguageModels2023} provide the first review of how LLMs enhance HRI across inquiry answering, commonsense, and instruction following, while highlighting key challenges in safety, context understanding, and scalability. Shi et al.\cite{shiHowCanLarge2024} demonstrate that, within socially assistive robotics, LLMs enable natural dialogue, multimodal user understanding, and policy synthesis. Zou et al.~\cite{zouLLMBasedHumanAgentCollaboration2025} propose a general taxonomy of LLM-driven human–agent systems, revealing that incorporation of environment and profiling, human feedback, interaction types, orchestration, and communication enhances system performance, reliability and safety. Atuhurra's meta-study of 250 HRI papers~\cite{atuhurraLeveragingLargeLanguage2024} corroborates these technological benefits while offering a critical counterpoint: the LLM substitution for traditional speech, intent, and perception modules that enriches knowledge, reasoning, and personalization in social robots simultaneously intensifies algorithmic bias, data-leakage risks, and computational overhead. 

While these prior studies have analyzed and evaluated the role of LLMs, and revealed possible directions in the post-LLM era, their scope remains limited. First, existing works often concentrate on the technical potential and performance of LLMs, such as advancements in core model architectures~\cite{zhangLargeLanguageModels2023}, improvements in training datasets~\cite{wangSurveyLargeLanguage2024, firooziFoundationModelsRobotics2023, liuIntegratingLargeLanguage2025}, and fine-tuning paradigms~\cite{liLargeLanguageModels2025}. Second, most works lack systematic and transparent procedures for literature identification, inclusion, and analysis. Further, LLMs have quickly expanded from text-only models to increasingly capable multimodal systems, and these developments are poised to introduce new opportunities, challenges, and design considerations for HRI that require timely integration and synthesis. To fill the gap, we provide the first systematic review of LLM-driven HRI. We adopt a rigorous and transparent methodology to synthesize how HRI is being reconceptualized in the era of LLMs and to outline potential directions for future LLM-driven HRI research.

\section{Methodology}
This systematic review was conducted and structured following the Preferred Reporting Items for Systematic Reviews and Meta-Analyses (PRISMA) 2020 guidelines \cite{page2021prisma}.

\subsection{Search Strategy}
A complete list of all databases, search results, and inclusion counts is provided in the Appendix~\ref {section:appendix_C}. In this section, we focus on detailing the formulation of our search queries and the rationale behind our database selection.

\subsubsection{Search Query Formulation} 

\label{subsubsec:SearchQuery}

After surveying a substantial body of literature, we established the scope of our review as outlined in \textbf{Section~\ref{subsec:Scope}}. To construct a robust search strategy, we first conducted an exploratory search on Google Scholar using the query: \textit{robot AND (llm OR vlm OR mllm OR gpt) AND hri}. However, this initial attempt revealed two limitations: (1) a large number of non–peer-reviewed arXiv manuscripts, and (2) considerable noise caused by acronym-based searches. For example, LLM is also commonly used to refer to the Master of Laws degree~\cite{tangLLMIntegrationExtended2025}. To mitigate these issues, we turned to ACM DL~\footnote{The search was conducted on July 22, 2025.}, whose technical focus helped reduce such ambiguity.

From this initial query, we identified 157 papers, which our research team divided for close reading. After a week of analysis and group discussion, several insights emerged that helped refine and supplement our scope. In this process, we found that some papers only mentioned LLMs briefly in discussion or future work sections, without integrating or evaluating them as core system components~\cite{rasouliCoDesignUserEvaluation2025, axelssonParticipantPerceptionsRobotic2025, kimUnderstandingExpectationsRobotic2025, boonyardFirefightingDroneAssistance2025a}. Others identified limitations of existing LLMs to emphasize the strengths of their own proposed methods~\cite{williamsImprovisingInteractionApplied2025, martin-ozimekLearningNonverbalCues2025, bouzidaCARMENCognitivelyAssistive2024}. These observations helped establish a stricter inclusion criterion: we focus exclusively on studies that integrate, evaluate, or operationalize LLMs as technical components for addressing HRI problems.

During the close reading, we noticed that many HRI studies also used the term Human–Robot Collaboration (HRC) to place greater emphasis on collaborative scenarios~\cite{hostettlerRealTimeAdaptiveIndustrial2025, goubardCognitiveModellingVisual, ranasingheLargeLanguageModels2025, zhangCanYouPass2025a}. As we know, the robotics community commonly regards physical Human–Robot Collaboration (pHRC) as falling within the broader scope of pHRI~\cite{alamiSafeDependablePhysical2006, ajoudaniProgressProspectsHuman2018, murphyHumanrobotInteractionRescue2004}. Further, recent HRI reviews place particular emphasis on collaborative interaction~\cite{campagnaSystematicReviewTrust2025} and even include HRC-related terms in their search strategies to ensure adequate coverage~\cite{wangSystematicReviewXREnabled2025}. Therefore, to avoid omissions, we accordingly expanded our terminology. A similar refinement was needed for robot-related keywords. Our decision was guided by two considerations. First, because our review adopts a user-centered perspective, social robots and humanoid robots play a particularly important role. Social robots represent highly interaction-oriented platforms~\cite{pinto-bernalDesigningSocialRobots2025}, and humanoid robots tend to afford more natural and embodied user-centered interaction~\cite{strathearnModellingUserPreference2020}. Second, our preliminary screening revealed that a large proportion of retrieved studies already centered on social and humanoid robots. Although the general term "robot''could technically encompass these categories, we chose—after consulting domain experts—to include "social robot''and "humanoid robot''in our search terms to better reflect our human-centered orientation and ensure more comprehensive coverage of interaction-focused HRI work.

These iterative refinements led to our final search query, which incorporates full-form terms, representative model names, and key concept variants: \textit{(``large language model''OR LLM OR ChatGPT OR GPT-3 OR GPT-4) AND (robot OR robotics OR ``social robot''OR ``humanoid robot") 
AND (``human-robot interaction''OR HRI OR ``human robot collaboration''OR HRC)}. We intentionally did not include general keywords such as "artificial intelligence", since our aim is to identify work that specifically involves LLMs rather than the broader AI literature (as seen in prior reviews of human–AI interaction~\cite{zhengUXResearchConversational2022, duanTrustingAutonomousTeammates2025, mehrotraSystematicReviewFostering2024}). Similarly, we did not enumerate additional model names such as Gemini or LLaMA, and instead opted for broad conceptual terms (e.g., large language model) to ensure comprehensive field coverage without overfitting to specific model families~\cite{tangLLMIntegrationExtended2025}.

\subsubsection{Database Selection}

Given the interdisciplinary nature of our review, which integrates perspectives from both HCI and robotics, we examined high-impact publication venues in the two fields using Google Scholar's ``Top Publications'' lists for HCI\footnote{\url{https://scholar.google.com/citations?view_op=top_venues&hl=en&vq=eng_humancomputerinteraction}} and Robotics\footnote{\url{https://scholar.google.com/citations?view_op=top_venues&hl=en&vq=eng_robotics}}. We identified the ACM DL as the primary repository for HCI research and IEEE Xplore as the leading database for robotics publications. Therefore, we selected ACM DL and IEEE Xplore as our main databases. To broaden the scope and include interdisciplinary perspectives crucial to HRI, the search was expanded following consultation with subject matter experts. This expansion included four high-impact, cross-disciplinary publication venues: Nature, Science Robotics, Computers in Human Behavior, and the International Journal of Social Robotics.

\subsection{Screening and Selection}

\subsubsection{Inclusion and Exclusion Criteria}

The screening and selection process was conducted in multiple stages to systematically refine the initial pool of literature down to a final, relevant corpus. The process was guided by a predefined set of inclusion and exclusion criteria, applied consistently by two independent reviewers.

To be included in the final review, a study had to meet several key criteria. Specifically, the publication was required to be a full-length, peer-reviewed research article in English that presented an empirical study on the integration of an LLM with a robotic system for an HRI application \textbf{(Section~\ref{subsubsec:HRI})}. A crucial inclusion criterion was the presence of an embodied agent, which we defined as either a physical robot or a high-fidelity simulated proxy that allows for spatial and interactive presence, such as avatars in VR or AR environments \textbf{(Section~\ref{subsubsec:Robot})}. Conversely, studies were excluded based on several factors. We excluded non-empirical works such as literature reviews, surveys, conceptual frameworks, and theoretical discussions that lacked a direct robotic application. Articles were also removed if LLMs were only mentioned superficially (e.g., in background or future work sections) rather than being a core component of the reported system~\textbf{(Section~\ref{subsubsec:LLM})}. Furthermore, we excluded studies that focused purely on technical improvements or system architecture without involving a user interaction or evaluation process~\textbf{(Section~\ref{subsubsec:Robot})}. Research centered on purely disembodied agents, like text-based chatbots, was not included~\textbf{(Section~\ref{subsubsec:Robot})}, nor were non-full papers such as workshop articles, technical reports, or abstracts.

\subsubsection{Resolution of Disagreements}

During screening and coding, disagreements between the two primary reviewers were resolved through a staged process. The coders first discussed each discrepancy to reach consensus. If agreement could not be achieved, they discussed with a third reviewer to ensure reliability.

The first point of disagreement concerned papers from Scientific Reports. Although labeled as a "report'', this venue predominantly publishes full empirical studies rather than non-refereed reports. After discussion, we included two papers that satisfied the criteria~\cite{loLLMbasedRobotPersonality2025a, herathFirstImpressionsHumanoid2025}. In total, sixteen papers produced inclusion–exclusion disagreements. For example, Hsu et al.~\cite{hsuResearchCareReflection2025} present a three-year robot study with people living with dementia, shifting from WoZ control to GPT-driven autonomy. It delivers actionable "research as care''guidelines, bridging technical HRI design with humanistic care for vulnerable populations. Therefore, we include it for unique insights on adapting LLMs for translating care ethics into HRI practices. Conversely, papers such as Promises~\cite{gunawanPromisesPromisesUnderstanding2025} were excluded. Although participants interacted with a commercially available LLM-enabled robot (e.g., Moxie), the study did not analyze or evaluate the LLM components themselves, nor were LLMs methodologically central to the research design. Through adjudication, eleven papers were ultimately included, and five were excluded.

Given that the disagreement rate was higher than expected, we additionally conducted a false-negative check to assess whether any relevant papers might have been mistakenly excluded. We randomly sampled 100 papers from the 306 studies that had been explicitly excluded during screening~\footnote{Sampling was performed using a random integer generator with uniform sampling without replacement.}. Two coders independently re-evaluated these papers against the inclusion criteria. Only three borderline papers were identified as potentially relevant; however, closer examination showed that one primarily centered on dataset construction~\cite{liToD4IRHumanisedTaskOriented2022}, while the other two relied on video or image demonstration and therefore lacked any real interaction with a physical robot~\cite{schnitzerPrototypingZoomorphicInteractive2024a, lemasurierTemplatedVsGenerative2024a}.

During the first round of open coding, we noticed that certain categories produced a higher number of disagreements. For instance, categories such as Contextual Perception and Understanding and Evaluation Metrics were sometimes defined implicitly by authors, which made interpretation less consistent across coders. In the case of usability, for example, some papers reported standardized instruments such as the system usability scale~\cite{kodurExploringDynamicsHumanRobot2025, dellannaSONARAdaptiveControl2024, geGenComUIExploringGenerative2025, wangCrowdBotOpenenvironmentRobot2024}, whereas others embedded usability-related assessments within broader interview or questionnaire responses~\cite{cuiNoRightOnline2023a, starkDobbyConversationalService2024}. To quantify the degree of coder agreement, we computed inter-rater consistency (IRR) using Cohen's kappa coefficient~\cite{cohenCoefficientAgreementNominal1960}. By contrast, highly observable categories such as modality (IRR = 0.768), robot morphology (IRR = 0.894), and application domain (IRR = 0.904) demonstrated strong agreement due to their concrete and surface-level characteristics. To address lower-agreement categories, we held a calibration meeting and refined the operational definitions before conducting a second coding pass. This resulted in substantially improved IRR scores. Finally, all the categories, subcodes, and IRR values, is provided in Table~\ref{tab:codebook}.

\subsection{Final Corpus}
\subsubsection{Paper Selection Process}
\begin{figure}[t] 
    \centering 
    \includegraphics[width= 1\linewidth]{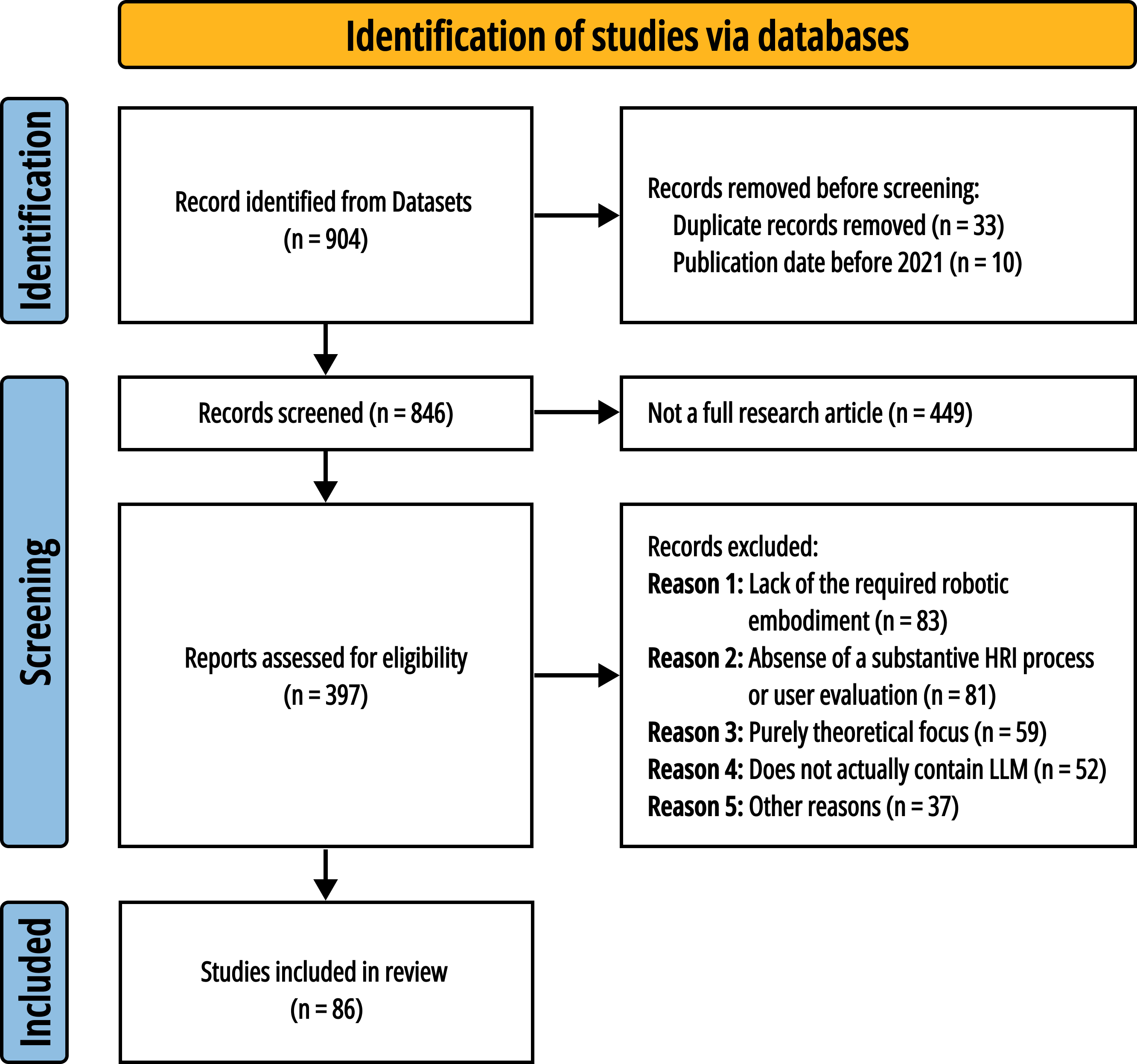} 
    \caption{PRISMA flow diagram outlining the literature screening and inclusion process for this systematic review.} 
    \label{fig:PRISMA} 
    \Description{This PRISMA flow diagram illustrates the process of study selection for a review, which is divided into three main stages: Identification, Screening, and Included. The process begins with the Identification stage, where an initial 904 records were identified from databases. From this set, records were removed before the main screening: 33 were removed for being duplicates, and 10 were removed because they were published before 2021. This left 846 records to enter the Screening stage. During the first part of screening, 449 of these records were excluded because they were not full research articles, leaving 397 reports to be assessed for eligibility. In the next step, these reports were further evaluated, and a number were excluded for five specific reasons. Reason 1: 83 reports were excluded for a``Lack of the required robotic embodiment.''Reason 2: 81 were excluded for the``Absence of a substantive HRI process or user evaluation.''Reason 3: 59 were excluded for having a``Purely theoretical focus.''Reason 4: 52 were excluded because they did not``actually contain LLM.''Reason 5: 37 were excluded for``Other reasons.''After this comprehensive filtering process, the final stage, labeled``Included,''shows that a total of 86 studies were included in the final review.}
\end{figure}

A complete PRISMA flow diagram illustrating this entire process is provided in Figure~\ref{fig:PRISMA}. The initial database search returned a total of 904 records. After removing 33 duplicates and 10 articles published before 2021, 846 unique records remained for the initial screening stage. During the title and abstract screening, 449 records were excluded, primarily because they were not full research articles, leaving 397 articles for a full-text eligibility assessment.
In the final full-text review stage, each of the 397 articles was read in its entirety. This rigorous assessment resulted in the exclusion of a further 311 articles. The primary reasons for exclusion were a lack of the required robotic embodiment (83 studies), the absence of a substantive HRI process or user evaluation (81 studies), and a purely theoretical focus, such as being a survey or conceptual framework (59 studies). Additionally, 52 studies did not substantively integrate an LLM in their system, and a further 37 were excluded for other reasons, such as having a primary goal of dataset collection rather than HRI investigation, or meeting multiple exclusion criteria simultaneously. This multi-stage selection process resulted in a final corpus of 86 studies, which form the foundation of our systematic analysis. 

\subsubsection{Overview of Included Papers}
\begin{figure}[h]
    \centering
    \includegraphics[width= 1\linewidth]{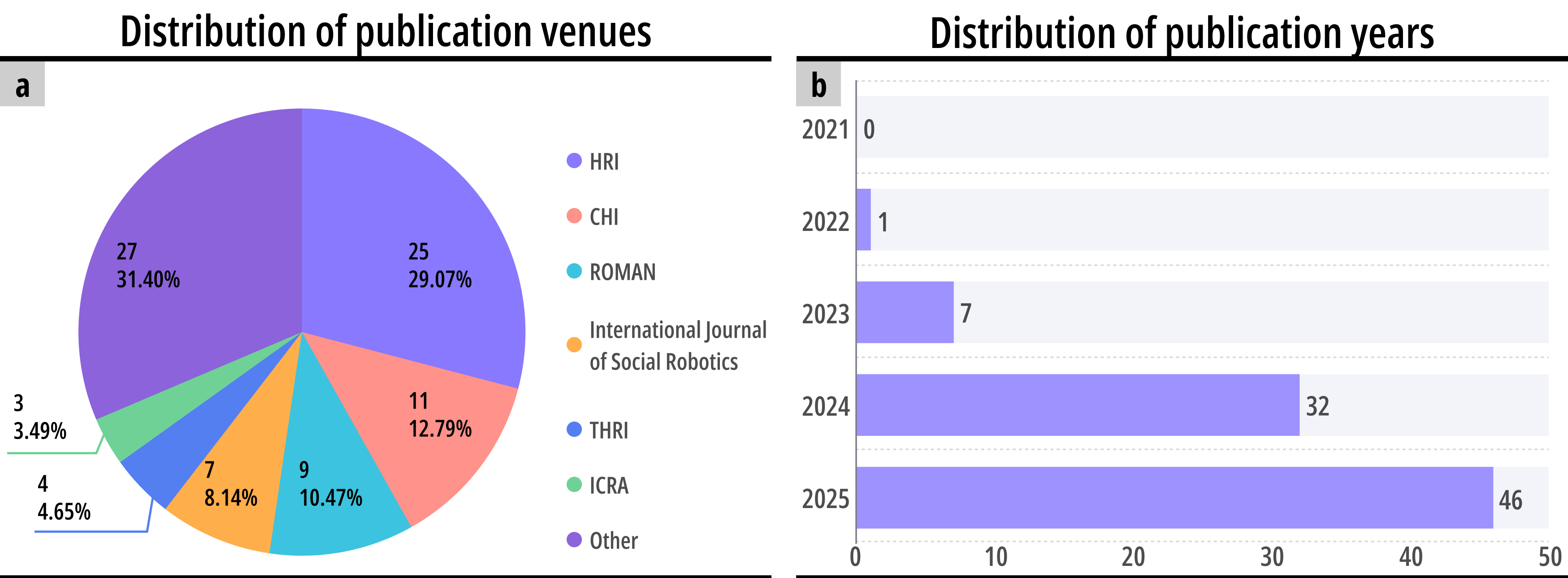}
    \caption{Overview of publication venues and years: (a) Distribution of included papers by venue. Venues contributing fewer than two included papers were not reported individually and are grouped under ``Other'' to ensure a meaningful representation. (b) Annual numbers of included papers.}
    \Description{This composite figure consists of two subplots: (a) A pie chart (titled "Distribution of publication venues") illustrating the proportional distribution of the 86 included papers across their publication venues. Specifically, the venues account for the following shares: HRI (31.40\%), CHI (12.79\%), ROMAN (10.47\%), *International Journal of Social Robotics* (8.14\%), THRI (4.65\%), ICRA (3.49\%), and "Other''(29.07\%)—where "Other''groups venues that contributed fewer than two included papers to ensure a meaningful proportional representation. (b) A bar chart (titled "Distribution of publication years") presenting the annual counts of included papers from 2021 to 2025: 0 papers in 2021, 1 in 2022, 7 in 2023, 32 in 2024, and 46 in 2025, reflecting a notable upward trend in publications over the period.}
    \label{fig:PubVenuesAndYear}
\end{figure}

The distribution of publication venues and years for the papers included in this review is presented in Figure~\ref{fig:PubVenuesAndYear}. An analysis of the publication venues reveals that the included papers are disseminated across premier conferences and journals in the intersecting fields, which underscores the highly interdisciplinary nature of research on LLM-driven HRI research. Notably, publications from the ACM/IEEE International Conference on Human-Robot Interaction and the ACM CHI Conference on Human Factors in Computing Systems collectively account for the largest proportion of the literature included in our review. Regarding the publication time in Figure~\ref{fig:PubVenuesAndYear}.b, the number of included publications demonstrates a marked increase over time, with a significant concentration of works appearing in 2024 and 2025.

\begin{table*}[t]
\centering
\small
  \caption{The final codebook with 9 code categories and 60 subcodes; the average IRR is computed across the subcodes for each code. Multiple refers to ``multiple codes can apply''. In ``Methodology'', ``Other Methods'' occurred too infrequently to be included in the quantitative analysis.}
  \label{tab:codebook}

  \Description{This table presents the classification framework (encompassing 9 core analytical categories) used for literature coding in the LLM-driven Human-Robot Interaction (HRI) systematic review, along with the corresponding coding items (Codes), mean inter-rater reliability (Mean IRR, a metric reflecting coding consistency) and whether multiple selections are allowed for each category (Multiple): (1) Contextual Perception and Understanding: Includes 6 coding items (e.g., Static and Semi-Static Context Injection, Emotional Grounding), with a mean IRR of 0.720 (SD=0.108), and multiple selections are permitted. (2) Generative and Agentic Interaction: Comprises 6 coding items (e.g., Persona Adaptation, Anticipatory Assistance), with a mean IRR of 0.796 (SD=0.073), and multiple selections are allowed. (3) Iterative Optimization and Alignment: Covers 5 coding items (e.g., Sustained Personalization, Ethical Repair), with a mean IRR of 0.684 (SD=0.070), and multiple selections are permitted. (4) Modality and Interaction Channels: Includes 7 coding items (e.g., Text, Tangible and Haptic Interaction), with a mean IRR of 0.768 (SD=0.070), and multiple selections are allowed. (5) Morphology: Consists of 5 coding items (e.g., Humanoid, AR/VR), with a mean IRR of 0.894, and multiple selections are not permitted. (6) Autonomy: Contains 3 coding items (e.g., Full Autonomy, Teleoperation), with a mean IRR of 0.846, and multiple selections are not allowed. (7) Methodology: Includes 7 coding items (e.g., Laboratory Experiment, Co-Design Workshops), with a mean IRR of 0.684 (SD=0.172), and multiple selections are permitted. (8) Evaluation Metrics: Comprises 9 coding items (e.g., Task Efficiency, Perceived Intelligence), with a mean IRR of 0.832 (SD=0.100), and multiple selections are allowed. (9) Application: Covers 8 coding items (e.g., Healthcare, Industrial Manufacturing), with a mean IRR of 0.904, and multiple selections are not permitted.}

\begin{tabular}{p{0.15\textwidth} p{0.60\textwidth} p{0.12\textwidth} p{0.05\textwidth}}
    \toprule
    \textbf{Category} & \textbf{Codes} & \textbf{Mean IRR} & \textbf{Multiple} \\
    \midrule

    Contextual Perception and Understanding &
    Static and Semi-Static Context Injection; Modular Perception and Textual Abstraction; Integrated Visual-Language Reasoning; Emotional Grounding; Task Intent Formulation; Human Model Alignment &
    0.720 (SD=0.108) &
    Yes \\

    Generative and Agentic Interaction &
    Persona Adaptation and Conversational Fluidity; Embodied Social Expressiveness; Task-Oriented Planning and Execution; Creative Storytelling and Social Engagement; Social Initiation; Anticipatory Assistance &
    0.796 (SD=0.073) &
    Yes \\

    Iterative Optimization and Alignment &
    Sustained Personalization; Episodic Memory Integration; Behavioral Repair in Task Execution; Emotional Repair in Social Interaction; Repair in Ethical and Normative Alignment &
    0.684 (SD=0.070) &
    Yes \\

    Modality and Interaction Channels &
    Text; Voice; Visuals; Motion; Hybrid; Tangible and Haptic Interaction; Proximity &
    0.768 (SD=0.070) &
    Yes \\

    Morphology &
    Humanoid; Functional; Zoomorphic; Desktop Companions; AR/VR &
    0.894 &
    No \\

    Autonomy &
    Full Autonomy; Semi-Autonomy; Teleoperation &
    0.846 &
    No \\

    Methodology &
    Laboratory Experiment; Field Deployments; Interviews; Questionnaires; Technical Evaluation; Other Methods (WoZ; Case Study; Simulation; Co-Design Workshops; BodyStorming; Think-Aloud Protocols) &
    0.684 (SD=0.172) &
    Yes \\

    Evaluation Metrics &
    Task Efficiency and Timing; Task Accuracy and Performance; LLM-Specific Performance; User's Perceptual and Relational Experience; Perceived Intelligence; Anthropomorphism; Usability; Safety; Cognitive Load and Workload &
    0.832 (SD=0.100) &
    Yes \\

    Application &
    Social and Conversational Systems; Healthcare and Wellbeing; Domestic and Everyday Use; Teaching and Education; Industrial Manufacturing; AR/VR-enabled Interactions; Public Spaces Service; Other &
    0.904 &
    No \\

    \bottomrule
  \end{tabular}
\end{table*}
\section{Large Language Models in Human-Robot Interaction} \label{section:LLMs-HRI}
\begin{figure*}[ht]
    \centering
    \includegraphics[width= 1\linewidth]{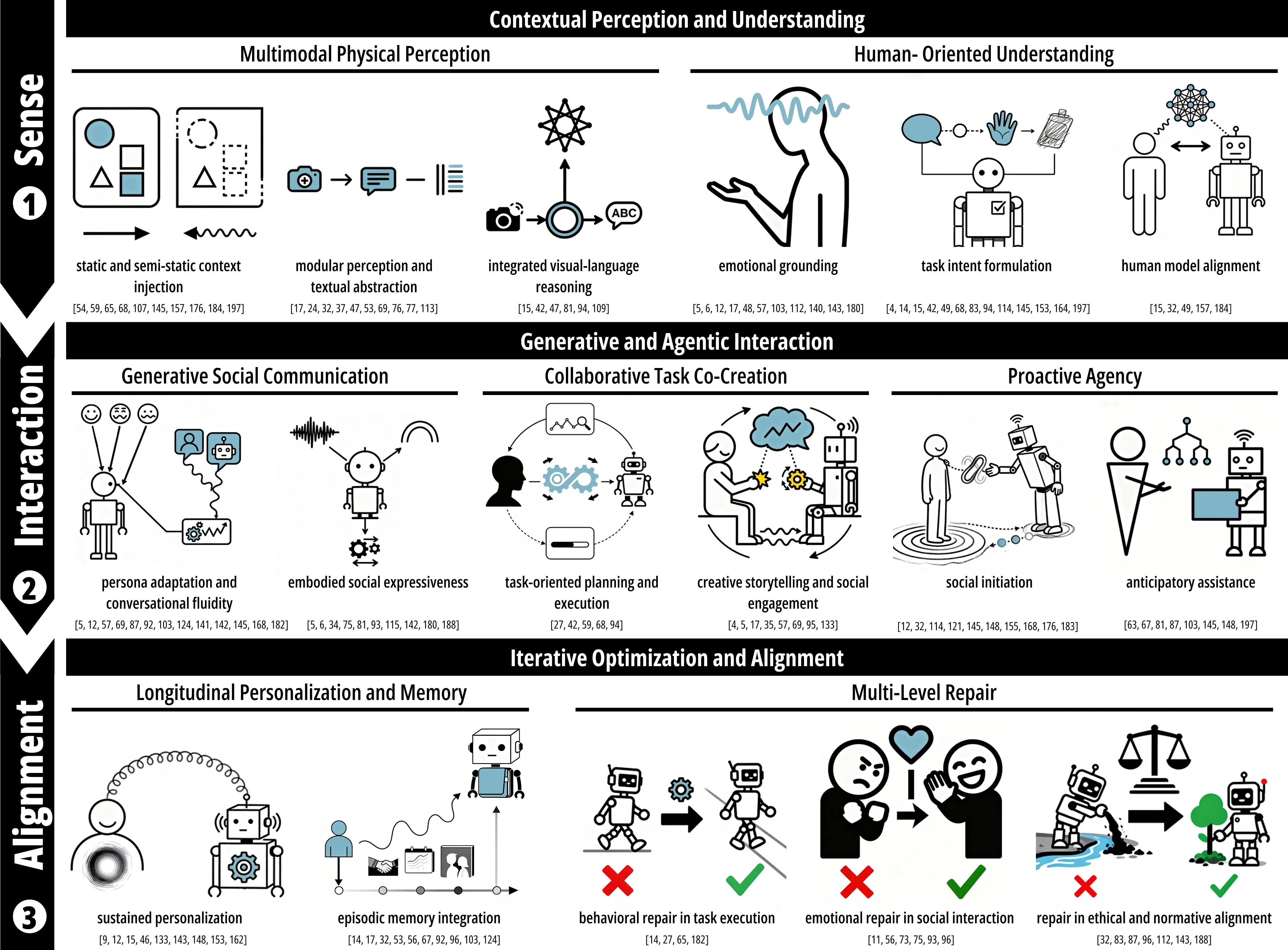}
    \caption{The proposed Sense-Interaction-Alignment framework for LLM-driven HRI research. This model adapts classical robotic paradigms to address the unique demands of embodied, social collaboration. It transitions from context grounding (Sense) and generative, multi-agent co-creation (Interaction), to continuous iterative optimization (Alignment).}
    \Description{Diagram illustrating the Sense-Interaction-Alignment framework for LLM-driven Human-Robot Interaction (HRI), structured into 3 core stages:(1) Sense (Contextual Perception and Understanding): Split into ``Multimodal Physical Perception'' (sub-items: static/semi-static context injection, modular perception \& textual abstraction, integrated visual-language reasoning) and ``Human-Oriented Understanding'' (sub-items: emotional grounding, task intent formulation, human model alignment), each paired with illustrative icons and citation identifiers. (2) Interaction (Generative and Agentic Interaction): Covering ``Generative Social Communication'' (persona adaptation, embodied social expressiveness), ``Collaborative Task Co-Creation'' (task-oriented planning/execution, creative storytelling), and``Proactive Agency'' (social initiation, anticipatory assistance), with matching icons and citations. (3) Alignment (Iterative Optimization and Alignment): Encompassing``Longitudinal Personalization and Memory'' (sustained personalization, episodic memory integration) and ``Multi-Level Repair'' (behavioral repair, emotional repair, ethical/normative alignment repair), each with icons, status indicators (red X/green check), and citations.}
    \label{fig:Sense-Interaction-Alignment}
\end{figure*}

In this section, we aim to answer \textbf{RQ1}. Overall,
LLMs have endowed robots with a formidable cognitive foundation, characterized by zero-shot capabilities, complex reasoning, and in-context learning \cite{BrownLanguagemodels2020, talmor2022commonsenseqa, WeiChainofthought2022}. However, transposing these capabilities from disembodied text processing to the physical reality of HRI involves multiple transformations. To structure this transition, we adapt the classical ``Sense-Plan-Act''~\cite{murphyIntroductionAIRobotics2001} and ``Reason + Act'' ~\cite{yao2022react} paradigms into a ``Sense-Interaction-Alignment'' framework, as seen in Figure~\ref{fig:Sense-Interaction-Alignment}.
In the \textbf{Sense} phase, we argue that robotic systems ground abstract LLM capabilities within specific physical and social contexts to achieve true embodied intelligence, distinguishing them from disembodied AI agents~\cite{liuAligningCyberSpace2025, Incao_2025}. Second, we reframe the traditional notion of ``Action'' as \textbf{Interaction}, positing that LLM-driven behaviors are not solitary executions but proactive, multi-agent collaborations. Finally, we introduce \textbf{Alignment} as the critical adaptive phase, where the ``Human in the Loop'' (HITL) \cite{firminodesouzaTrustTrustworthinessHumanCentered2025} necessitates continuous optimization through personalization and repair mechanisms to ensure robot behaviors remain congruent with human needs and social norms over time.

\subsection{Contextual Perception and Understanding}\label{subsection:Perception&Understanding}
In the \textbf{Sense} phase, LLMs transcend raw data processing to construct a semantic understanding of the environment. This section elucidates the transformation of sensory inputs into actionable cognitive contexts~\cite{leusmannInvestigatingLLMDrivenCuriosity2025}, progressing from the perception of physical surroundings to the understanding of complex social and affective dynamics.

\subsubsection{ Multimodal Physical Perception}\label{subsubsection:MultimodalPhysicalPerception}
The prerequisite for embodied intelligence is the ability to perceive and semantically interpret physical space. Unlike traditional robotics, which relies on rigid metric representations, LLMs facilitate the semantic parsing of environmental cues. We categorize these grounding strategies by their degree of semantic integration.

- \textbf{\textit{Static and Semi-Static Context Injection.}}
Primitive approaches rely on manual context injection, embedding environmental constraints directly into system prompts~\cite{padmanabhaVoicePilotHarnessingLlms2024, starkDobbyConversationalService2024, vermaTheoryMindAbilities2024, zhangLargeLanguageModels2023b, ikedaMARCERMultimodalAugmented2025a, yanoUnifiedUnderstandingEnvironment2024} or utilizing pre-configured semantic maps~\cite{zuLanguageSketchingLLMdriven2024}. These methods often incorporate granular object details, such as the dimensions of laboratory equipment~\cite{karliAlchemistLLMAidedEndUser2024a} and spatial relationships~\cite{hoSETPAiREdDesigningParental2025, jinRobotGPTRobotManipulation2024}. While effective in controlled settings, their reliance on ``semi-static information'' creates a bottleneck, limiting adaptation in dynamic or unmapped environments where real-time updates are essential~\cite{yanoUnifiedUnderstandingEnvironment2024}.

- \textbf{\textit{Modular Perception and Textual Abstraction.}}
To overcome the limitations of static prompts, researchers have adopted dynamic Sensor-to-Text pipelines~\cite{grassiGroundingConversationalRobots2024, laiNaturalMultimodalFusionBased2025, herathFirstImpressionsHumanoid2025}. Systems like ARECA translate quantitative metrics (e.g., temperature, location) into narrative descriptions~\cite{choARECADesignSpeculation2023}. Concurrently, modular vision algorithms (e.g., YOLO, SAM) extract object labels~\cite{laiNaturalMultimodalFusionBased2025, blancoAIenhancedSocialRobots2024}, while automatic speech recognition converts audio into text~\cite{latifPhysicsAssistantLLMpoweredInteractive2024, perella-holfeldParentEducatorConcerns2024a, kimUnderstandingLargelanguageModel2024d}. These text-based percepts serve as intermediate abstractions, enabling LLMs to perform high-level reasoning—such as determining object saliency from feature lists—without processing raw visual data directly~\cite{dellannaSONARAdaptiveControl2024, ferriniPerceptsSemanticsMultimodala}.

- \textbf{\textit{Integrated Visual-Language Reasoning.}}
Advanced implementations are increasingly replacing intermediate textual abstractions with VLMs for direct scene interpretation~\cite{bastinGPTAllySafetyorientedSystem2025}. This shift enables real-time interaction captioning~\cite{grassiGroundingConversationalRobots2024, mahadevanImageInThatManipulatingImages2025, geGenComUIExploringGenerative2025} and the integration of dynamic knowledge graphs for affordance planning~\cite{panACKnowledgeComputationalFramework2025}. Hybrid approaches further bridge precision and reasoning by combining fiducial markers (e.g., ArUco) with MLLMs to support complex behaviors like curiosity-driven exploration~\cite{leusmannInvestigatingLLMDrivenCuriosity2025}. However, the reliance on converting multimodal inputs into a unified semantic space remains a significant challenge for achieving deep, lossless multimodal fusion. We will discuss this further in subsequent sections.

\subsubsection{Human-Oriented Understanding} \label{subsection:Human-Oriented}
Beyond physical environment, effective HRI requires navigating the nuanced landscape of social interaction. Here, LLMs enable robots to shift from determining ``what is there'' to deciphering ``who is there'' and ``why,''  processing invisible states such as emotion, intent, and social model.

- \textbf{\textit{Emotional Grounding.}} 
Social understanding begins with the interpretation of emotional cues. Systems increasingly employ multimodal fusion, inferring affective states by combining facial detection with LLM-based textual analysis~\cite{nardelliIntuitiveInteractionCognitive2025, yuImprovingPerceivedEmotional2024, sieversInteractingSentimentalRobot2024, blancoAIenhancedSocialRobots2024} and incorporating temporal audio cues for greater accuracy~\cite{pereiraMultimodalUserEnjoyment2024a, spitaleVITAMultiModalLLMBased2025a}. Recent research pushes towards ``empathic grounding,''enabling robots to grasp complex nuances like nostalgia or implicit regret~\cite{hsuBittersweetSnapshotsLife2025a, bannaWordsIntegratingPersonality2025, antonyXpressSystemDynamic2025a, grassiEnhancingLLMBasedHumanRobot2024}. This macro-level understanding, often spanning multiple dialogue turns, is critical for sensitive applications such as longitudinal wellbeing monitoring~\cite{arjmandEmpathicGroundingExplorations2024}.

- \textbf{\textit{Task Intent Formulation.}} 
Moving from emotion to pragmatics, LLMs facilitate the separation of explicit task requests from broader communicative intent~\cite{pinedaSeeYouLater2025a, starkDobbyConversationalService2024, grassiStrategiesControllingConversation2025}. For explicit commands, LLMs parse unstructured language into rigid specifications (e.g., task type, time, equipment)~\cite{wangCrowdBotOpenenvironmentRobot2024, karliAlchemistLLMAidedEndUser2024a, bastinGPTAllySafetyorientedSystem2025}. This capability extends to multimodal inputs, allowing robots to infer goals from sketches~\cite{geGenComUIExploringGenerative2025, zuLanguageSketchingLLMdriven2024} or body language~\cite{liStargazerInteractiveCamera2023, bassiounyUJIButlerSymbolicNonsymbolic2025}. Furthermore, context-aware systems can predict task progression, such as detecting completed workflow steps~\cite{taoLAMSLLMDrivenAutomatic2025a, mahadevanImageInThatManipulatingImages2025}, or supporting creative intents for inclusive interactions with neurodiverse populations~\cite{aliInclusiveCocreativeChildrobot2025a}.

- \textbf{\textit{Human Model Alignment.}} 
The highest level of social cognition involves internalizing implicit rules of engagement. At the conversational level, LLMs identify subtle shifts, such as intentions to change topics~\cite{grassiStrategiesControllingConversation2025, bastinGPTAllySafetyorientedSystem2025}. More broadly, hybrid architectures enforce social compliance by integrating interpretation rules with prohibitions and obligations~\cite{dellannaSONARAdaptiveControl2024}. At a deeper level, LLMs provide a pathway to Theory-of-Mind capabilities, serving as Zero-Shot Human Models that simulate ``what the human would think,'' thereby enabling robots to predict and align with complex social dynamics~\cite{vermaTheoryMindAbilities2024, zhangLargeLanguageModels2023b}.

\subsection{Generative and Agentic Interaction} \label{subsection:Generative&Agentic}
In the \textbf{Interaction} phase, LLMs fundamentally reshape HRI from rigid command-response loops to fluid, generative, and agentic collaborations. Unlike traditional systems constrained by pre-scripted behaviors, LLM-driven robots exhibit the capacity to generate novel social signals, co-create complex plans, and autonomously initiate interactions based on environmental context.

\subsubsection{Generative Social Communication} \label{subsubsection:GenerativeSocial}
LLMs empower robots to transcend static dialogue trees, enabling communication that is emotionally tailored, stylistically adaptive, and multimodally expressive~\cite{kimUnderstandingLargelanguageModel2024d, starkDobbyConversationalService2024}.

- \textbf{\textit{Persona Adaptation and Conversational Fluidity.}} 
Effective social communication begins with linguistic adaptation, where robots dynamically modulate their personality rather than relying on generic responses~\cite{nardelliIntuitiveInteractionCognitive2025, antonyXpressSystemDynamic2025a}. Researchers achieve this by conditioning models on psychological frameworks like the Big Five traits~\cite{zhangExploringRobotPersonality2025a, bannaWordsIntegratingPersonality2025, loLLMbasedRobotPersonality2025a}. This allows robots to manifest distinct personas adapted to specific social roles—from ``cheerfully ironic'' tones that increase warmth~\cite{sieversIntroducingNoteLevity2024} to self-disclosing styles suitable for health mediation~\cite{westerFacingLLMsRobot2024a} or emotional elicitation in elder care~\cite{hsuBittersweetSnapshotsLife2025a, limaPromotingCognitiveHealth2025}. Sustaining these personas requires moving beyond slot-filling systems to LLMs that process unstructured inputs for context-aware coherence~\cite{kimUnderstandingLargelanguageModel2024d, starkDobbyConversationalService2024}. Modern systems thus focus on maintaining conversational flow by analyzing dialogue history and syntactic cues to predict precise turn-taking moments, ensuring social presence remains believable~\cite{rosenPreviousExperienceMatters2024, skantzeApplyingGeneralTurntaking2025}.

- \textbf{\textit{Embodied Social Expressiveness.}} 
Linguistic fluency alone is insufficient for embodied presence; the core challenge lies in synchronizing verbal output with non-verbal behaviors~\cite{zhangPromptingEmbodiedAI2025}. Moving beyond isolated motion generation, recent systems leverage LLMs to orchestrate holistic behavioral responses, simultaneously outputting verbal utterances, emotional states, and physical cues—such as head nodding—to facilitate empathic grounding~\cite{arjmandEmpathicGroundingExplorations2024, mahadevanGenerativeExpressiveRobot2024}. This coordination is refined by control strategies that align physical signals with linguistic intent, such as using gaze aversion to mark turn-taking boundaries~\cite{pintoPredictiveTurntakingLeveraging2024, skantzeApplyingGeneralTurntaking2025} or synchronizing facial expressions with utterance sentiment~\cite{antonyXpressSystemDynamic2025a}. To support this expressivity without compromising real-time performance, architectures often hybridize generative LLM planning with rule-based execution, minimizing latency while maximizing social impact~\cite{leusmannInvestigatingLLMDrivenCuriosity2025, yuImprovingPerceivedEmotional2024, kontogiorgosQuestioningRobotUsing2025, elfleetInvestigatingImpactMultimodal2024}.

\subsubsection{Collaborative Task Co-Creation}\label{subsubsection:CollaborativeTask}
LLMs transform robots from passive tools into active partners, with this collaborative paradigm evident in both goal-oriented tasks and open-ended, creative undertakings.

- \textbf{\textit{Task-Oriented Planning and Execution.}} 
Collaboration in physical tasks has shifted towards shared autonomy, where agency is distributed between human and machine~\cite{cuiNoRightOnline2023a}. Systems like GenComUI visualize LLM-generated plans, allowing users to verify task flows before action~\cite{geGenComUIExploringGenerative2025}. During execution, frameworks like LILAC enable users to provide natural language corrections that update the robot's control space in real-time, facilitating the learning of complex manipulations from minimal demonstrations~\cite{cuiNoRightOnline2023a}. This co-authorship is further supported by multimodal tools bridging abstract intent and precise control via augmented reality and timeline adjustments~\cite{karliAlchemistLLMAidedEndUser2024a, ikedaMARCERMultimodalAugmented2025a, mahadevanImageInThatManipulatingImages2025}.

- \textbf{\textit{Creative Storytelling and Social Engagement.}} 
In open-ended domains, robots utilize LLMs as creativity scaffolds to foster engagement through joint imagination. Rather than passively delivering content, systems like Jibo actively co-construct stories with children by offering divergent narrative ideas~\cite{aliInclusiveCocreativeChildrobot2025a, elgarfCreativeBotCreativeStoryteller2022, malnatskyFittingHumorAgeBased2025}. This paradigm extends to therapeutic contexts, where robots generate personalized narratives dynamically adapted to the cognitive needs of older adults~\cite{blancoAIenhancedSocialRobots2024, hsuBittersweetSnapshotsLife2025a}, transforming social interaction into a co-creative and cognitively stimulating experience~\cite{antonyXpressSystemDynamic2025a, kimUnderstandingLargelanguageModel2024d, shenSocialRobotsSocial2025}.

\subsubsection{Proactive Agency}\label{subsubsection:ProactiveAgency}
The emergence of social agency from LLMs allows robots to initiate actions based on inferred context, shifting the interaction dynamic from reactive execution to proactive engagement.

- \textbf{\textit{Social Initiation.}} 
Research suggests that users prefer robots that proactively communicate their capabilities, enabling smoother cooperation~\cite{reimannWhatCanYou2025a}. Proactive robots bridge the gap between presence and interaction by autonomously identifying and approaching users in socially appropriate ways~\cite{starkDobbyConversationalService2024, westerFacingLLMsRobot2024a, bannaWordsIntegratingPersonality2025, yanoUnifiedUnderstandingEnvironment2024, tsushimaTaskPlanningFactory2025}. Beyond physical initiation, systems like SONAR exhibit ``proactive social agency'' by engaging in small talk during warm-up phases~\cite{pinedaSeeYouLater2025a, dellannaSONARAdaptiveControl2024}, or employing ``situation controlling'' strategies to manage user expectations before interaction begins~\cite{reimannWhatCanYou2025a, suChatAdpChatGPTpoweredAdaptation2024}. However, balancing agency with user control is critical; while low-granularity adjustments align with user preferences, excessive granular control can undermine the perception of the robot as an intelligent social agent~\cite{zhangBalancingUserControl2025a}.

- \textbf{\textit{Anticipatory Assistance.}} 
The agency further manifests through the anticipation of user needs. Robots leverage multimodal perception to proactively offer assistance or initiate context-aware conversations~\cite{suChatAdpChatGPTpoweredAdaptation2024, limaPromotingCognitiveHealth2025, kamelabadComparingMonolingualBilingual2025}. This extends to emotionally intelligent behaviors, such as taking the initiative to elicit positive states based on user personality~\cite{nardelliIntuitiveInteractionCognitive2025}, or employing curiosity-driven strategies to bridge knowledge gaps~\cite{leusmannInvestigatingLLMDrivenCuriosity2025}. By autonomously investigating uncertainties, robots evolve into active partners that not only answer questions but strategically ask them to deepen mutual understanding~\cite{itoRobotDynamicallyAsking2025a}.
To maintain alignment during such proactive behaviors, agents explicitly inform users of completed actions via dialogue cues~\cite{starkDobbyConversationalService2024} and actively seek clarification when facing ambiguity~\cite{zuLanguageSketchingLLMdriven2024}. In the event of errors, structured feedback detailing the cause allows users to understand both the outcome and the underlying reasoning, fostering trust~\cite{leusmannInvestigatingLLMDrivenCuriosity2025}.

\subsection{Iterative Optimization and Alignment}
\label{subsection:Optimization&Alignment}
While sense and interaction provide the operational capability for behavior, they do not guarantee its long-term suitability. Generative interaction introduces risks of hallucination, misalignment, or drift. The \textbf{Alignment} phase addresses the critical feedback loops required to optimize interaction, ensuring it remains safe, ethical, and personalized over time.

\subsubsection{Longitudinal Personalization and Memory} \label{subsubsection:Personalization&Memory}

To achieve embodied intelligence, robots transition from episodic interactions to continuous relationships. This process relies on active personalization strategies supported by persistent memory repositories.

- \textbf{\textit{Sustained Personalization.}} 
Personalization encompasses the long-term optimization of system parameters beyond immediate reactions~\cite{bannaWordsIntegratingPersonality2025, goubardCognitiveModellingVisual, shenSocialRobotsSocial2025}. While short-term adaptation handles real-time adjustments for safety~\cite{bastinGPTAllySafetyorientedSystem2025} or dialogue pacing~\cite{axelssonYouFollow2023}, true embodiment requires refining decision-making models over extensive periods. Addressing the data-inefficiency of traditional Reinforcement Learning, recent advancements leverage LLMs to accelerate policy tuning. For instance, ChatAdp uses ChatGPT to generate synthetic feedback, creating personalized training data with minimal human effort~\cite{suChatAdpChatGPTpoweredAdaptation2024}. Additionally, frameworks like LAMS allow users to explicitly teach robots new logic through natural language~\cite{taoLAMSLLMDrivenAutomatic2025a}, while systems like VITA evolve dedicated user models to align with personality traits over time~\cite{spitaleVITAMultiModalLLMBased2025a, wangChallengesAdoptingCompanion2025}.

- \textbf{\textit{Episodic Memory Integration.}} 
Long-term personalization is also founded on the retention of shared history~\cite{hsuResearchCareReflection2025, herathFirstImpressionsHumanoid2025, nardelliIntuitiveInteractionCognitive2025, bassiounyUJIButlerSymbolicNonsymbolic2025, rosenPreviousExperienceMatters2024}. Systems utilize sophisticated memory architectures, such as DSR graphs, to accumulate context across sessions, capturing preferred topics and cognitive patterns~\cite{blancoAIenhancedSocialRobots2024, dellannaSONARAdaptiveControl2024}. This historical awareness enables dynamic profiling, allowing robots to tailor coaching strategies or speech rates based on past interactions~\cite{mannavaExploringSuitabilityConversational2024, kamelabadComparingMonolingualBilingual2025}. Operationalizing this involves live memory repositories where dual LLMs form a continuous feedback loop between perception, retrieval, and action~\cite{loLLMbasedRobotPersonality2025a}. Crucially, this process is personality-dependent; the robot's distinct character influences how memories are encoded and retrieved, guiding consistent and appropriate actions~\cite{nardelliIntuitiveInteractionCognitive2025}.

\subsubsection{Multi-Level Repair}
\label{subsubsection:Multi-LevelRepair}
As personalization deepens, so does the need for defensive mechanisms to prevent misalignment. We categorize these mechanisms into three levels: task execution, social interaction, and ethical compliance.

- \textbf{\textit{Behavioral Repair in Task Execution.}} 
When physical tasks fail, repair mechanisms address both control and logic. For real-time control, frameworks like LILAC allow users to refine the control space via natural language corrections~\cite{cuiNoRightOnline2023a}. To preempt errors, systems like UJI-Butler incorporate Human-in-the-Loop verification for planned actions~\cite{bassiounyUJIButlerSymbolicNonsymbolic2025, zhangExploringRobotPersonality2025a}. At the logic level, systems like RobotGPT utilize simulation-based corrector bots to iteratively analyze and fix runtime errors in LLM-generated code until successful execution is achieved~\cite{jinRobotGPTRobotManipulation2024}.

- \textbf{\textit{Emotional Repair in Social Interaction.}} 
In social contexts, breakdowns require strategies to mitigate frustration and preserve trust~\cite{axelssonOhSorryThink2024, hsuResearchCareReflection2025}. Beyond error correction, robots navigate the dynamics of politeness~\cite{kodurExploringDynamicsHumanRobot2025, mannavaExploringSuitabilityConversational2024}. Research emphasizes the use of multimodal signals—such as visual cues or gestures—to resolve misunderstandings effectively~\cite{kontogiorgosQuestioningRobotUsing2025, mahadevanGenerativeExpressiveRobot2024}. Longitudinal studies further suggest that repair strategies evolve with the user relationship, transitioning from generic apologies to complex empathic structures that acknowledge user feelings~\cite{axelssonOhSorryThink2024}.

- \textbf{\textit{Repair in Ethical and Normative Alignment.}} 
Finally, behaviors strictly adhere to societal norms and ethical guidelines. Architectures like SONAR enforce social appropriateness through formal rules regarding prohibitions and obligations. These systems may also employ online learning to adjust fuzzy definitions of norms, such as social distance, based on validated interaction data~\cite{dellannaSONARAdaptiveControl2024, liStargazerInteractiveCamera2023}. Maintaining trust with vulnerable populations particularly requires transparency about limitations to prevent unrealistic expectations~\cite{limaPromotingCognitiveHealth2025, zhangPromptingEmbodiedAI2025, spitaleVITAMultiModalLLMBased2025a} and implementing guardrails for sensitive topics~\cite{mannavaExploringSuitabilityConversational2024}. Balancing this deep personalization with privacy concerns remains a pivotal challenge in ethical alignment~\cite{pereiraMultimodalUserEnjoyment2024a}.

\section{Design Components and Strategies}
\label{sec: design}
Building upon the preceding discussion of LLMs fundamental capabilities in HRI, this section addresses \textbf{RQ2}. Beyond exploring foundational technicalities, we specifically examine the transformative impact of LLMs across three pivotal dimensions: modality, morphology, and autonomy. 

\subsection{Modality and Interaction Channels}
\begin{figure*}[h]
    \centering
    \includegraphics[width= 1\textwidth]{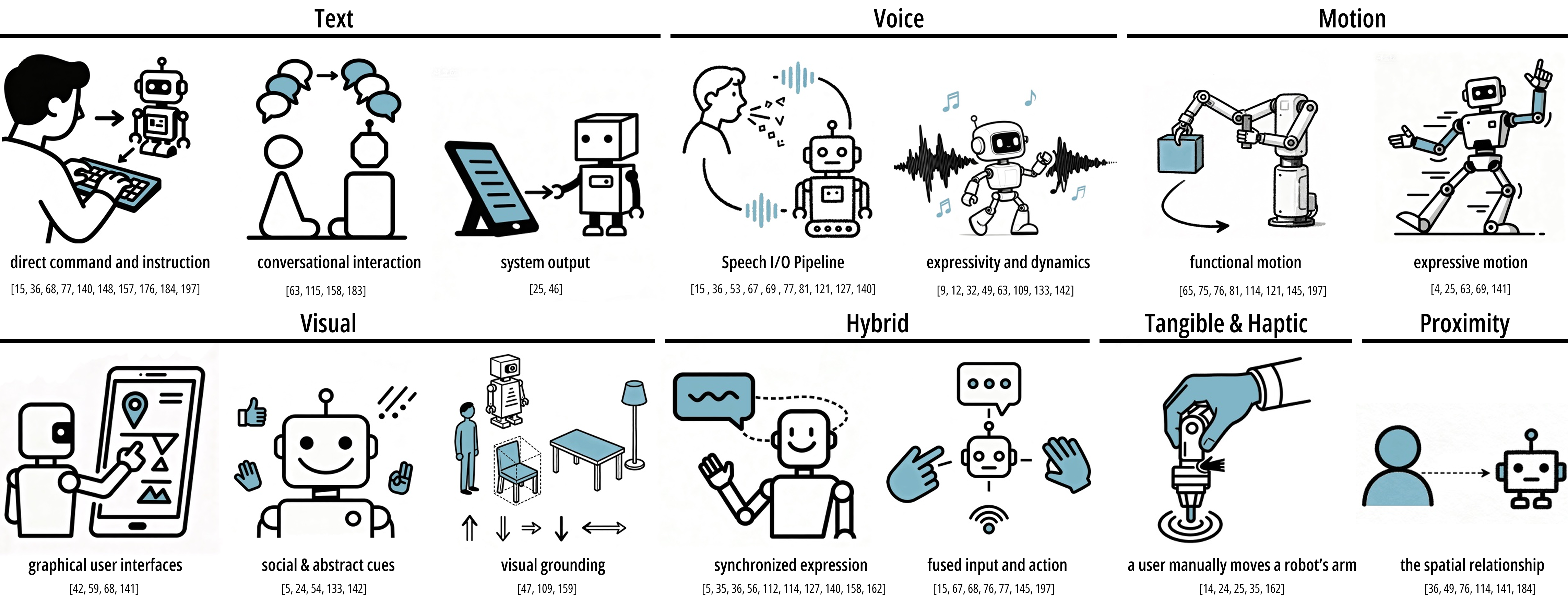}
    \caption{Interaction modalities, through which the human and robot will interact.}
    \label{fig:modality}
    \Description{This diagram classifies interaction modalities in LLM-driven human-robot interaction (HRI) — a core dimension of the systematic review’s taxonomy — with six modality categories, each linked to functional interactions and literature references: (1) Text: Direct command/instruction, conversational interaction, system output. (2) Voice: Speech I/O pipeline, vocal expressivity and dynamics. (3) Motion: Functional motion (task action), expressive motion (social gesture). (4) Visual: Graphical user interfaces, social/abstract cues, visual grounding. (5) Hybrid: Synchronized expression (language + motion), fused input/action. (6) Tangible & Haptic/Proximity: Manual robot arm manipulation, spatial relationship reasoning. Iconographic illustrations depict corresponding robot-human interaction scenarios, and numerical brackets indicate the reviewed studies associated with each functional implementation.}
\end{figure*}

To design and conduct research in LLM-driven HRI, a critical early step is selecting the appropriate modalities, or channels, through which the human and robot will interact. The literature demonstrates a wide array of choices in Figure \ref{fig:modality}, ranging from fundamental text-based commands to complex, multi-sensory experiences.

\subsubsection{Text}

Text serves as the most direct and foundational modality for interacting with LLMs, utilized for both user input and robot output. This includes its use for \textbf{direct command and instruction}, where many systems rely on users providing textual commands through a console or editor~\cite{farooqDAIMHRINewHumanRobot2024, zhangLargeLanguageModels2023b, suChatAdpChatGPTpoweredAdaptation2024, latifPhysicsAssistantLLMpoweredInteractive2024, sieversInteractingSentimentalRobot2024, bastinGPTAllySafetyorientedSystem2025, karliAlchemistLLMAidedEndUser2024a, zuLanguageSketchingLLMdriven2024, yanoUnifiedUnderstandingEnvironment2024, vermaTheoryMindAbilities2024}. Furthermore, text is the core of \textbf{conversational interaction} beyond simple commands, as LLMs are used to generate conversational turns\cite{wangChildRobotRelationalNorm2025a}, rephrase dialogue for specific personas\cite{zhangBalancingUserControl2025a, pintoPredictiveTurntakingLeveraging2024}, and formulate questions\cite{itoRobotDynamicallyAsking2025a}. Finally, robots also employ text for \textbf{system output}, for instance, by displaying instructions or feedback on tablets and e-paper screens\cite{goubardCognitiveModellingVisual, choLivingAlongsideAreca2025}.

\subsubsection{Voice}
Voice enables a wide range of spoken interactions from dialogues and small talk \cite{starkDobbyConversationalService2024, pinedaSeeYouLater2025a} to live, unscripted conversations \cite{pereiraMultimodalUserEnjoyment2024a, kodurExploringDynamicsHumanRobot2025}, with the naturalness often enhanced by the underlying LLM \cite{wangChallengesAdoptingCompanion2025}. The predominant \textbf{Speech I/O Pipeline} converts user speech to text via Automatic Speech Recognition (ASR) for the LLM, then vocalizes responses using Text-to-Speech (TTS) \cite{kimUnderstandingLargelanguageModel2024d, reimannWhatCanYou2025a, farooqDAIMHRINewHumanRobot2024, salemComparativeHumanrobotInteraction2024, latifPhysicsAssistantLLMpoweredInteractive2024, sieversInteractingSentimentalRobot2024, bastinGPTAllySafetyorientedSystem2025, leusmannInvestigatingLLMDrivenCuriosity2025, herathFirstImpressionsHumanoid2025, kamelabadComparingMonolingualBilingual2025}. Implementations leverage commercial services from OpenAI \cite{aliInclusiveCocreativeChildrobot2025a, zhangBalancingUserControl2025a, xuExploringUseRobots2025a, karliAlchemistLLMAidedEndUser2024a}, Google \cite{antonyXpressSystemDynamic2025a, skantzeApplyingGeneralTurntaking2025, hoSETPAiREdDesigningParental2025, bassiounyUJIButlerSymbolicNonsymbolic2025}, Microsoft \cite{liStargazerInteractiveCamera2023, grassiGroundingConversationalRobots2024, nardelliIntuitiveInteractionCognitive2025}, and Amazon \cite{axelssonYouFollow2023, arjmandEmpathicGroundingExplorations2024}, alongside open-source \cite{laiNaturalMultimodalFusionBased2025} and platform-native engines \cite{wangChildRobotRelationalNorm2025a, rosenPreviousExperienceMatters2024, westerFacingLLMsRobot2024a, sakamotoEffectivenessConversationalRobots2025, blancoAIenhancedSocialRobots2024}. A nascent trend \textbf{explores paralinguistic features} like pitch and speed \cite{bannaWordsIntegratingPersonality2025, axelssonYouFollow2023, dellannaSONARAdaptiveControl2024} and synchronizes speech with physical embodiment \cite{shenSocialRobotsSocial2025, itoRobotDynamicallyAsking2025a}. This allows robots to voice internal states \cite{panACKnowledgeComputationalFramework2025}, selectively address users \cite{grassiStrategiesControllingConversation2025}, or manage turn-taking and recognition errors \cite{skantzeApplyingGeneralTurntaking2025}.

\subsubsection{Visuals}
The visual channel encompasses graphical interfaces, social cues, and environmental perception, with LLMs used to interpret visual input and generate visual output. Within this modality, robots \textbf{present information on screens}, from tablets \cite{sieversIntroducingNoteLevity2024} to dynamic maps with generative visual aids \cite{geGenComUIExploringGenerative2025}. AR interfaces allow users to preview actions and define objects in the workspace \cite{ikedaMARCERMultimodalAugmented2025a, karliAlchemistLLMAidedEndUser2024a}.
 To \textbf{appear more social}, LLMs generate context-aware facial expressions \cite{antonyXpressSystemDynamic2025a, shenSocialRobotsSocial2025} and select gestures \cite{hoSETPAiREdDesigningParental2025}. Abstract cues are also used, such as LED indicators that emulate breathing or manage turn-taking \cite{choARECADesignSpeculation2023, skantzeApplyingGeneralTurntaking2025}.
 Critically, \textbf{VLMs ground language in reality}. They allow robots to interpret scenes through dense captioning \cite{grassiGroundingConversationalRobots2024, panACKnowledgeComputationalFramework2025} or understand human pose to provide context-aware feedback \cite{wangPepperPoseFullbodyPose2024}.

\subsubsection{Motion}

In physical interactions, motion is a primary modality where LLMs typically manage high-level task planning by translating abstract commands into action sequences, while low-level control relies on traditional robotics techniques. The application of motion can be broadly categorized into functional and expressive purposes.
\textbf{Functional motion} involves goal-oriented physical actions to complete tasks. Examples include complex manipulation like pick-and-place, pouring, or grasping \cite{jinRobotGPTRobotManipulation2024, kontogiorgosQuestioningRobotUsing2025, laiNaturalMultimodalFusionBased2025, leusmannInvestigatingLLMDrivenCuriosity2025}, navigation to guide users or follow paths \cite{reimannWhatCanYou2025a, starkDobbyConversationalService2024, zuLanguageSketchingLLMdriven2024}, and performing assembly procedures \cite{pinedaSeeYouLater2025a}.
In parallel, \textbf{expressive motion} serves a communicative or social function to enhance interactional fidelity. It encompasses various forms of gestures \cite{kimUnderstandingLargelanguageModel2024d, sieversIntroducingNoteLevity2024, itoRobotDynamicallyAsking2025a}, platform-specific expressive movements \cite{aliInclusiveCocreativeChildrobot2025a}, and physical changes to convey system status, such as an air purifier's operational intensity \cite{choLivingAlongsideAreca2025}.

\subsubsection{Hybrid}
Most sophisticated HRI systems are inherently multimodal, combining language with other channels to create more robust and intuitive user experiences \cite{elfleetInvestigatingImpactMultimodal2024, leusmannInvestigatingLLMDrivenCuriosity2025}. The effective orchestration of these channels is a key characteristic of advanced systems.
 A primary approach \textbf{synchronizes text-to-speech with non-verbal behaviors}. This includes generating corresponding facial expressions \cite{pereiraMultimodalUserEnjoyment2024a, antonyXpressSystemDynamic2025a, elgarfCreativeBotCreativeStoryteller2022}, body motion, and gaze \cite{pinedaSeeYouLater2025a, wangChallengesAdoptingCompanion2025} to enhance social presence and convey intent \cite{farooqDAIMHRINewHumanRobot2024, sieversInteractingSentimentalRobot2024, salemComparativeHumanrobotInteraction2024, hsuResearchCareReflection2025, wangChildRobotRelationalNorm2025a}.
 Beyond expression, systems \textbf{fuse language with other modalities} for enhanced capability. This includes processing parallel voice commands with deictic gestures to resolve ambiguity \cite{laiNaturalMultimodalFusionBased2025}, combining speech with sketches and sensor data for spatial tasks \cite{zuLanguageSketchingLLMdriven2024, karliAlchemistLLMAidedEndUser2024a}, or linking dialogue to physical actions like navigation \cite{starkDobbyConversationalService2024, bastinGPTAllySafetyorientedSystem2025, latifPhysicsAssistantLLMpoweredInteractive2024}. Some also integrate traditional interfaces like tablets \cite{kamelabadComparingMonolingualBilingual2025}.

\subsubsection{Tangible and Haptic Interaction} This modality ranges from \textbf{direct physical guidance}, where a user manually moves a robot's arm to demonstrate a task \cite{bassiounyUJIButlerSymbolicNonsymbolic2025}, to \textbf{touch-based inputs on screens or sensitive surfaces} \cite{elgarfCreativeBotCreativeStoryteller2022, wangChallengesAdoptingCompanion2025}. Advanced applications of touch aim to simulate the experience of interacting with living entities, moving beyond simple functional feedback toward a more social, life-like feel \cite{choARECADesignSpeculation2023, choLivingAlongsideAreca2025}.

\subsubsection{Proximity}
The \textbf{spatial relationship} between a user and a robot is a subtle yet powerful social cue. Robots can manage conversational dynamics by adjusting their distance to users \cite{grassiStrategiesControllingConversation2025}, and proximity can be explicitly modeled to interpret social situations \cite{pinedaSeeYouLater2025a}. The fundamental importance of this channel is underscored by the many systems designed for co-located, face-to-face interaction within a shared workspace \cite{farooqDAIMHRINewHumanRobot2024, zhangLargeLanguageModels2023b, laiNaturalMultimodalFusionBased2025, sieversIntroducingNoteLevity2024}.

\subsection{Morphology}

\begin{figure}[h]
    \centering
    \includegraphics[width= 1\linewidth]{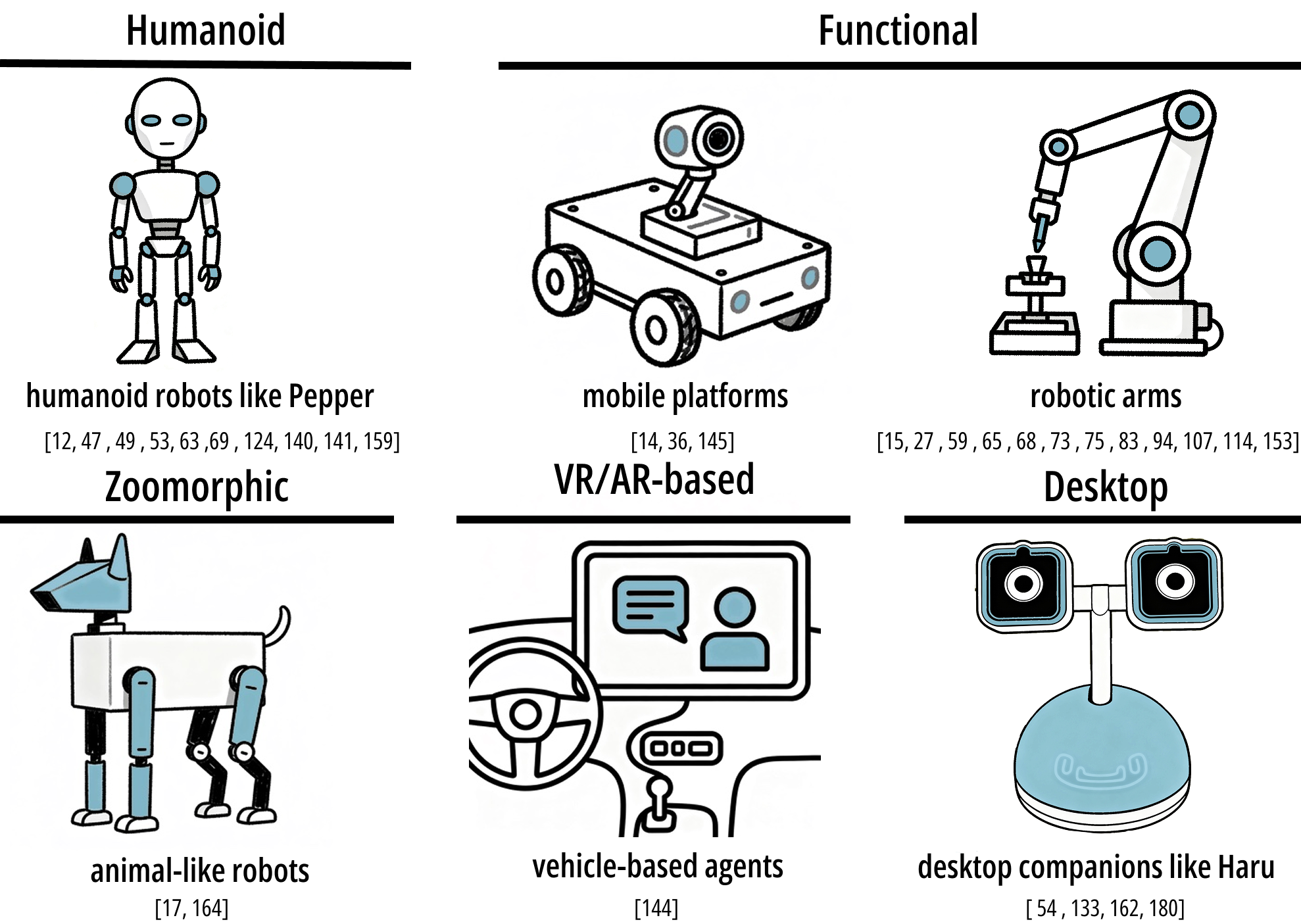}
    \caption{Different types of physical embodiment of a robot.}
    \label{fig:morphology}
    \Description{This diagram classifies the morphological types of robots in LLM-driven human-robot interaction (HRI) systems (under the “Design Components and Strategies” dimension of the 86 reviewed studies), with six core categories paired with descriptive definitions, iconographic representations, and corresponding literature references (numerical brackets): (1) Humanoid: Refers to humanoid robots (e.g., Pepper), illustrated by a humanoid robot figure, supported by studies [12, 47, 49, 53, 63, 69, 124, 140, 141, 159]. (2) Functional: Encompasses task-oriented robots, including mobile platforms (illustrated by a wheeled robotic platform, references [14, 36, 145]) and robotic arms (illustrated by an industrial robotic arm, references [15, 27, 59, 65, 68, 73, 75, 83, 94, 107, 114, 153]). (3) Zoomorphic: Denotes animal-like robots, illustrated by a dog-shaped robotic figure, associated with studies [17, 164]. (4) VR/AR-based: Refers to vehicle-integrated intelligent agents (linked to VR/AR interfaces), illustrated by a vehicle-mounted agent interface, supported by study [144]. (5) Desktop: Encompasses desktop companion robots (e.g., Haru), illustrated by a dual-camera desktop robotic device, with references [54, 133, 162, 180].}
\end{figure}

Once the interaction channels are established, researchers further consider the physical morphology that serves as the carrier for these modalities. The physical embodiment of a robot represents a critical design consideration that can fundamentally shape the nature of interaction~\cite{fongSurveySociallyInteractive2003a}. 
Figure \ref{fig:morphology} provides a survey of the diverse platforms utilized in the reviewed literature to support various interaction goals.

\subsubsection{Humanoid}
 Humanoid robots (e.g., Pepper~\cite{kimUnderstandingLargelanguageModel2024d,sieversInteractingSentimentalRobot2024,sieversIntroducingNoteLevity2024,itoRobotDynamicallyAsking2025a,wangPepperPoseFullbodyPose2024,grassiGroundingConversationalRobots2024,bannaWordsIntegratingPersonality2025,grassiStrategiesControllingConversation2025,rosenPreviousExperienceMatters2024,herathFirstImpressionsHumanoid2025}, Nao~\cite{salemComparativeHumanrobotInteraction2024,zhangExploringRobotPersonality2025a,dellannaSONARAdaptiveControl2024,grassiEnhancingLLMBasedHumanRobot2024}, Furhat~\cite{pereiraMultimodalUserEnjoyment2024a,kamelabadComparingMonolingualBilingual2025,salemComparativeHumanrobotInteraction2024,axelssonYouFollow2023,arjmandEmpathicGroundingExplorations2024,skantzeApplyingGeneralTurntaking2025,elgarfCreativeBotCreativeStoryteller2022}) remain the most prevalent due to their suitability for socially aligned interactions, with additional platforms such as QTrobot~\cite{hsuBittersweetSnapshotsLife2025a,spitaleVITAMultiModalLLMBased2025a}, ARI~\cite{reimannWhatCanYou2025a}, Navel~\cite{nardelliIntuitiveInteractionCognitive2025}, Geminoid F~\cite{sakamotoEffectivenessConversationalRobots2025}, and Mobi~\cite{loLLMbasedRobotPersonality2025a} are also frequently employed. 

\subsubsection{Functional}
These robots are designed primarily for task performance rather than social embodiment. This group including mobile platforms (e.g., TurtleBot~\cite{farooqDAIMHRINewHumanRobot2024,bassiounyUJIButlerSymbolicNonsymbolic2025}, Segway~\cite{starkDobbyConversationalService2024}) and robotic arms~\cite{kodurExploringDynamicsHumanRobot2025,jinRobotGPTRobotManipulation2024,pinedaSeeYouLater2025a,bastinGPTAllySafetyorientedSystem2025,karliAlchemistLLMAidedEndUser2024a,ikedaMARCERMultimodalAugmented2025a,mahadevanImageInThatManipulatingImages2025,liStargazerInteractiveCamera2023,taoLAMSLLMDrivenAutomatic2025a,kontogiorgosQuestioningRobotUsing2025,cuiNoRightOnline2023a,padmanabhaVoicePilotHarnessingLlms2024} dominate task-oriented applications such as manipulation.

\subsubsection{Other}

Some studies explore alternative morphologies. These include zoomorphic (animal-like) robots~\cite{wangCrowdBotOpenenvironmentRobot2024,blancoAIenhancedSocialRobots2024}, desktop companions like Haru~\cite{shenSocialRobotsSocial2025,hoSETPAiREdDesigningParental2025,wangChallengesAdoptingCompanion2025,yuImprovingPerceivedEmotional2024}, and even vehicle-based agents for human-vehicle interaction research~\cite{stampfExploringPassengerAutomatedVehicle2024}. This diversity highlights the expanding design space for LLM-driven robots beyond traditional forms.

\subsection{Levels of Autonomy}
\begin{figure}[h]
    \centering
    \includegraphics[width= 1\linewidth]{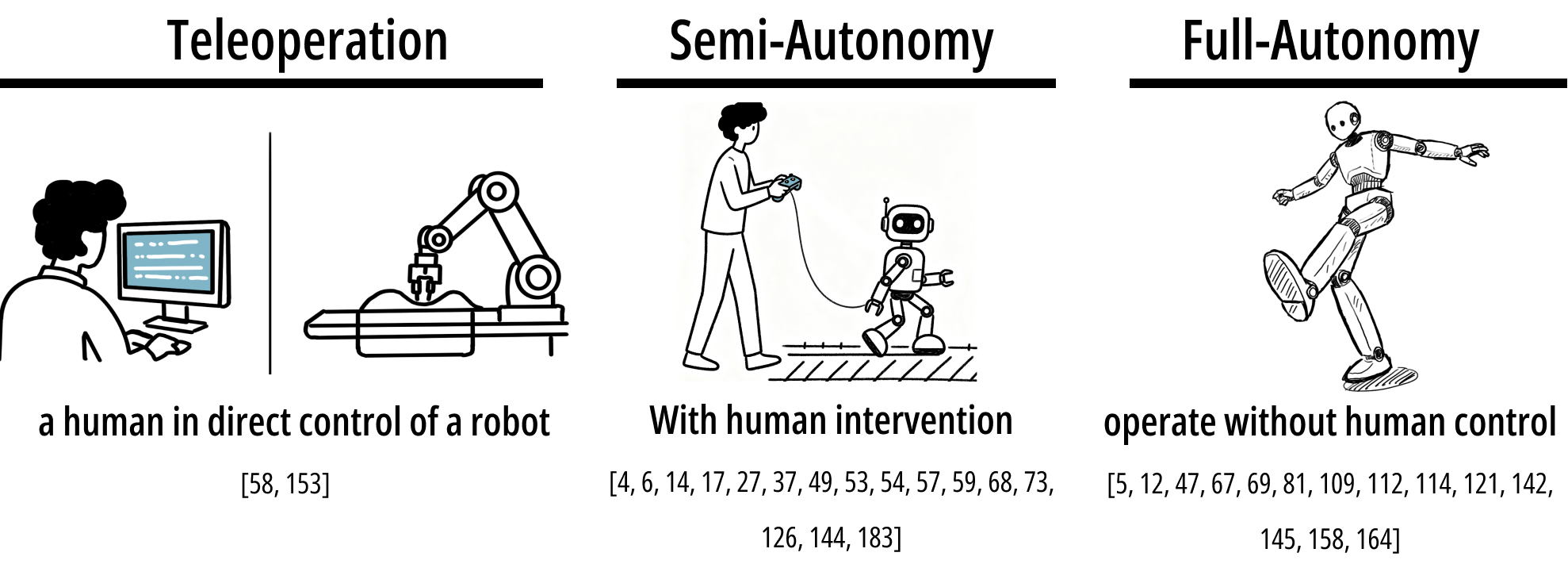}
    \caption{The increasing autonomy level from direct control to deciding everything and acting.}
    \label{fig:autonomy}
    \Description{This diagram categorizes the autonomy spectrum of LLM-driven human-robot interaction (HRI) systems (under the “Design Components and Strategies” dimension of the reviewed 86 studies), with three core levels paired with descriptive definitions, iconographic scenarios, and corresponding literature references (numerical brackets): (1) Teleoperation: Defined as “a human in direct control of a robot” (illustrated by a human operating a robotic arm via a computer interface), supported by studies [58, 153]. (2) Semi-Autonomy: Characterized by robot operation “with human intervention” (illustrated by a human guiding a mobile robot via a controller), associated with studies [4, 6, 14, 17, 22, 37, 49, 53, 54, 57, 59, 68, 73, 126, 144, 183]. (3) Full-Autonomy: Refers to robot operation “without human control” (illustrated by an independent humanoid robot), supported by studies [5, 12, 47, 69, 81, 109, 112, 114, 142, 145, 158, 164]. This taxonomy captures the human-robot control relationships enabled by LLMs, as identified in this systematic review.}
\end{figure}

Moving from external form to internal logic, the design process further involves determining the appropriate level of autonomy. This dimension describes the extent to which a robot acts on its own accord and defines the distribution of control between the user and the LLM. It is important 
the optimal degree of independence often depends on the specific research context, the target application, and the nature of the task. Figure \ref{fig:autonomy} illustrates the spectrum of autonomy, from direct control to more independent decision-making and execution.

\subsubsection{Teleoperation}
Teleoperation places a human in direct control of a robot, a method used to study user experience when autonomy is not yet feasible. In an assistive teleoperation model, LLMs can translate an operator's high-level commands into low-level robot actions to reduce cognitive load \cite{taoLAMSLLMDrivenAutomatic2025a}. Even in these systems, human verification is common; for instance, an experimenter might approve LLM-generated responses before the robot speaks to ensure safety and appropriateness \cite{huDesigningTelepresenceRobots2025}. Additionally, teleoperation serves as a relatively important source for collecting data for robot imitation learning.

\subsubsection{Semi-Autonomy}
Semi-autonomous control serves as a practical strategy to mitigate system failures, including conversational breakdowns \cite{sakamotoEffectivenessConversationalRobots2025, grassiStrategiesControllingConversation2025}, inappropriate LLM responses \cite{herathFirstImpressionsHumanoid2025}, and physical safety risks \cite{kodurExploringDynamicsHumanRobot2025}, acknowledging the continued need for human oversight when implicit cues are missed \cite{hsuBittersweetSnapshotsLife2025a}. This approach is applied in diverse contexts: collaborative tasks \cite{karliAlchemistLLMAidedEndUser2024a, ikedaMARCERMultimodalAugmented2025a}, creative content co-creation \cite{aliInclusiveCocreativeChildrobot2025a, blancoAIenhancedSocialRobots2024, liStargazerInteractiveCamera2023}, and managing social dynamics like turn-taking and gaze \cite{sakamotoEffectivenessConversationalRobots2025, ferriniPerceptsSemanticsMultimodala, hoSETPAiREdDesigningParental2025}. Implementations typically involve a human-in-the-loop, sometimes via a Wizard-of-Oz setup \cite{sakamotoEffectivenessConversationalRobots2025, arjmandEmpathicGroundingExplorations2024}. Other methods include direct human intervention to advance tasks \cite{zhangBalancingUserControl2025a}, provide online corrections \cite{cuiNoRightOnline2023a, stampfExploringPassengerAutomatedVehicle2024}, or curate AI-generated content \cite{aliInclusiveCocreativeChildrobot2025a} within a human-in-the-loop learning framework \cite{bassiounyUJIButlerSymbolicNonsymbolic2025, blancoAIenhancedSocialRobots2024}.

\subsubsection{Full Autonomy}
Full autonomy aims to create robots that operate without direct human control \cite{kimUnderstandingLargelanguageModel2024d}, using LLMs to build end-to-end systems for complex social interactions \cite{pinedaSeeYouLater2025a, starkDobbyConversationalService2024}. The objective is to produce socially aware partners \cite{grassiGroundingConversationalRobots2024} that generate their own real-time dialogue and behaviors, allowing researchers to study emergent user interactions \cite{reimannWhatCanYou2025a, pereiraMultimodalUserEnjoyment2024a}. To enhance believability, these systems generate dynamic facial expressions \cite{antonyXpressSystemDynamic2025a} and context-sensitive gestures \cite{bannaWordsIntegratingPersonality2025}. For complex tasks, they integrate multiple data sources using frameworks that combine LLMs with VLMs \cite{leusmannInvestigatingLLMDrivenCuriosity2025} and knowledge graphs \cite{panACKnowledgeComputationalFramework2025}. Some systems even use reinforcement learning for dynamic adaptation \cite{wangCrowdBotOpenenvironmentRobot2024}. Implementations are often hybrid architectures augmenting a central LLM with specialized components like computer vision for body tracking \cite{wangChildRobotRelationalNorm2025a}, models for turn-taking prediction \cite{skantzeApplyingGeneralTurntaking2025}, and standard speech I/O \cite{kamelabadComparingMonolingualBilingual2025}.

\section{Study Methods and Evaluation Strategies}
\label{sec: method}
\begin{figure*}[h]
    \centering
    \includegraphics[width= 1\linewidth]{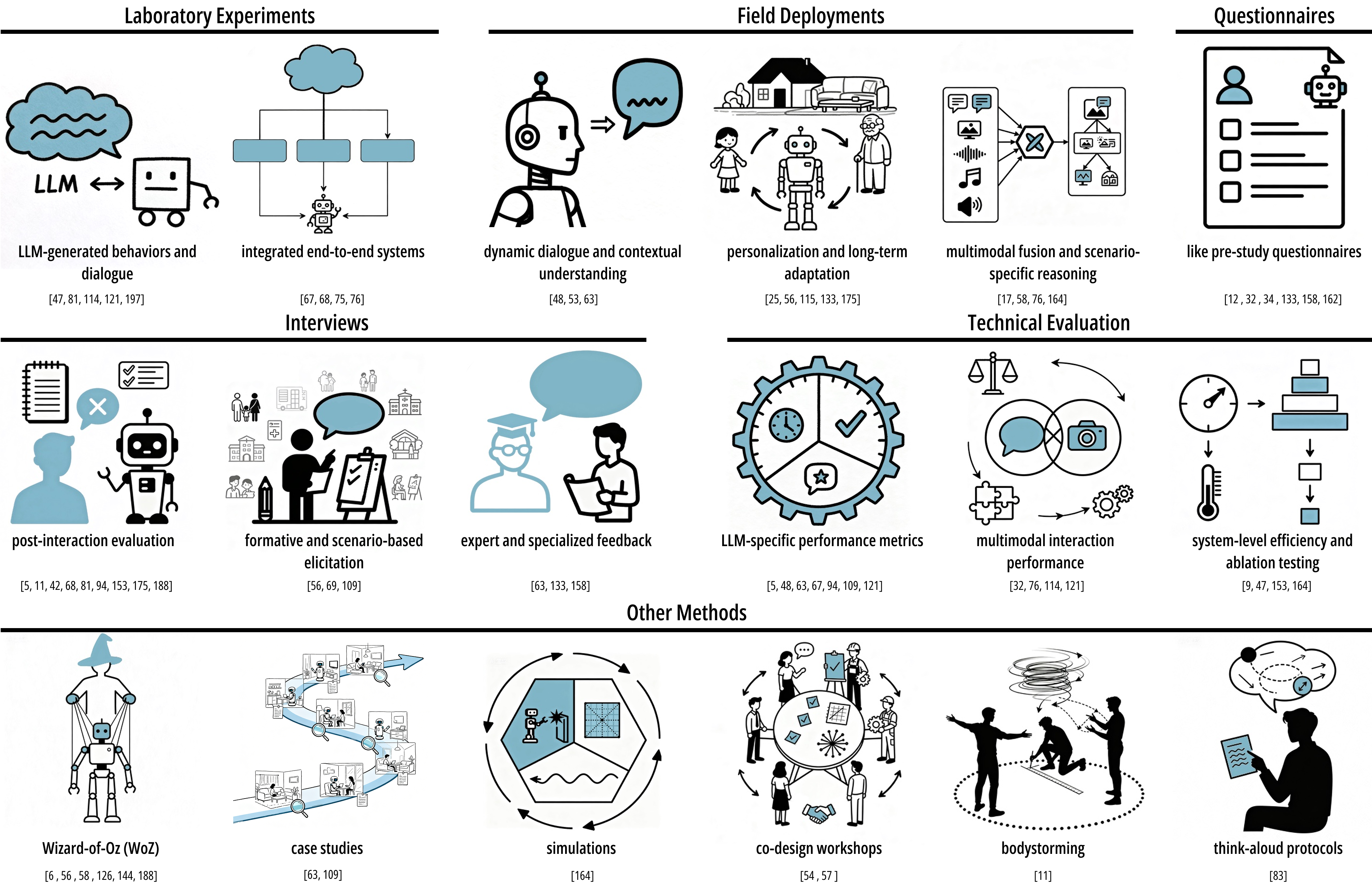}
    \caption{Study methods used in HRI.}
    \Description{This figure presents the classification of study methods (tailored to human-robot interaction (HRI) research scenarios) employed in LLM-driven HRI studies, derived from the 86 papers included in this systematic review. These methods cover core HRI research dimensions (controlled environment testing, real-world deployment, user-centric feedback, system performance verification, etc.), organized into six categories with scenario-specific implementations, iconographic illustrations, and corresponding literature references (numerical brackets): (1) Laboratory Experiments: As a foundational method in HRI research, this category includes “LLM-generated behaviors and dialogue” (used to test robot interaction behaviors driven by LLMs in controlled settings, illustrated by LLM-robot dialogue interaction, references [47, 81, 121, 197]) and “integrated end-to-end systems” (used to verify the performance of full LLM-robot interaction systems, illustrated by system architecture, references [67, 68, 75, 76]). (2) Field Deployments: Applied to explore HRI in natural contexts, this category covers “dynamic dialogue and contextual understanding” (used to test robot dialogue adaptability in real interactions, illustrated by robot-human dialogue, references [48, 53, 63]), “personalization and long-term adaptation” (used to study robot-user long-term interaction in daily scenarios, illustrated by multi-user domestic interaction, references [25, 56, 115, 133, 175]), and “multimodal fusion and scenario-specific reasoning” (used to verify cross-modality interaction effects in specific HRI scenarios, illustrated by cross-modality processing, references [17, 38, 164]). (3) Questionnaires: A common user-centric method in HRI, this category refers to “pre-study questionnaires” (used to collect user background or pre-interaction preferences before HRI experiments, illustrated by survey forms, references [12, 33, 34, 133, 158, 162]). (4) Interviews: Used to obtain in-depth user perceptions of HRI, this category includes “post-interaction evaluation” (used to collect user feedback after single HRI sessions, illustrated by post-interaction feedback, references [5, 11, 42, 68, 81, 155, 175, 188]), “formative and scenario-based elicitation” (used to gather user demands for HRI scenario design, illustrated by group scenario discussion, references [56, 69, 109]), and “expert and specialized feedback” (used to obtain professional evaluations of HRI system design, illustrated by expert interviews, references [63, 133, 158]). (5) Technical Evaluation (HRI System Performance Verification): Used to measure the technical effectiveness of LLM-driven HRI systems, this category covers “LLM-specific performance metrics” (used to evaluate LLM’s interaction-related performance in HRI, illustrated by metric visualization, references [5, 48, 63, 67, 94, 109, 121]), “multimodal interaction” (used to verify the coordination effect of multi-modal signals in HRI, illustrated by cross-modality alignment), and “system-level efficiency and ablation testing” (used to assess HRI system operation efficiency and component contributions, illustrated by efficiency assessment, references [32, 76, 114, 121]). (6) Other Methods: Encompasses methods tailored to specific HRI research needs, including “Wizard-of-Oz (WoZ)” (used to simulate LLM-driven robot behaviors in early HRI prototype testing, illustrated by remote-controlled robot interaction, references [6, 56, 58, 126, 144, 188]), “case studies” (used to analyze typical LLM-driven HRI application cases, references [53, 109]), “simulations” (used to predict HRI effects in virtual scenarios, references [164]), “co-design workshops” (used to co-develop HRI solutions with users, references [54, 57]), “bodystorming” (used to explore physical interaction designs in HRI, references [11]), and “think-aloud protocols” (used to record user cognitive processes during HRI, references [83]).
}
    \label{fig:studymthod}
\end{figure*}

\begin{figure*}[h]
    \centering
    \includegraphics[width= 1\linewidth]{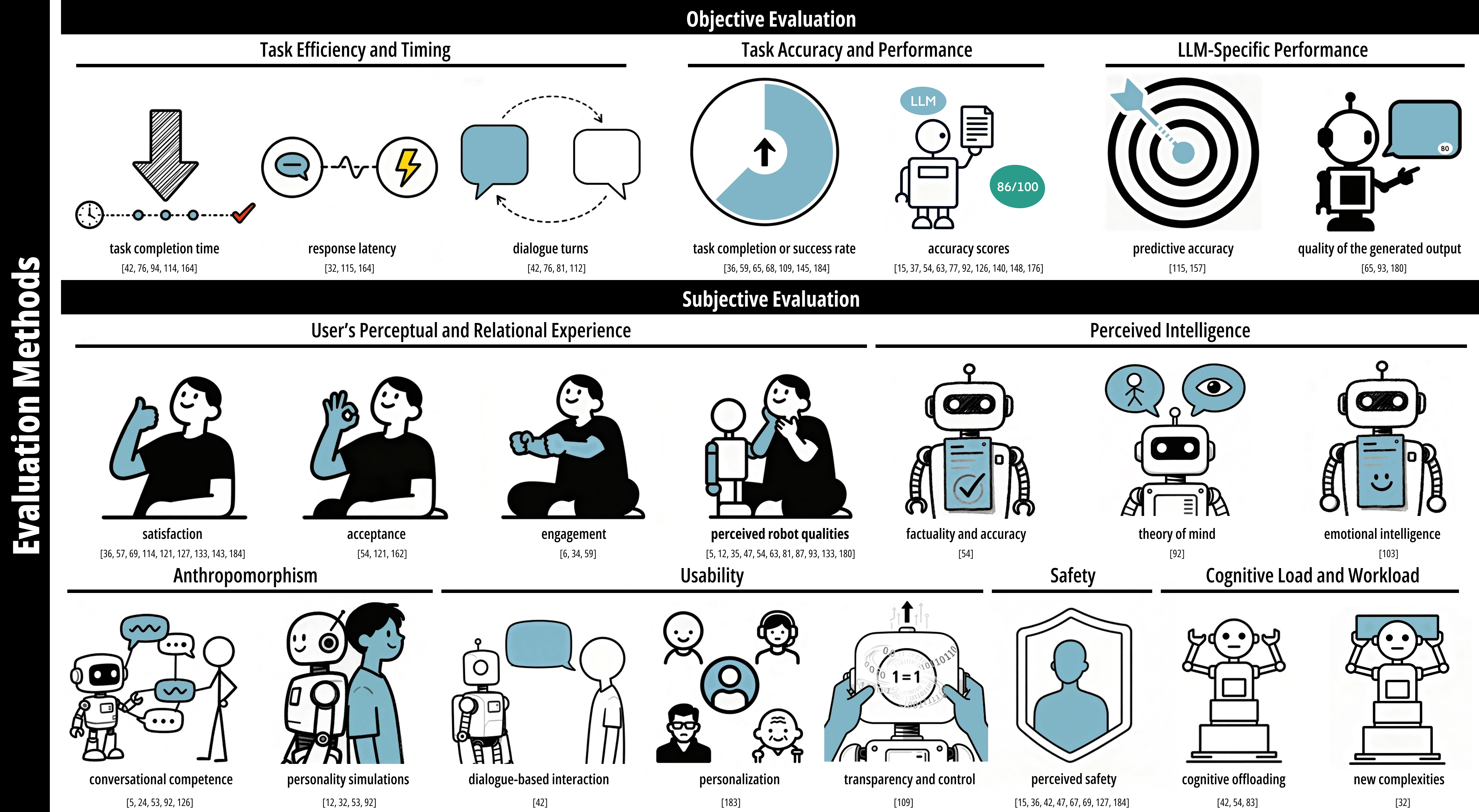}
    \caption{Evaluation methods used in HRI.}
    \Description{This figure classifies evaluation methods in LLM-driven human-robot interaction (HRI) research (based on 86 reviewed studies), dividing approaches into Objective Evaluation (quantitative system performance) and Subjective Evaluation (user perceptual experience), with each subdimension linked to specific metrics, iconographic scenarios, and literature references (numerical brackets): 1. Objective Evaluation: (1) Task Efficiency and Timing: Task completion time [32,76,114,164] (icon: timeline + task completion badge); Response latency [32,115,164] (icon: dialogue + latency waveform); Dialogue turns [47,85,112] (icon: circular dialogue bubbles). (2) Task Accuracy and Performance: Task completion/success rate [36,68,98,109,145,184] (icon: rising proportion chart); Accuracy scores [13,36,47,68,77,126,140,148,178] (icon: robot with scorecard). (3) LLM-Specific Performance: Predictive accuracy [115,157] (icon: target hit); Generated output quality [65,93,180] (icon: robot outputting content). 2. Subjective Evaluation: (1) User's Perceptual & Relational Experience: Satisfaction [36,57,64,114,121,127,133,145,184] (icon: user thumbs-up); Acceptance [34,121,162] (icon: user OK gesture); Engagement [36,114,159] (icon: interactive user pose); Perceived robot qualities [5,12,35,47,64,83,87,133,153,180] (icon: user reflecting on robot). (2) Perceived Intelligence: Factuality/accuracy [54] (icon: robot with verification badge); Theory of mind [92] (icon: robot linking user intent); Emotional intelligence [103] (icon: robot with emotion badge). (3) Anthropomorphism: Conversational competence [5,24,33,82,180] (icon: multi-turn robot dialogue); Personality simulations [12,22,32,93] (icon: user + anthropomorphic robot); Dialogue-based interaction [42] (icon: robot-user dialogue). (4) Usability: Personalization [183] (icon: multi-user personalized labels); Transparency/control [109] (icon: user operating robot interface). (5) Safety & Cognitive Load: Perceived safety [13,36,47,62,67,127,137,184] (icon: shield + user); Cognitive offloading [42,54,83] (icon: robot reducing cognitive load); New complexities [42] (icon: robot increasing cognitive load).}
    \label{fig:evaluation}
\end{figure*}

Regarding \textbf{RQ3}, this section systematically reviews the study methodologies in Figure~\ref{fig:studymthod} and evaluation strategies for HRI in the age of LLMs in Figure~\ref{fig:evaluation}. Distinct from purely computational benchmarks, HRI research prioritizes human-centered evaluation, predominantly through user studies.

\subsection{Methodology}

\subsubsection{Laboratory Experiment}
Laboratory experiments remain a cornerstone for empirically evaluating LLM-driven HRI in controlled settings. These studies are essential for isolating the impact of LLM-driven capabilities on interaction quality and user perception. Researchers typically evaluate both \textbf{LLM-generated behaviors and dialogue} and \textbf{integrated end-to-end systems}. This involves using structured, task-based scenarios—such as collaborative manipulation \cite{kontogiorgosQuestioningRobotUsing2025}, language learning \cite{kamelabadComparingMonolingualBilingual2025}, robot programming \cite{karliAlchemistLLMAidedEndUser2024a}, or assistive tasks \cite{laiNaturalMultimodalFusionBased2025}—to gather performance metrics and subjective feedback. Dedicated labs allow for precise control over variables, facilitating high-quality data collection on how LLM-driven personalities \cite{pinedaSeeYouLater2025a}, curiosity \cite{leusmannInvestigatingLLMDrivenCuriosity2025}, or multimodal reasoning \cite{reimannWhatCanYou2025a, grassiGroundingConversationalRobots2024, zuLanguageSketchingLLMdriven2024} affect user experience and task success.

\subsubsection{Field Deployments}
\label{subsec:field}
Field deployments test the robustness and adaptability of LLM-driven robots in real-world environments, moving beyond the constraints of the lab. These studies are crucial for understanding how systems perform over extended periods and with diverse user populations. Key applications include leveraging LLMs for \textbf{dynamic dialogue and contextual understanding} in unpredictable settings like classrooms \cite{itoRobotDynamicallyAsking2025a} or public festivals \cite{herathFirstImpressionsHumanoid2025, grassiEnhancingLLMBasedHumanRobot2024}. Another major focus is on \textbf{personalization and long-term adaptation}, particularly in home environments where robots support child wellbeing, act as companions \cite{xuExploringUseRobots2025a, shenSocialRobotsSocial2025, choLivingAlongsideAreca2025}, or assist older adults \cite{pintoPredictiveTurntakingLeveraging2024, hsuResearchCareReflection2025}. Finally, deployments often evaluate \textbf{multimodal fusion and scenario-specific reasoning} in complex settings like care centers \cite{laiNaturalMultimodalFusionBased2025, blancoAIenhancedSocialRobots2024} and university campuses \cite{wangCrowdBotOpenenvironmentRobot2024, huDesigningTelepresenceRobots2025}.

\subsubsection{Interviews}
Interviews are a key qualitative method for capturing nuanced human perceptions and expectations regarding LLM-driven robots. They are often used for \textbf{post-interaction evaluation}, like semi-structured formats allow researchers to gather detailed feedback on user experiences, preferences, and the reasoning behind them \cite{leusmannInvestigatingLLMDrivenCuriosity2025, karliAlchemistLLMAidedEndUser2024a, antonyXpressSystemDynamic2025a, axelssonOhSorryThink2024, xuExploringUseRobots2025a, geGenComUIExploringGenerative2025, zhangPromptingEmbodiedAI2025, mahadevanImageInThatManipulatingImages2025, taoLAMSLLMDrivenAutomatic2025a}. Interviews are also vital in \textbf{formative and scenario-based elicitation} during the early design stages, helping to align system concepts with user needs before implementation \cite{panACKnowledgeComputationalFramework2025, hsuResearchCareReflection2025, kimUnderstandingLargelanguageModel2024d}. Furthermore, they are used to gather \textbf{expert and specialized feedback} from researchers or domain specialists on the broader implications of LLM-driven robotic systems \cite{itoRobotDynamicallyAsking2025a, wangChildRobotRelationalNorm2025a, shenSocialRobotsSocial2025}.

\subsubsection{Questionnaires}
Questionnaires are a versatile tool for collecting quantitative, self-reported data at various stages of HRI research. \textbf{Pre-study questionnaires} gather demographic data and assess participants' prior experience or expectations with AI and robots \cite{wangChallengesAdoptingCompanion2025, elfleetInvestigatingImpactMultimodal2024, shenSocialRobotsSocial2025, bannaWordsIntegratingPersonality2025, dellannaSONARAdaptiveControl2024, wangChildRobotRelationalNorm2025a}. Their most common application is in \textbf{post-study questionnaires}, which measure subjective metrics like user enjoyment, perceived intelligence, and usability, capturing dimensions specific to LLM capabilities \cite{reimannWhatCanYou2025a, pereiraMultimodalUserEnjoyment2024a, karliAlchemistLLMAidedEndUser2024a, ikedaMARCERMultimodalAugmented2025a, kimUnderstandingLargelanguageModel2024d, grassiGroundingConversationalRobots2024, elgarfCreativeBotCreativeStoryteller2022, zhangBalancingUserControl2025a, spitaleVITAMultiModalLLMBased2025a, axelssonOhSorryThink2024}. Questionnaires also function as a \textbf{standalone research tool} for scalable data collection, such as gathering task instructions to fine-tune models \cite{wangCrowdBotOpenenvironmentRobot2024, loLLMbasedRobotPersonality2025a} or crowdsourcing evaluations of LLM-generated dialogue scripts \cite{mannavaExploringSuitabilityConversational2024}.

\subsubsection{Technical Evaluation}
Technical evaluations of LLM-driven HRI systems assess functional correctness, fluency, and efficiency. This goes beyond traditional robotics to include \textbf{LLM-specific performance metrics} such as response latency, code execution success rates, and the semantic quality of generated language \cite{grassiEnhancingLLMBasedHumanRobot2024, kamelabadComparingMonolingualBilingual2025, panACKnowledgeComputationalFramework2025, antonyXpressSystemDynamic2025a, mahadevanImageInThatManipulatingImages2025, itoRobotDynamicallyAsking2025a, reimannWhatCanYou2025a}. Since LLMs often act as a central hub, evaluations must also measure \textbf{multimodal interaction performance}, analyzing the fusion of language with perceptual data and comparing the integrated system against specialized baselines \cite{pinedaSeeYouLater2025a, reimannWhatCanYou2025a, laiNaturalMultimodalFusionBased2025, dellannaSONARAdaptiveControl2024}. Finally, \textbf{system-level efficiency and ablation testing} are used to measure computational overhead and isolate the LLM's specific contribution to performance gains, often validated with statistical tests \cite{wangCrowdBotOpenenvironmentRobot2024, grassiGroundingConversationalRobots2024, taoLAMSLLMDrivenAutomatic2025a, axelssonYouFollow2023}.

\subsubsection{Other Methods}
To address the unique challenges of evaluating LLM-based robots, researchers also employ a range of complementary methods. The \textbf{Wizard-of-Oz} technique remains prevalent for simulating advanced dialogue capabilities and mitigating model unreliability during user studies \cite{sakamotoEffectivenessConversationalRobots2025, arjmandEmpathicGroundingExplorations2024, huDesigningTelepresenceRobots2025, stampfExploringPassengerAutomatedVehicle2024, zhangPromptingEmbodiedAI2025, hsuResearchCareReflection2025}. \textbf{Case studies} and longitudinal deployments offer deep, contextual insights into long-term use \cite{itoRobotDynamicallyAsking2025a, panACKnowledgeComputationalFramework2025}, while \textbf{simulations} enable safe and scalable testing in complex scenarios \cite{wangCrowdBotOpenenvironmentRobot2024}. Lastly, \textbf{participatory methods} such as co-design workshops \cite{hsuBittersweetSnapshotsLife2025a, hoSETPAiREdDesigningParental2025}, bodystorming \cite{axelssonOhSorryThink2024}, and think-aloud protocols \cite{liStargazerInteractiveCamera2023} are critical for gathering formative feedback and aligning system design with user needs.

\subsection{Evaluation Metrics}
\label{section:metrics}
\subsubsection{Objective}
In the age of LLMs, objective evaluation in HRI combines traditional performance metrics with new measures assessing the models themselves. 

- \textit{\textbf{Task Efficiency and Timing.}}
Metrics in this category quantify the speed and smoothness of the interaction. Researchers commonly measure \textbf{task completion time (TCT)} to assess overall efficiency \cite{laiNaturalMultimodalFusionBased2025, geGenComUIExploringGenerative2025, pinedaSeeYouLater2025a, mahadevanImageInThatManipulatingImages2025, wangCrowdBotOpenenvironmentRobot2024}. Another key metric is \textbf{response latency}, which includes both the robot's general response time \cite{dellannaSONARAdaptiveControl2024, wangCrowdBotOpenenvironmentRobot2024} and the specific latency of LLM API calls, a factor critical for real-time turn-taking \cite{pintoPredictiveTurntakingLeveraging2024}. Additionally, studies evaluate the number of \textbf{dialogue turns} or interaction durations to gauge conversational compactness \cite{geGenComUIExploringGenerative2025, pereiraMultimodalUserEnjoyment2024a, leusmannInvestigatingLLMDrivenCuriosity2025}. For instance, LLM integration has demonstrated the potential to reduce interaction time by up to 50\% in certain multimodal tasks \cite{laiNaturalMultimodalFusionBased2025}, highlighting its impact on efficiency.

- \textit{\textbf{Task Accuracy and Performance.}}
This category evaluates the correctness and success of the HRI system. A primary metric is the \textbf{task completion or success rate}, often supplemented with error logs to identify failure modes \cite{starkDobbyConversationalService2024, jinRobotGPTRobotManipulation2024, karliAlchemistLLMAidedEndUser2024a, ikedaMARCERMultimodalAugmented2025a, panACKnowledgeComputationalFramework2025, farooqDAIMHRINewHumanRobot2024, zhangLargeLanguageModels2023b}. Beyond binary success/failure, researchers employ more granular \textbf{accuracy scores} tailored to specific tasks. Examples include accuracy in question generation \cite{itoRobotDynamicallyAsking2025a}, semantic precision in understanding commands \cite{ferriniPerceptsSemanticsMultimodala}, accuracy in estimating user states or preferences \cite{hoSETPAiREdDesigningParental2025, sakamotoEffectivenessConversationalRobots2025}, and performance on benchmark datasets \cite{loLLMbasedRobotPersonality2025a}, with some systems achieving over 90\% accuracy \cite{yanoUnifiedUnderstandingEnvironment2024}. Other works also focus on performance in specialized domains like physics tutoring \cite{latifPhysicsAssistantLLMpoweredInteractive2024} or adaptive interfaces \cite{suChatAdpChatGPTpoweredAdaptation2024, bastinGPTAllySafetyorientedSystem2025, sieversInteractingSentimentalRobot2024}.

- \textit{\textbf{LLM-Specific Performance.}}
With LLMs as a core component, evaluation extends to the model's intrinsic performance. This involves measuring the LLM's \textbf{predictive accuracy} using standard machine learning metrics like recall, precision, and F1-scores for tasks such as turn-taking prediction \cite{pintoPredictiveTurntakingLeveraging2024} or theory of mind assessments \cite{vermaTheoryMindAbilities2024}. The \textbf{quality of the generated output} is another crucial aspect, evaluated through metrics like Pylint scores for code generation \cite{jinRobotGPTRobotManipulation2024}, similarity to expert-designed behaviors \cite{mahadevanGenerativeExpressiveRobot2024}, or accuracy in matching a target voice profile. These metrics provide direct insights into the LLM's reliability and capability, directly measuring its contribution to the overall system performance \cite{yuImprovingPerceivedEmotional2024}.

\subsubsection{Subjective}
The integration of LLMs has profoundly reshaped the evaluation of subjective user experience in HRI. While established metrics for constructs like usability, safety, and perceived intelligence remain vital, they are now augmented by new dimensions that capture the complexities of LLM-driven interaction, such as conversational depth and relational quality.

- \textit{\textbf{User's Perceptual and Relational Experience.}}
The integration of LLMs has pivoted subjective HRI evaluation from static, task-oriented metrics toward a holistic assessment of the user's perceptual and relational experience. While foundational metrics like the Godspeed Questionnaire's \textit{Likeability} subscale and general user \textbf{satisfaction} ratings remain prevalent \cite{kimUnderstandingLargelanguageModel2024d, shenSocialRobotsSocial2025, farooqDAIMHRINewHumanRobot2024, zhangLargeLanguageModels2023b, salemComparativeHumanrobotInteraction2024}, their scope has expanded. Satisfaction is now also judged by social dialogue quality \cite{pinedaSeeYouLater2025a, spitaleVITAMultiModalLLMBased2025a}, emotional safety \cite{hsuBittersweetSnapshotsLife2025a}, and enjoyment from long-term interactions \cite{reimannWhatCanYou2025a}. Similarly, \textbf{acceptance} now extends beyond reuse intention \cite{reimannWhatCanYou2025a} to include attitudes toward AI-generated content \cite{hoSETPAiREdDesigningParental2025} and concerns about transparency and deception \cite{wangChallengesAdoptingCompanion2025}. LLM-driven fluency also deepens \textbf{engagement}, transforming it into a measure of partnership and interaction quality \cite{ikedaMARCERMultimodalAugmented2025a}. This is often evaluated through the robot’s ability to generate empathetic reactions that encourage user expression \cite{arjmandEmpathicGroundingExplorations2024, antonyXpressSystemDynamic2025a, elfleetInvestigatingImpactMultimodal2024}. Finally, researchers increasingly assess \textbf{perceived robot qualities} like intelligence \cite{grassiGroundingConversationalRobots2024}, empathy \cite{yuImprovingPerceivedEmotional2024}, curiosity \cite{leusmannInvestigatingLLMDrivenCuriosity2025}, and creativity \cite{elgarfCreativeBotCreativeStoryteller2022}, focusing on the alignment between generated content and its physical, perceivable, and low-latency enactment \cite{antonyXpressSystemDynamic2025a, mahadevanGenerativeExpressiveRobot2024, bannaWordsIntegratingPersonality2025, itoRobotDynamicallyAsking2025a}. Longitudinal and multimodal methods are becoming essential to capture these dynamic aspects \cite{limaPromotingCognitiveHealth2025, yuImprovingPerceivedEmotional2024, shenSocialRobotsSocial2025, hoSETPAiREdDesigningParental2025}.

- \textit{\textbf{Perceived Intelligence.}}
Perceived intelligence is a critical subjective metric in the LLM era, reflecting users' judgment of a robot's cognitive abilities. While traditional measures like the Godspeed questionnaire are still widely used to assess dimensions like competence and knowledge \cite{zhangBalancingUserControl2025a, geGenComUIExploringGenerative2025, grassiGroundingConversationalRobots2024}, LLMs introduce novel evaluative dimensions. Key among these are concerns about \textbf{factuality and accuracy}, as LLM hallucinations can negatively impact perceptions of intelligence \cite{hoSETPAiREdDesigningParental2025}. Conversely, LLMs enable advanced social-cognitive abilities, such as \textbf{theory of mind} \cite{loLLMbasedRobotPersonality2025a} and \textbf{emotional intelligence} \cite{nardelliIntuitiveInteractionCognitive2025}, which expand the construct of perceived intelligence beyond task competence. Studies consistently show that the enhanced dialogue quality and responsiveness of LLM-powered robots lead to higher ratings in perceived intelligence \cite{huDesigningTelepresenceRobots2025, sakamotoEffectivenessConversationalRobots2025}, an effect often underscored by direct user feedback such as ``You're smart''\cite{itoRobotDynamicallyAsking2025a}.

- \textit{\textbf{Anthropomorphism.}}
The integration of LLMs has significantly reshaped the evaluation of anthropomorphism in HRI. Foundational dimensions like animacy, intelligence, and likeability, often assessed with the Godspeed questionnaire, remain relevant \cite{kimUnderstandingLargelanguageModel2024d, shenSocialRobotsSocial2025, salemComparativeHumanrobotInteraction2024, sieversInteractingSentimentalRobot2024, geGenComUIExploringGenerative2025}. However, LLMs introduce more nuanced facets of human-likeness. Evaluation has expanded to include the robot's \textbf{conversational competence}, such as its ability to perform role-taking, maintain a consistent personality, and articulate values \cite{herathFirstImpressionsHumanoid2025, sakamotoEffectivenessConversationalRobots2025, loLLMbasedRobotPersonality2025a}. The perceived human-likeness is now deeply tied to the quality of dialogue, with compelling storytelling and personal reflections blurring the machine-human boundary \cite{antonyXpressSystemDynamic2025a, choARECADesignSpeculation2023}. Furthermore, LLMs facilitate more complex and adaptive \textbf{personality simulations}, enhancing relatability \cite{loLLMbasedRobotPersonality2025a, bannaWordsIntegratingPersonality2025}. Nevertheless, user expectations are varied, underscoring the context-dependent nature of preferred anthropomorphism in LLM-driven HRI \cite{herathFirstImpressionsHumanoid2025, dellannaSONARAdaptiveControl2024}.

- \textit{\textbf{Usability.}}
In the age of LLMs, HRI research frequently employs standard metrics like the System Usability Scale (SUS) to assess perceived usability across diverse applications \cite{leusmannInvestigatingLLMDrivenCuriosity2025, karliAlchemistLLMAidedEndUser2024a, ikedaMARCERMultimodalAugmented2025a, xuExploringUseRobots2025a, mahadevanImageInThatManipulatingImages2025, wangCrowdBotOpenenvironmentRobot2024, kodurExploringDynamicsHumanRobot2025, limaPromotingCognitiveHealth2025}. However, the unique capabilities of LLMs necessitate an expanded view of usability. Evaluations now also incorporate the quality of \textbf{dialogue-based interaction}, using instruments like the Chatbot Usability Scale \cite{geGenComUIExploringGenerative2025}. Other critical LLM-specific dimensions include adherence to user preferences for \textbf{personalization} \cite{zhangBalancingUserControl2025a} and the provision of adjustable features that enhance \textbf{transparency and control} \cite{panACKnowledgeComputationalFramework2025}. These aspects are crucial for building trust \cite{bassiounyUJIButlerSymbolicNonsymbolic2025} and managing potential usability issues, such as users being distracted by overly engaging content \cite{hoSETPAiREdDesigningParental2025}.

- \textit{\textbf{Safety.}}
The integration of LLMs has broadened the scope of safety evaluation in HRI. While traditional measures of \textbf{perceived safety}—assessing users' comfort and security, often with the Godspeed Questionnaire—remain standard practice \cite{kimUnderstandingLargelanguageModel2024d, kamelabadComparingMonolingualBilingual2025, geGenComUIExploringGenerative2025, grassiGroundingConversationalRobots2024, farooqDAIMHRINewHumanRobot2024, zhangLargeLanguageModels2023b, salemComparativeHumanrobotInteraction2024, bastinGPTAllySafetyorientedSystem2025}, LLMs introduce critical new safety dimensions. A primary concern is \textbf{content safety}, as the generative nature of LLMs can produce outputs that are inappropriate or misaligned with user values \cite{hoSETPAiREdDesigningParental2025}. Furthermore, the inherent \textbf{unpredictability and potential for logical failures} in LLMs can translate directly into physical risks when these models guide a robot's actions \cite{bassiounyUJIButlerSymbolicNonsymbolic2025}. Consequently, contemporary research also evaluates the efficacy of technical safeguards, such as constraining output tokens \cite{laiNaturalMultimodalFusionBased2025} or modifying robot behavior \cite{kodurExploringDynamicsHumanRobot2025}, to mitigate these multifaceted risks.

- \textit{\textbf{Cognitive Load and Workload.}}
LLMs introduce a duality to workload assessment in HRI, with the potential to both reduce and create new cognitive demands. Traditional tools like the NASA-TLX questionnaire are still widely used to measure perceived cognitive load \cite{leusmannInvestigatingLLMDrivenCuriosity2025, mahadevanImageInThatManipulatingImages2025}. On one hand, LLMs can significantly lower \textbf{communication effort} by making instructions more intuitive \cite{geGenComUIExploringGenerative2025} and enable \textbf{cognitive offloading} by taking over tasks for the user \cite{hoSETPAiREdDesigningParental2025, liStargazerInteractiveCamera2023}. On the other hand, they can introduce \textbf{new complexities}, particularly in coordinating multi-agent systems \cite{dellannaSONARAdaptiveControl2024} or in safety-critical domains where supervised autonomy must be carefully managed to avoid increasing stress during error recovery. Importantly, users' pre-existing expectations about LLM capabilities can also shape workload perceptions independently of the actual interaction, highlighting a key methodological consideration \cite{rosenPreviousExperienceMatters2024}.

\section{Applications}
\label{sec:application}
Through our analysis, we identified eight principal application domains where LLMs were incorporated into HRI. We classified the existing works into the following high-level clusters: 1) social and conversational systems, 2) healthcare and wellbeing, 3) domestic and everyday use, 4) public spaces service, 5)industrial manufacturing, 6) AR/VR-enabled interactions, 7) teaching and education, 8) other. 

\begin{figure*}[h]
    \centering
    \includegraphics[width= 1\textwidth]{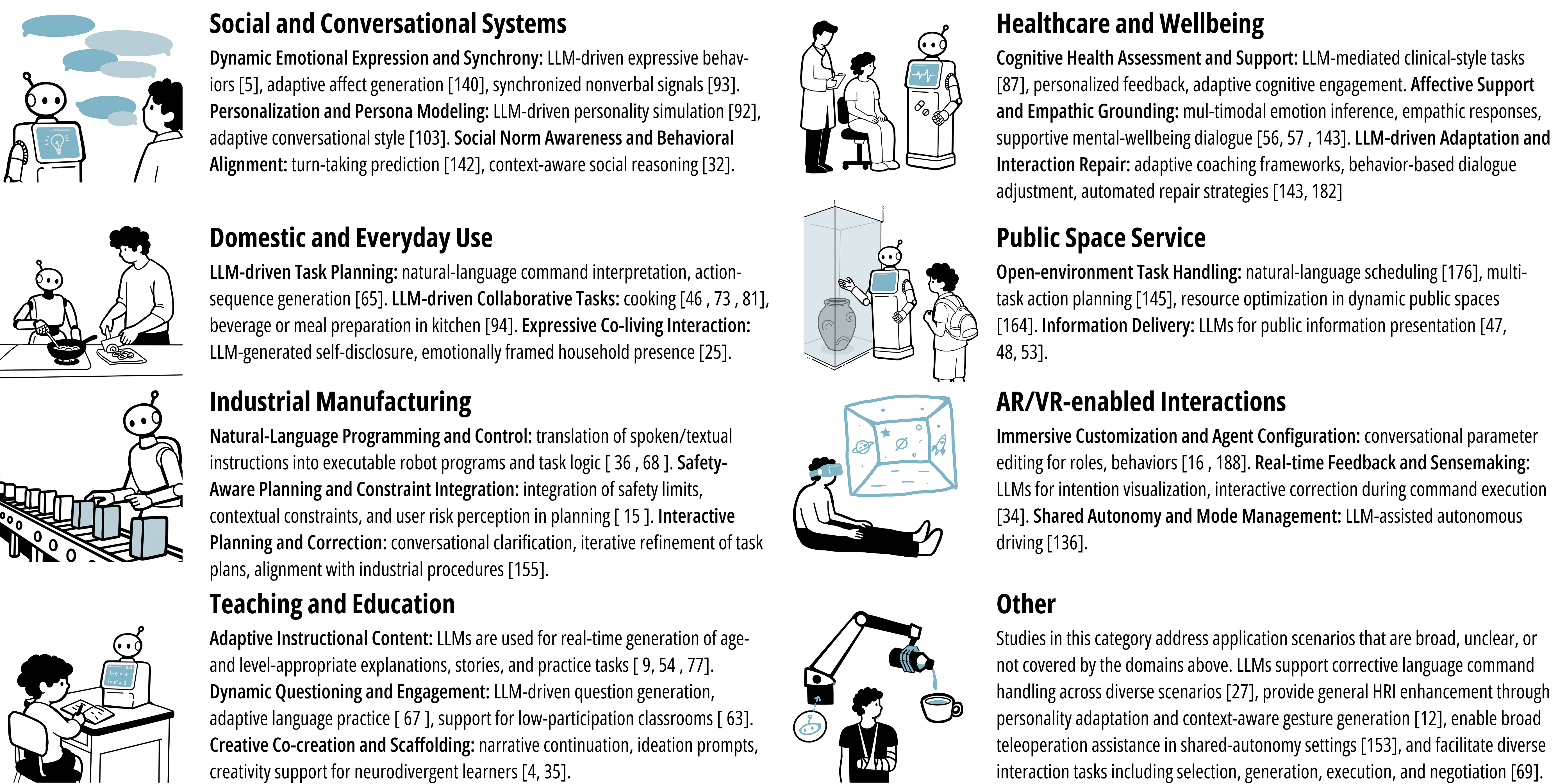}
    \caption{Usage scenarios and sub-categorization by LLM capabilities.}
    \label{fig:application}
    \Description{This figure presents the classification of application domains for LLM-driven human-robot interaction (HRI) systems, derived from the 86 studies included in this systematic review. Each domain is paired with LLM-enabled functional implementations, scenario-matching iconography, and corresponding literature references (numerical brackets): (1) Social and Conversational Systems: Encompasses LLM-driven dynamic emotional expression/synchrony [5, 140, 93], personalization/persona modelling [92, 103], and social norm awareness/behavioral alignment [142, 32]; illustrated by a robot engaging in dialogue with a human. (2) Domestic and Everyday Use: Includes LLM-driven task planning [65], collaborative tasks (e.g., cooking, meal preparation) [46, 73, 81, 94], and expressive co-living interaction [25]; illustrated by a robot assisting with household activities. (3) Industrial Manufacturing: Covers LLM-enabled natural-language programming/control [36, 68], safety-aware planning/constraint integration [15], and interactive planning/correction [155]; illustrated by a robot operating in an industrial production scenario. (4) Teaching and Education: Features LLM-driven adaptive/level-appropriate explanations [9, 54, 77], dynamic questioning/engagement [67, 63], and creative co-creation/scaffolding [4, 35]; illustrated by a robot supporting a learner in a classroom setting. (5) Healthcare and Wellbeing: Includes LLM-mediated cognitive health assessment/support [87], affective support/empathic grounding [56, 57, 143], and interaction repair [143, 182]; illustrated by a robot assisting a user in a healthcare context. (6) Public Space Service: Encompasses LLM-driven open-environment task handling [176, 145] and public information delivery [46, 47, 153]; illustrated by a robot providing service in a public setting. (7) AR/VR-enabled Interactions: Covers LLM-powered immersive customization/agent configuration [16, 188], real-time feedback/sensemaking [34], and shared autonomy/mode management [136]; illustrated by a user engaging with a robot via AR/VR. (8) Other: Addresses broad/unclear scenarios not covered by prior domains, including LLM-supported corrective language command handling [27], general HRI enhancement [12], teleoperation assistance [153], and diverse interaction tasks [69]; illustrated by a robot assisting with daily tasks.}
\end{figure*}

Beyond grouping works by their application domains, we further refined each category based on the specific capabilities that LLMs contribute to these scenarios—such as enhanced perception and contextual understanding, generative and agentic interaction capabilities, and iterative optimization and alignment mechanisms. This allowed us to highlight not only where LLMs are used, but how they reshape the functional roles of robots in these contexts. Figure~\ref{fig:application} summarizes these categories, including the associated papers and the LLM-enabled sub-capabilities that characterize each domain. The identified domains demonstrate the primary ways LLM integration advances HRI research by addressing longstanding challenges and enabling more adaptive robotic systems.

\section{Key Design Considerations and Challenges}
\label{sec:challenges}
Building upon the systematic synthesis of existing literature, this section integrates broader academic perspectives to address \textbf{RQ4}. Specifically, we distill eleven design considerations and challenges identified across our corpus. Building on Sense-Interaction-Alignment framework in~\textbf{Section~\ref{section:LLMs-HRI}}, these considerations are categorized similarly into three aspects: (1) sense—understanding and perception (challenges 1-4), (2) interaction—action and agency (challenges 5-8), and (3) alignment-adaptation and repair (challenges 9-11).

- \textbf{\textit{Challenge-1. Reliability of LLM-driven Understanding:}}
The primary challenge in LLM-driven HRI is the inherent unreliability of robotic understanding, which spans from technical performance to high-level reasoning. First, foundational limitations such as latency, inconsistency, and unpredictability ~\cite{limaPromotingCognitiveHealth2025, kimUnderstandingLargelanguageModel2024d, wangCrowdBotOpenenvironmentRobot2024} fundamentally undermine the real-time nature of robot perception. Beyond these surface-level issues, a deeper layer of the challenge resides in the fragility of high-level cognitive reasoning, particularly in understanding human social cues. For instance, studies have identified significant failures in LLMs’ ability to interpret humor ~\cite{grassiEnhancingLLMBasedHumanRobot2024} or perform spatial and quantitative reasoning in embodied contexts ~\cite{hoSETPAiREdDesigningParental2025}. Second, while emerging strategies have attempted to mitigate these shortcomings through grounded prompting, multi-role validation, or human-in-the-loop corrections ~\cite{karliAlchemistLLMAidedEndUser2024a, jinRobotGPTRobotManipulation2024, cuiNoRightOnline2023a}, these interventions often introduce new layers of complexity and opacity. 
The ultimate challenge remains achieving a level of reliable, interpretable understanding without constant external supervision.

- \textbf{\textit{Challenge-2. Multimodal Perception of Emotional Intelligence:}}
While the integration of LLMs has enabled robots to respond fluently and enrich interactions through prosody, intonation, and onomatopoeic expressions such as ``oh,'' ``wow,'' or ``haha''~\cite{grassiGroundingConversationalRobots2024,wangChallengesAdoptingCompanion2025}, achieving genuine emotional intelligence remains a multi-layered challenge. First, despite advances in empathy calibration and contextual sensitivity in controlled scenarios~\cite{elfleetInvestigatingImpactMultimodal2024, kamelabadComparingMonolingualBilingual2025}, robots struggle with the instability of multimodal affective signals. In heterogeneous and evolving contexts, signals like hesitation, stress, or collective affect remain inherently noisy and ambiguous~\cite{choARECADesignSpeculation2023, nardelliIntuitiveInteractionCognitive2025}, making it difficult for robots to perceive emotional nuances consistently beyond surface-level linguistic alignment.
Second, a gap persists between generating affective responses and possessing a functional Theory of Mind. Even with hybrid architectures for socio-cognitive grounding \cite{dellannaSONARAdaptiveControl2024, vermaTheoryMindAbilities2024}, most frameworks fail to maintain stability in longitudinal settings, often creating an illusion of understanding rather than genuine intent clarification~\textbf{(Section~\ref{subsection:Human-Oriented})}. Finally, this perceptual gap creates a turn-taking bottleneck \cite{skantzeApplyingGeneralTurntaking2025}. Current systems cannot yet replicate the subtle non-verbal cues (like gaze, fillers, and micropauses) used to negotiate conversational flow. 
Consequently, bridging fluent generative output with robust multimodal perception remains a critical hurdle for social HRI.

- \textbf{\textit{Challenge-3. Multimodal information sensing and alignment:}}
While LLM-driven robots rely on integrating various modalities for context-aware interaction (sensing), the core challenge lies in achieving semantic alignment across disparate data streams (e.g., the resolution of modality conflicts). 
For instance, a robot may face a disconnect when a user's verbal input (e.g., a joke) contradicts their physical cues (e.g., an angry facial expression or aggressive posture)~\cite{elfleetInvestigatingImpactMultimodal2024}. Current LLM-driven architectures often lack the nuanced arbitration logic to weigh these conflicting signals, leading to cascading errors in social interpretation that compromise interactional safety. Furthermore, multimodal alignment requires making the sensing process itself more legible and transparent. Prior work suggests that simple alignment of timestamps is insufficient; instead, the system must communicate its internal interpretation of fused data to the user. As demonstrated by MARCER~\cite{ikedaMARCERMultimodalAugmented2025a}, while transparent feedback via hybrid modalities can mitigate misalignment, the failure to resolve subtle social cue discrepancies often exacerbates user frustration or perceived unfairness. The challenge, therefore, remains in developing robust ``cross-modal reasoning'' that can dynamically prioritize modalities based on the situational context, rather than merely aggregating them into a singular text-based prompt for the LLM.

- \textbf{\textit{Challenge-4. Equitable Engagement in Multi-User Scenarios:}}
As LLM-driven robots are increasingly deployed in environments involving multiple users, such as cafés, classrooms, and domestic settings, equitable engagement becomes a critical yet underexplored challenge~\cite{grassiStrategiesControllingConversation2025}. To elaborate, robots must not only balance diverse preferences, but also perceive and interpret complex social dynamics among co-present users to avoid conflict and exclusion~\cite{panACKnowledgeComputationalFramework2025}. Prior work has shown that multi-user settings substantially complicate social sensing; for instance, Skantze and Irfan indicated that multiple users makes it more challenging for robots to determine if the users are addressing the robot or each other~\cite{skantzeApplyingGeneralTurntaking2025}. Ethical risks arise when one user attempts to elicit actions that could harm others, such as simulating emergencies in VR studies~\cite{stampfExploringPassengerAutomatedVehicle2024}. Moreover, implicit biases from human-human interactions can carry over into human-robot interactions when robots lack sufficient social awareness to detect and compensate for unequal engagement patterns, thereby exacerbating inequities in multi-user contexts. To mitigate these, robots could integrate seamlessly into the social and informational frameworks of their environment rather than operate as isolated agents, ensuring responsible coordination, fairness, and safety across users~\cite{wangChallengesAdoptingCompanion2025}.

- \textbf{\textit{Challenge-5. Morphology-Aligned Social Intelligence:}}
Morphology constrains the range of behavioral expressions a robot can produce, thereby shaping how social agency is physically expressed during interaction. LLMs enable robots to interpret semantic meaning and engage in flexible dialogue, which in turn raises expectations for the robot’s morphology (e.g., head, arms) to express corresponding social and emotional capabilities~\cite{kimUnderstandingExpectationsRobotic2025}. For example, LLMs can be used to generate or refine expressive robot behaviors such as nodding~\cite{taoLAMSLLMDrivenAutomatic2025a}, as well as to produce richly articulated motion sequences~\cite{leusmannInvestigatingLLMDrivenCuriosity2025, mahadevanGenerativeExpressiveRobot2024}~\textbf{ (Section~\ref{subsubsection:GenerativeSocial})}. However, this enhancement is not a linear progression toward unconditionally anthropomorphizing robots~\cite{choLivingAlongsideAreca2025}. Achieving fully human-like expressiveness remains technically challenging and ethically fraught~\cite{pinedaSeeYouLater2025a}. Mismatches between highly fluent linguistic output and comparatively rudimentary physical behavior can produce expectation gaps~\cite{herathFirstImpressionsHumanoid2025, grassiStrategiesControllingConversation2025}; LLM hallucinations may trigger trust ruptures~\cite{wilcockErrRoboticEarn2023a, laiNaturalMultimodalFusionBased2025}; and overly anthropomorphic presentations may heighten user discomfort like uncanny valley~\cite{salemComparativeHumanrobotInteraction2024}. 
Consequently, we suggest future HRI design carefully balance LLM-driven social agency with the morphology limitations through deliberate design choices, such as prioritizing physical compliance over anthropomorphic realism~\cite{bastinGPTAllySafetyorientedSystem2025} or incorporating explicit physical cues~\cite{wangChallengesAdoptingCompanion2025, zhangPromptingEmbodiedAI2025}, to mitigate risks associated with over-humanization.

- \textbf{\textit{Challenge-6. Balancing Autonomy and Human Oversight:}}
LLMs substantially expand robots’ autonomy by enhancing natural language control~\cite{farooqDAIMHRINewHumanRobot2024, padmanabhaVoicePilotHarnessingLlms2024, jinRobotGPTRobotManipulation2024}, dynamic code generation~\cite{bastinGPTAllySafetyorientedSystem2025, karliAlchemistLLMAidedEndUser2024a}, and common sense reasoning~\cite{mahadevanGenerativeExpressiveRobot2024, yeImprovedTrustHumanRobot2023}. However, this shift raises challenges for balancing LLM-driven agency with human control. First, high-granularity end-user programming can cause robots to be perceived as tools rather than autonomous partners, reducing the sense of intelligence and social presence~\cite{zhangBalancingUserControl2025a}. Second, while LLM-driven automation may increase task efficiency, it can shift focus toward individual outcomes and inadvertently suppress opportunities for meaningful communication with other people~\cite{shiradoRealismDrivesInterpersonal2025}. Third, socially proactive behaviors (e.g., small talk and emotional responses) may impose unintended interactional obligations on users~\cite{pinedaSeeYouLater2025a}.
At the same time, some domains (e.g., assistive and manipulation tasks) benefit from shared autonomy, where users can provide natural-language online corrections to refine robot behavior~\cite{cuiNoRightOnline2023a}~\textbf{ (Section~\ref{subsubsection:CollaborativeTask})}. This highlights the need for hybrid autonomy designs that balance LLM-driven initiative with task efficiency~\cite{shiradoRealismDrivesInterpersonal2025}. Moving forward, autonomy could be structured around shared-control or Human-in-the-Loop frameworks, ensuring that LLM-driven autonomy remains aligned with user intent even in complex or high-stakes environments~\cite{antonyXpressSystemDynamic2025a, limaPromotingCognitiveHealth2025, leusmannInvestigatingLLMDrivenCuriosity2025}.

- \textbf{\textit{Challenge-7. Balancing Trust and Overtrust:}}
In LLM-driven HRI, trust mediates the delegation of authority to robots, yet LLMs often exacerbate overtrust and overreliance \cite{axelssonOhSorryThink2024, hoSETPAiREdDesigningParental2025, itoRobotDynamicallyAsking2025a, kontogiorgosQuestioningRobotUsing2025, leusmannInvestigatingLLMDrivenCuriosity2025}. 
First, trust is induced multidimensionally; as synthesized in \textbf{Section~\ref{subsubsection:GenerativeSocial}}, factors like personalization \cite{nardelliIntuitiveInteractionCognitive2025, bannaWordsIntegratingPersonality2025}, social norm awareness \cite{dellannaSONARAdaptiveControl2024, elfleetInvestigatingImpactMultimodal2024}, and human-likeness \cite{antonyXpressSystemDynamic2025a, choLivingAlongsideAreca2025} lower trust thresholds. Beyond factual accuracy, this process requires contextual grounding and expectation management \cite{wilcockErrRoboticEarn2023a}; however, anthropomorphic cues often trigger social interpretations that exceed actual system reliability.
Second, technical opacity hinders trust calibration. While objective metrics like the Attention Arbitration Ratio can predict trust levels \cite{goubardCognitiveModellingVisual}, technical fragilities (like hallucinations and inconsistent reasoning) foster uncalibrated overtrust \cite{geGenComUIExploringGenerative2025, liStargazerInteractiveCamera2023, reimannWhatCanYou2025a}. This is critical when perceived confidence masks underlying uncertainty \cite{kontogiorgosQuestioningRobotUsing2025, shiradoRealismDrivesInterpersonal2025}, causing difficult-to-mitigate trust ruptures. Finally, trust repair remains asymmetric. Despite multi-level strategies involving apologies or explanations \cite{arjmandEmpathicGroundingExplorations2024} (\textbf{Section~\ref{subsubsection:Multi-LevelRepair}}), effectively recalibrating lost confidence is still an open question.

- \textbf{\textit{Challenge-8. Safeguarding Privacy and Mitigating Safety Risks:}}
In LLM-driven HRI, privacy and safety transcend traditional cybersecurity as robots act as embodied observers in private spaces. 
First, LLMs enable semantic privacy risks: rather than just capturing raw data, robots perform continuous multimodal reasoning to infer user habits and social dynamics \cite{wangPepperPoseFullbodyPose2024, shenSocialRobotsSocial2025}. This ``semantic surveillance'' creates intrusive risks of accidental data exposure based on deep contextual understanding.
Second, LLMs as decision engines introduce unpredictable physical risks. Model hallucinations or opacity can lead to hazardous actions, such as misinterpreting intent in high-stakes tasks~\cite{limaPromotingCognitiveHealth2025, shiradoRealismDrivesInterpersonal2025}. While strategies like prompt deletion or age-specific guidelines exist~\cite{grassiStrategiesControllingConversation2025, aliInclusiveCocreativeChildrobot2025a}, these reactive measures struggle with the inherent unpredictability of an agent physicalizing its reasoning. Finally, the contextual variability of safety norms across cultures complicates the design of universal frameworks. Consequently, safeguarding LLM-driven HRI requires moving beyond encryption toward frameworks addressing the unique vulnerabilities of embodied, context-aware interaction.

- \textbf{\textit{Challenge-9. Stabilizing Personalization and Social Alignment:}}
Personalization has been widely praised for sustaining user enjoyment ~\cite{kamelabadComparingMonolingualBilingual2025,spitaleVITAMultiModalLLMBased2025a,wangChallengesAdoptingCompanion2025}~\textbf{ (Section~\ref{subsubsection:Personalization&Memory})}. However, personalization can also create unintended emotional dependence in vulnerable users, where the sudden withdrawal of robots leads to significant harm ~\cite{hsuResearchCareReflection2025}. Empathic calibration strategies, though powerful in the short term, risk losing their novelty and eroding trust over time~\textbf{(Section~\ref{subsection:Generative&Agentic})}. For instance, affective styles are central to sustaining comfort and naturalness in short-term interactions ~\cite{karliAlchemistLLMAidedEndUser2024a,pinedaSeeYouLater2025a,wangCrowdBotOpenenvironmentRobot2024}, yet the inconsistency of LLMs (e.g., humor understanding) challenges the reliability of such strategies~\cite{grassiEnhancingLLMBasedHumanRobot2024}. Similarly, the unpredictability of LLM outputs raises risks for repair strategies: while transparency and empathy are expected in many domains~\cite{arjmandEmpathicGroundingExplorations2024,limaPromotingCognitiveHealth2025, shenSocialRobotsSocial2025}, overly elaborate or irrelevant repairs can disrupt conversational flow~\cite{axelssonOhSorryThink2024, pinedaSeeYouLater2025a}. Moreover, some users may actually prefer robotic companionship precisely because it bypasses the unpredictability and emotional complexity of human relationships~\cite{wangChallengesAdoptingCompanion2025}. If these preferences become widespread, how designers should balance personalization-driven alignment with long-term user well-being remains an open question.

- \textbf{\textit{Challenge-10. Sustaining Long-Term Engagement:}}
Effective integration of LLM-driven robots into daily routines hinges on sustaining engagement beyond the initial novelty effect~\cite{huDesigningTelepresenceRobots2025}. The first challenge lies in evolving personalization; while LLMs support varied interactions in healthcare and domestic settings~\cite{laiNaturalMultimodalFusionBased2025, wangPepperPoseFullbodyPose2024}, existing systems often fail to transform short-term memory into a stable, growing personality. Current behavioral modeling methods~\textbf{(Section~\ref{subsubsection:Personalization&Memory})} struggle to keep pace with shifting human expectations, leading to ``stale'' interactions as the honeymoon phase fades.
Secondly, a significant methodological gap persists in longitudinal validation. Despite emerging field studies in real-world environments~\textbf{(Section~\ref{subsec:field})}, most empirical evidence remains limited to ``snapshot'' observations. As noted in the SET-PAiREd~\cite{hoSETPAiREdDesigningParental2025}, brief deployments fail to capture how roles and trust dynamics evolve as LLM integrates into the domestic fabric. This lack of sustained data prevents a systematic understanding of adaptation fatigue over time.
Finally, the transition from tools to persistent companions raises profound ethical concerns regarding emotional dependency and value drift. 
Thus, the challenge lies in designing frameworks that co-evolve with users while safeguarding social well-being and ethical boundaries in longitudinal settings.

- \textbf{\textit{Challenge-11. Proactive Repair for Diverse Contexts:}}
Repair presents a critical design space for LLM-driven HRI. Robots today can already leverage dialogue history to provide explanations, apologies, and repair strategies~\cite{arjmandEmpathicGroundingExplorations2024} \textbf{(Section~\ref{subsubsection:Multi-LevelRepair})}. However, repair strategies cannot remain reactive templates; they must become proactive and adaptive, evolving with longitudinal expectations~\cite{aliInclusiveCocreativeChildrobot2025a,axelssonOhSorryThink2024,geGenComUIExploringGenerative2025,kontogiorgosQuestioningRobotUsing2025,padmanabhaVoicePilotHarnessingLlms2024,pinedaSeeYouLater2025a,reimannWhatCanYou2025a}. In-context learning provides early evidence for this shift, demonstrating that robots can anticipate errors, distinguish genuine interruptions from noise, propose grounded alternatives, and transparently explain their limitations ~\cite{leusmannInvestigatingLLMDrivenCuriosity2025,panACKnowledgeComputationalFramework2025,reimannWhatCanYou2025a}. This reconfiguration moves repair from a corrective action into a continuous socio-emotional negotiation that preserves trust and transparency across long-term engagement ~\cite{axelssonOhSorryThink2024,choARECADesignSpeculation2023,geGenComUIExploringGenerative2025,reimannWhatCanYou2025a}. However, there are still limited investigations in how such adaptive repair mechanisms can be calibrated to different domains, from casual conversation to high-stakes contexts.

\section{Discussion and Limitations}
Several limitations arise from our corpus construction and assessment process. First, although our search terms were iteratively refined to cover major HRI and robotics venues, they could not fully capture the breadth of emerging work across interdisciplinary domains. This may have resulted in the under-representation of studies from psychology, sociology, and communication research that examine LLM-mediated social interaction and human–machine boundaries.
Further, our inclusion criteria included studies with both explicit (e.g., LLMs as system components) and implicit (e.g., LLMs as tools for experimental design) applications, which may introduce heterogeneity to the data collection. 
For example, Rosén et al.~\cite{rosenPreviousExperienceMatters2024} used GPT-3 to replace Wizard-of-Oz methods to avoid introducing human actor expectations, while Grassi et al.~\cite{grassiStrategiesControllingConversation2025} employed LLMs to increase response variability. Compared with explicit use, such implicit uses were not central to the studies and often lacked rigorous evaluation, limiting cross-study comparability. Moreover, the assessment of conceptually ambiguous and borderline studies inevitably reflects our own interpretive stance, and other researchers might reasonably draw different boundaries. Future reviews may benefit from broader interdisciplinary coverage and differentiated treatment of implicit versus explicit LLM integration.

In addition to the scoping of research themes, our selection of publication types also warrants discussion. We excluded Late-Breaking Reports (LBRs)~\footnote{Late-Breaking Reports are a publication category in the ACM/IEEE International Conference on Human–Robot Interaction (HRI).} to ensure our dataset primarily reflects rigorously validated, methodologically comprehensive research. While we acknowledge LBRs often showcase cutting-edge LLM applications in HRI, focusing on full archival papers prioritizes synthesis reliability over preliminary findings. This trade-off ensures a robust foundation for our analysis. Future reviews could specifically synthesize these emerging LBRs to capture the field's rapid, real-time evolution. It has been truly inspiring to witness and synthesize the rapid evolution of this flourishing field through our work. We hope this review contributes to the collective endeavor of building more intelligent, human-centric robotic systems.


\section{Conclusion}
In this paper, we present our systematic review and taxonomy of LLM-driven HRI research, synthesizing existing research approaches, interaction designs, and evaluation strategies across identified studies. Our goal is to provide a common ground for researchers to understand how LLMs are shaping HRI systems. Building on this synthesis, we introduce the Sense–Interaction–Alignment framework to consolidate how LLMs enable new embodied capabilities, including enhanced contextual sensing, generative and socially grounded interaction, and continuous human-aligned adaptation across scenarios. In addition, to further stimulate future research at the intersection of LLMs and HRI, we discuss key design considerations and emerging challenges, such as multimodal grounding, morphology-aligned social intelligence, and sustaining long-term alignment in real-world settings. We hope our review, taxonomy, and identified research directions will guide and inspire future work in LLM-driven HRI.

\section{Disclosure about Use of LLM}
Portions of this paper were refined using GPT and DeepL for clarity and grammatical accuracy; all content has been thoroughly proofread by the authors.


\begin{acks}
We gratefully acknowledge the support of the Guangdong Provincial Key Lab of Integrated Communication, Sensing and Computation for Ubiquitous Internet of Things Grant (No. 2023B1212010007), the 111 Center Grant (No. D25008), and Natural Science Foundation of China (No. 62402271). This work was completed during the internships of Yufeng Wang, Anastasia Nikolova, and Yuxuan Wang in the Human-Centered Intelligence+ Summer Research Program, and we sincerely thank the program and support team. 
\end{acks}
\bibliographystyle{ACM-Reference-Format}
\bibliography{chi26-649}

\onecolumn

\appendix
\section{Retrieval Keywords for Publication Trend}
\label{section:appendix_A}

This appendix elaborates on the search keywords and Boolean retrieval strategies adopted in the ACM digital library for analyzing the publication trends of Large Language Models (LLMs), Human-Robot Interaction (HRI), and their interdisciplinary research (LLM-driven HRI) over the decade from 2015 to 2025. All retrieval operations were uniformly executed on September 9, 2025. The annual number of relevant publications retrieved for each category is as follows:

\begin{itemize}
    \item \textbf{Large Language Models (LLMs)}: 2015 (15), 2016 (11), 2017 (12), 2018 (13), 2019 (10), 2020 (19), 2021 (33), 2022 (89), 2023 (1119), 2024 (5074), 2025 (5498)
    \item \textbf{Human–Robot Interaction (HRI)}: 2015 (384), 2016 (408), 2017 (549), 2018 (565), 2019 (384), 2020 (840), 2021 (681), 2022 (701), 2023 (908), 2024 (1206), 2025 (814)
    \item \textbf{LLM-driven HRI}: 2015 (1), 2016 (2), 2017 (0), 2018 (1), 2019 (1), 2020 (5), 2021 (7), 2022 (21), 2023 (108), 2024 (311), 2025 (268)
\end{itemize}

It should be noted that these retrieved publication counts are solely for trend reference purposes, and the relevance of the retrieved articles to the core research themes (LLMs, HRI, and LLM-driven HRI) remains to be further verified. The detailed temporal publication trends (including visualizations of annual volume changes) are presented in Figure \ref{fig:trend}.

\begin{table}[ht!]
\centering
\small
\caption{Search keywords and strategies used for literature retrieval in the ACM digital library (2015–2025).}
\label{tab:search_keywords}
\Description{This table outlines the search keywords and strategies used to retrieve literature from the ACM Digital Library for the period of 2015 to 2025, organized into three distinct categories. For the Large Language Models (LLM) category, the search query looked for articles containing any of the following terms: large language model, LLM, or foundation model. For the Human-Robot Interaction (HRI) category, the query included the terms human-robot interaction, HRI, human-robot collaboration, or HRC. To identify papers at the intersection of these fields, under the LLM-driven HRI category, a complex boolean query was used which required an article to contain at least one term from each of three groups.}
\begin{tabular}{p{0.25\linewidth} p{0.70\linewidth}}
\toprule
\textbf{Category} & \textbf{Search Keywords / Query} \\
\midrule
Large Language Models (LLMs) & ``large language model" OR``LLM" OR``foundation model" \\
Human–Robot Interaction (HRI) &``human-robot interaction" OR ``HRI" OR``human-robot collaboration" OR``HRC" \\
LLM-driven HRI & ("large language model" OR LLM OR ChatGPT OR GPT-3 OR GPT-4) AND (robot OR robotics OR``social robot" OR``humanoid robot") AND ("human-robot interaction" OR HRI OR``human-robot collaboration" OR HRC) \\
\bottomrule
\end{tabular}
\end{table}

\section{Related Works and Contributions}
\label{section:appendix_B}
\begin{table}[ht!]
\centering
\small
\caption{Overview of representative review, survey, and meta-study works on LLMs and foundation models in robotics and HRI, including authors, publication year, research focus, and core contributions.}

\begin{tabular}{p{0.12\linewidth} p{0.07\linewidth} p{0.06\linewidth} p{0.15\linewidth} p{0.50\linewidth}}
\toprule
\textbf{Author} & \textbf{Type} & \textbf{Year} & \textbf{Focus} & \textbf{Contribution} \\
\midrule
Zeng et al. \cite{zengLargeLanguageModels2023} 
& Survey & 2023 
& LLMs in Robotics
& Highlights the benefits of LLMs for robotics. \\

Wang et al. \cite{wangLargeLanguageModels2025} 
& Review & 2025 
& LLMs in Robotics
& Provides an overview of the integration of LLMs into robotic systems and tasks. \\

Firoozi et al. \cite{firooziFoundationModelsRobotics2023} 
& Survey & 2023
& Foundation Models in Robotics
& Surveys the promising applications of foundation models in robotics. \\

Jeong et al. \cite{jeongSurveyRobotIntelligence2024} 
& Survey & 2024
& Robotic Systems
& Explores the potential impact and applicability of LLMs on robotics. \\

Kim et al. \cite{kimSurveyIntegrationLarge2024} 
& Survey & 2024
& LLMs in Intelligent Robotics
& Offers detailed prompt engineering guidelines and categorizes LLMs in robotics. \\

Wang et al. \cite{wangSurveyLargeLanguage2024} 
& Survey & 2024
& LLM-based Autonomous Agents
& Proposes a review of LLM-based autonomous agents from a holistic perspective. \\

Liu et al. \cite{liuIntegratingLargeLanguage2025} 
& Review & 2025
& Robotic Autonomy
& Surveys the integration of LLMs into autonomous robotics. \\

Guo et al. \cite{liuIntegratingLargeLanguage2025} 
& Survey & 2024
& LLM-based Multi-Agents
& Provides a taxonomy of LLM-MA systems. \\

Li et al. \cite{liLargeLanguageModels2025} 
& Survey & 2025
& Multi-Robot Systems
& Provides the first exploration of integrating LLMs into MRS. \\

Xi et al. \cite{xiRisePotentialLarge2025} 
& Survey & 2025
& LLM-based Agents
& Present a framework for LLM-based agents, comprising three components. \\

Lin et al. \cite{linEmbodiedAILarge2024b} 
& Survey & 2024
& Embodied AI
& Proposes PALoop framework, an emotional logic engine based on LLMs. \\

Salimpour et al. \cite{salimpourEmbodiedAgenticAI2025} 
& Survey & 2025
& Embodied Agentic AI
& Proposes a taxonomy of LLM integration and provides a comparative analysis. \\

Zhang et al. \cite{zhangLargeLanguageModels2023} 
& Review & 2023
& LLMs in HRI
& Provides the first review of LLM applications in HRI across three categories and identifies three major challenges. \\\addlinespace

Shi et al. \cite{shiHowCanLarge2024} 
& Survey & 2024
& LLMs in Socially Assistive HRI
& The first surveys to focus on LLMs in SARs, informing the potential of LLMs. \\\addlinespace

Atuhurra \cite{atuhurraLeveragingLargeLanguage2024} 
& Meta-study & 2024
& LLMs in HRI
& Identifies thirteen key benefits of LLMs in HRI and three key risks. \\\addlinespace

Zou et al. \cite{zouLLMBasedHumanAgentCollaboration2025} 
& Survey & 2024
& LLM-Based HAI and HAC
& Proposes a taxonomy for LLM-HAS, indicating challenges and opportunities. \\
\bottomrule
\end{tabular}
\label{tab:relatedworks}
\Description{This table summarizes recent academic works and their contributions regarding Large Language Models (LLMs) in robotics and related fields. First, a 2023 survey by Zeng et al. focuses on``LLMs in Robotics" and highlights their benefits. Following this, a 2025 review by Wang et al. also centers on``LLMs in Robotics," aiming to provide an overview of their integration into robotic systems and tasks. A 2023 survey from Firoozi et al. delves into``Foundation Models in Robotics," surveying their promising applications. Moving to 2024, a survey by Jeong et al. addresses``Robotic Systems," exploring the potential impact and applicability of LLMs; in the same year, a survey by Kim et al. focuses on``LLMs in Intelligent Robotics," offering detailed prompt engineering guidelines and categorizing LLMs in the field. Also in 2024, another survey by Wang et al. discusses``LLM-based Autonomous Agents," proposing a review from a holistic perspective. Looking ahead to 2025, a review by Liu et al. will survey the integration of LLMs into``Robotic Autonomy." Back in 2024, Guo et al. published a survey on``LLM-based Multi-Agents," providing a taxonomy of LLM-MA systems. In 2025, a survey by Li et al. will present the first exploration of integrating LLMs into``Multi-Robot Systems" (MRS); in the same year, a survey by Xi et al. will present a three-component framework for``LLM-based Agents." In 2024, a survey by Lin et al. on``Embodied AI" proposes the PALoop framework, an emotional logic engine based on LLMs. Then in 2025, a survey by Salimpour et al. will discuss``Embodied Agentic AI," proposing a taxonomy for LLM integration and providing a comparative analysis. Within the field of Human-Robot Interaction (HRI), a 2023 review by Zhang et al. provides the first overview of LLM applications in HRI and identifies three major challenges. In 2024, a survey by Shi et al. is the first to focus on``LLMs in Socially Assistive HRI," informing the potential of LLMs. During the same year, a meta-study by Atuhurra identifies thirteen key benefits and three key risks of``LLMs in HRI." Finally, a 2024 survey by Zou et al. focuses on``LLM-Based Human-Agent Interaction (HAI) and Collaboration (HAC)," proposing a taxonomy for LLM-HAS and indicating its challenges and opportunities.}
\end{table}

\section{Retrieval Databases, Queries and Screening Results}
\label{section:appendix_C}

\begin{table}[H]
\centering
\small
\caption{Overview of databases, search queries, and screening results. The temporal scope of the search covers literature published between January 1, 2021 and August 1, 2025. For studies indexed in both ACM DL and IEEE Xplore, we followed a conservative consolidation rule whereby duplicates were attributed to ACM DL to avoid double-counting across venues.}
\Description{Overview of search databases, corresponding query strategies, and literature screening outcomes. The search timeframe covers literature published from January 1, 2021 to August 1, 2025. For duplicate studies indexed in both ACM DL and IEEE Xplore, a conservative consolidation rule is adopted: duplicates are attributed to ACM DL to prevent double-counting across venues. Most databases use the unified query: ("large language model" OR LLM OR ChatGPT OR GPT-3 OR GPT-4) AND (robot OR robotics OR "social robot" OR "humanoid robot") AND ("human-robot interaction" OR HRI OR "human robot collaboration" OR HRC). Due to venue-specific search constraints, Computers in Human Behavior employs two simplified keyword-based queries. The table presents four columns: Database, Search Query, number of Retrieved literatures, and number of Included literatures for each database.}
\label{tab:appendix-db-results}

\begin{tabular}{p{0.16\linewidth} p{0.65\linewidth} p{0.06\linewidth} p{0.06\linewidth}}
\toprule
\textbf{Database} & \textbf{Search Query} & \textbf{Retrieved} & \textbf{Included} \\
\midrule
ACM DL & ("large language model" OR LLM OR ChatGPT OR GPT-3 OR GPT-4)  AND  (robot OR robotics OR "social robot" OR "humanoid robot")  AND  ("human-robot interaction" OR HRI OR "human robot collaboration" OR HRC) & 709 & 53 \\

IEEE Xplore & same query as above & 157 & 24 \\

Nature & same query as above & 10 & 2 \\

Science Robotics & same query as above & 2 & 0 \\

International Journal of Social Robotics& \small{same query as above} & 10 & 7 \\

Computers in Human Behavior & \small{simplified keyword-based search due to venue-specific search constraints: (1) (LLM OR ChatGPT OR GPT-3 OR GPT-4) AND robot AND ("human-robot interaction" OR HRI OR "human robot collaboration" OR HRC), and (2) "large language model" AND (robot OR robotics OR "social robot" OR "humanoid robot") AND ("human-robot interaction" OR HRI OR "human robot collaboration" OR HRC)} & 6 & 0 \\
\bottomrule
\end{tabular}
\end{table}

\section{Details of the Survey Literature Statistics}
\label{section:appendix_D}
\begin{table}[H] 
  \centering
  \small 
  \caption{References of Section 4.1. Contextual Perception and Understanding}
  \Description{Classification of literature on contextual perception and understanding in LLM-driven HRI. The table is structured with three columns: Sense (categorization of perception and understanding dimensions), Number (count of relevant literature), and Papers (corresponding literature reference numbers). The perception and understanding dimensions are divided into two major categories: Multimodal Physical Perception (including three subcategories: Static and Semi-Static Context Injection with 19 papers, Modular Perception and Textual Abstraction with 32 papers, and Integrated Visual-Language Reasoning with 16 papers) and Human-Oriented Understanding (including three subcategories: Emotional Grounding with 18 papers, Task Intent Formulation with 39 papers, and Human Model Alignment with 19 papers). Each subcategory lists the reference numbers of the included literature to support the corresponding research dimension.} 
  
  \begin{tabular}{p{0.30\linewidth} p{0.10\linewidth} p{0.56\linewidth}}
    \toprule
    \textbf{Sense} & \textbf{Number} & \textbf{Papers} \\
    \midrule

\multicolumn{3}{l}{\textit{\textbf{Multimodal Physical Perception}}} \\[0.3em]
Static and Semi-Static Context Injection      & 19 & \cite{limaPromotingCognitiveHealth2025, pinedaSeeYouLater2025a, kamelabadComparingMonolingualBilingual2025, karliAlchemistLLMAidedEndUser2024a, ikedaMARCERMultimodalAugmented2025a, vermaTheoryMindAbilities2024, zhangLargeLanguageModels2023b, yanoUnifiedUnderstandingEnvironment2024, padmanabhaVoicePilotHarnessingLlms2024, hoSETPAiREdDesigningParental2025, laiNaturalMultimodalFusionBased2025, zuLanguageSketchingLLMdriven2024, tsushimaTaskPlanningFactory2025, loLLMbasedRobotPersonality2025a, dellannaSONARAdaptiveControl2024, rosenPreviousExperienceMatters2024, starkDobbyConversationalService2024, grassiEnhancingLLMBasedHumanRobot2024, jinRobotGPTRobotManipulation2024} \\
Modular Perception and Textual Abstraction    & 32 & \cite{grassiEnhancingLLMBasedHumanRobot2024, kodurExploringDynamicsHumanRobot2025, bassiounyUJIButlerSymbolicNonsymbolic2025, dellannaSONARAdaptiveControl2024, herathFirstImpressionsHumanoid2025, grassiGroundingConversationalRobots2024, laiNaturalMultimodalFusionBased2025, geGenComUIExploringGenerative2025, blancoAIenhancedSocialRobots2024, kimUnderstandingLargelanguageModel2024d, perella-holfeldParentEducatorConcerns2024a,zhangPromptingEmbodiedAI2025, bastinGPTAllySafetyorientedSystem2025, pintoPredictiveTurntakingLeveraging2024, latifPhysicsAssistantLLMpoweredInteractive2024, cuiNoRightOnline2023a, skantzeApplyingGeneralTurntaking2025, mahadevanImageInThatManipulatingImages2025, ferriniPerceptsSemanticsMultimodala, axelssonYouFollow2023, choARECADesignSpeculation2023, pereiraMultimodalUserEnjoyment2024a, leusmannInvestigatingLLMDrivenCuriosity2025, xuExploringUseRobots2025a, ikedaMARCERMultimodalAugmented2025a, itoRobotDynamicallyAsking2025a, aliInclusiveCocreativeChildrobot2025a, kamelabadComparingMonolingualBilingual2025, reimannWhatCanYou2025a, pinedaSeeYouLater2025a, spitaleVITAMultiModalLLMBased2025a, limaPromotingCognitiveHealth2025} \\
Integrated Visual-Language Reasoning          & 16 & \cite{loMemoryRobotDesign2025, aliInclusiveCocreativeChildrobot2025a, leusmannInvestigatingLLMDrivenCuriosity2025, panACKnowledgeComputationalFramework2025, arjmandEmpathicGroundingExplorations2024, mahadevanImageInThatManipulatingImages2025, vermaTheoryMindAbilities2024, latifPhysicsAssistantLLMpoweredInteractive2024, bastinGPTAllySafetyorientedSystem2025, wangPepperPoseFullbodyPose2024, geGenComUIExploringGenerative2025, hoSETPAiREdDesigningParental2025, grassiGroundingConversationalRobots2024, loLLMbasedRobotPersonality2025a, bassiounyUJIButlerSymbolicNonsymbolic2025, grassiEnhancingLLMBasedHumanRobot2024} \\[0.6em]

\multicolumn{3}{l}{\textit{\textbf{Human-Oriented Understanding}}} \\[0.3em]
Emotional Grounding                           & 18 & \cite{limaPromotingCognitiveHealth2025, spitaleVITAMultiModalLLMBased2025a, hsuBittersweetSnapshotsLife2025a, bannaWordsIntegratingPersonality2025, karliAlchemistLLMAidedEndUser2024a, arjmandEmpathicGroundingExplorations2024, pereiraMultimodalUserEnjoyment2024a, mannavaExploringSuitabilityConversational2024, sieversInteractingSentimentalRobot2024, yuImprovingPerceivedEmotional2024, bastinGPTAllySafetyorientedSystem2025, choLivingAlongsideAreca2025, blancoAIenhancedSocialRobots2024, shenSocialRobotsSocial2025, loLLMbasedRobotPersonality2025a, nardelliIntuitiveInteractionCognitive2025, starkDobbyConversationalService2024, grassiEnhancingLLMBasedHumanRobot2024} \\
Task Intent Formulation                       & 39 & \cite{hsuBittersweetSnapshotsLife2025a, taoLAMSLLMDrivenAutomatic2025a, loMemoryRobotDesign2025, pinedaSeeYouLater2025a, reimannWhatCanYou2025a, kamelabadComparingMonolingualBilingual2025, aliInclusiveCocreativeChildrobot2025a, zhangWalkExperimentControlling2025, wangChildRobotRelationalNorm2025a, ikedaMARCERMultimodalAugmented2025a, xuExploringUseRobots2025a, stampfExploringPassengerAutomatedVehicle2024, panACKnowledgeComputationalFramework2025, huDesigningTelepresenceRobots2025, goubardCognitiveModellingVisual, mahadevanImageInThatManipulatingImages2025, kontogiorgosQuestioningRobotUsing2025, vermaTheoryMindAbilities2024, mahadevanGenerativeExpressiveRobot2024, sieversIntroducingNoteLevity2024, cuiNoRightOnline2023a, yeImprovedTrustHumanRobot2023, bastinGPTAllySafetyorientedSystem2025, elfleetInvestigatingImpactMultimodal2024, kimUnderstandingLargelanguageModel2024d, padmanabhaVoicePilotHarnessingLlms2024, salemComparativeHumanrobotInteraction2024, liStargazerInteractiveCamera2023, wangCrowdBotOpenenvironmentRobot2024, geGenComUIExploringGenerative2025, zuLanguageSketchingLLMdriven2024, farooqDAIMHRINewHumanRobot2024, tsushimaTaskPlanningFactory2025, loLLMbasedRobotPersonality2025a, grassiStrategiesControllingConversation2025, dellannaSONARAdaptiveControl2024, bassiounyUJIButlerSymbolicNonsymbolic2025, starkDobbyConversationalService2024, jinRobotGPTRobotManipulation2024} \\
Human Model Alignment                         & 19 & \cite{bannaWordsIntegratingPersonality2025, westerFacingLLMsRobot2024a, kamelabadComparingMonolingualBilingual2025, panACKnowledgeComputationalFramework2025, ferriniPerceptsSemanticsMultimodala, vermaTheoryMindAbilities2024, suChatAdpChatGPTpoweredAdaptation2024, zhangLargeLanguageModels2023b, bastinGPTAllySafetyorientedSystem2025, yanoUnifiedUnderstandingEnvironment2024, geGenComUIExploringGenerative2025, hoSETPAiREdDesigningParental2025, laiNaturalMultimodalFusionBased2025, tsushimaTaskPlanningFactory2025, grassiStrategiesControllingConversation2025, dellannaSONARAdaptiveControl2024, nardelliIntuitiveInteractionCognitive2025, sakamotoEffectivenessConversationalRobots2025, jinRobotGPTRobotManipulation2024} \\

\bottomrule
\end{tabular}
\end{table}

\begin{table}[H]
\centering
\small
\caption{References of Section 4.2. Generative and Agentic Interaction}
\Description{References for the generative and agentic interaction dimension in LLM-driven HRI (corresponding to Section 5.2). The table consists of three columns: Interaction (hierarchical classification of generative and agentic interaction dimensions), Number (count of relevant literature supporting each dimension), and Papers (reference numbers of the included literature). The core interaction dimensions are divided into three major categories: 1) Generative Social Communication, including two subcategories—Persona Adaptation and Conversational Fluidity (40 papers, referenced as [5,6,11,12,16,17,24,32,48,49,53,56,57,67-69,75,77,81,87,92,95,96,103,112,114,124,127,141,142,144,145,158,162,168,170,175,178,182,183]) and Embodied Social Expressiveness (22 papers, referenced as [4-6,12,16,34,54,57,75,81,87,92,93,103,115,124,142,158,175,180,183,188]); 2) Collaborative Task Co-Creation, including two subcategories—Task-Oriented Planning and Execution (28 papers, referenced as [9,14,16,17,27,36,42,46,54,59,63,65,68,75,76,81,92,94,107,109,145,153,155,164,170,178,184,197]) and Creative Storytelling and Social Engagement (13 papers, referenced as [4,5,9,12,17,24,35,54,56,57,69,95,133]); 3) Proactive Agency, including two subcategories—Social Initiation (24 papers, referenced as [12,14,25,32,47,48,53,56-58,81,96,114,121,126,133,140,142,145,148,155,168,176,183]) and Anticipatory Assistance (16 papers, referenced as [14,15,63,67,81,87,92,94,103,115,136,145,148,168,176,197]). The table systematically collates literature supporting each sub-dimension of generative and agentic interaction, providing a reference for relevant research in LLM-driven HRI.}
\begin{tabular}{p{0.30\linewidth} p{0.10\linewidth} p{0.55\linewidth}}
    \toprule
    \textbf{Interaction} & \textbf{Number} & \textbf{Papers} \\
    \midrule

\multicolumn{3}{l}{\textbf{\textit{Generative Social Communication}}} \\[0.3em]
Persona Adaptation and Conversational Fluidity      & 40 & \cite{limaPromotingCognitiveHealth2025, zhangExploringRobotPersonality2025a, hsuBittersweetSnapshotsLife2025a, bannaWordsIntegratingPersonality2025, westerFacingLLMsRobot2024a, pinedaSeeYouLater2025a, kamelabadComparingMonolingualBilingual2025, karliAlchemistLLMAidedEndUser2024a, wangChildRobotRelationalNorm2025a, xuExploringUseRobots2025a, leusmannInvestigatingLLMDrivenCuriosity2025, zhangBalancingUserControl2025a, antonyXpressSystemDynamic2025a, stampfExploringPassengerAutomatedVehicle2024, arjmandEmpathicGroundingExplorations2024, pereiraMultimodalUserEnjoyment2024a, axelssonOhSorryThink2024, hsuResearchCareReflection2025, choARECADesignSpeculation2023, kontogiorgosQuestioningRobotUsing2025, bellucciImmersiveTailoringEmbodied2025, skantzeApplyingGeneralTurntaking2025, mannavaExploringSuitabilityConversational2024, sieversIntroducingNoteLevity2024, yeImprovedTrustHumanRobot2023, latifPhysicsAssistantLLMpoweredInteractive2024, kimUnderstandingLargelanguageModel2024d, wangChallengesAdoptingCompanion2025, blancoAIenhancedSocialRobots2024, salemComparativeHumanrobotInteraction2024, loLLMbasedRobotPersonality2025a, grassiStrategiesControllingConversation2025, herathFirstImpressionsHumanoid2025, dellannaSONARAdaptiveControl2024, nardelliIntuitiveInteractionCognitive2025, rosenPreviousExperienceMatters2024, starkDobbyConversationalService2024, malnatskyFittingHumorAgeBased2025, grassiEnhancingLLMBasedHumanRobot2024, wilcockErrRoboticEarn2023a} \\
Embodied Social Expressiveness  & 22 & \cite{limaPromotingCognitiveHealth2025, hsuBittersweetSnapshotsLife2025a, bannaWordsIntegratingPersonality2025, aliInclusiveCocreativeChildrobot2025a, wangChildRobotRelationalNorm2025a, xuExploringUseRobots2025a, leusmannInvestigatingLLMDrivenCuriosity2025, zhangBalancingUserControl2025a, antonyXpressSystemDynamic2025a, arjmandEmpathicGroundingExplorations2024, kontogiorgosQuestioningRobotUsing2025, bellucciImmersiveTailoringEmbodied2025, mahadevanGenerativeExpressiveRobot2024, skantzeApplyingGeneralTurntaking2025, yuImprovingPerceivedEmotional2024, pintoPredictiveTurntakingLeveraging2024, zhangPromptingEmbodiedAI2025, elfleetInvestigatingImpactMultimodal2024, hoSETPAiREdDesigningParental2025, loLLMbasedRobotPersonality2025a, nardelliIntuitiveInteractionCognitive2025, rosenPreviousExperienceMatters2024} \\[0.6em]

\multicolumn{3}{l}{\textit{\textbf{Collaborative Task Co-Creation}}} \\[0.3em]

Task-Oriented Planning and Execution   & 28 & \cite{taoLAMSLLMDrivenAutomatic2025a, itoRobotDynamicallyAsking2025a, karliAlchemistLLMAidedEndUser2024a, ikedaMARCERMultimodalAugmented2025a, leusmannInvestigatingLLMDrivenCuriosity2025, panACKnowledgeComputationalFramework2025, goubardCognitiveModellingVisual, axelssonYouFollow2023, mahadevanImageInThatManipulatingImages2025, kontogiorgosQuestioningRobotUsing2025, bellucciImmersiveTailoringEmbodied2025, cuiNoRightOnline2023a, yeImprovedTrustHumanRobot2023, zhangLargeLanguageModels2023b, padmanabhaVoicePilotHarnessingLlms2024, blancoAIenhancedSocialRobots2024, wangCrowdBotOpenenvironmentRobot2024, geGenComUIExploringGenerative2025, hoSETPAiREdDesigningParental2025, laiNaturalMultimodalFusionBased2025, zuLanguageSketchingLLMdriven2024, farooqDAIMHRINewHumanRobot2024, tsushimaTaskPlanningFactory2025, loLLMbasedRobotPersonality2025a, bassiounyUJIButlerSymbolicNonsymbolic2025, starkDobbyConversationalService2024, jinRobotGPTRobotManipulation2024, wilcockErrRoboticEarn2023a} \\

Creative Storytelling and Social Engagement          & 13 & \cite{hsuBittersweetSnapshotsLife2025a, bannaWordsIntegratingPersonality2025, aliInclusiveCocreativeChildrobot2025a, antonyXpressSystemDynamic2025a, hsuResearchCareReflection2025, choARECADesignSpeculation2023, axelssonYouFollow2023, elgarfCreativeBotCreativeStoryteller2022, kimUnderstandingLargelanguageModel2024d, blancoAIenhancedSocialRobots2024, hoSETPAiREdDesigningParental2025, shenSocialRobotsSocial2025, malnatskyFittingHumorAgeBased2025} \\[0.6em]

\multicolumn{3}{l}{\textit{\textbf{Proactive Agency}}} \\[0.3em]

Social Initiation   & 24 & \cite{hsuBittersweetSnapshotsLife2025a, bannaWordsIntegratingPersonality2025, westerFacingLLMsRobot2024a, pinedaSeeYouLater2025a, reimannWhatCanYou2025a, leusmannInvestigatingLLMDrivenCuriosity2025, zhangBalancingUserControl2025a, huDesigningTelepresenceRobots2025, hsuResearchCareReflection2025, skantzeApplyingGeneralTurntaking2025, mannavaExploringSuitabilityConversational2024, sieversInteractingSentimentalRobot2024, suChatAdpChatGPTpoweredAdaptation2024, yanoUnifiedUnderstandingEnvironment2024, choLivingAlongsideAreca2025, grassiGroundingConversationalRobots2024, tsushimaTaskPlanningFactory2025, shenSocialRobotsSocial2025, herathFirstImpressionsHumanoid2025, dellannaSONARAdaptiveControl2024, bassiounyUJIButlerSymbolicNonsymbolic2025, sakamotoEffectivenessConversationalRobots2025, starkDobbyConversationalService2024, grassiEnhancingLLMBasedHumanRobot2024} \\

Anticipatory Assistance          & 16 & \cite{limaPromotingCognitiveHealth2025, westerFacingLLMsRobot2024a, kamelabadComparingMonolingualBilingual2025, itoRobotDynamicallyAsking2025a, leusmannInvestigatingLLMDrivenCuriosity2025, mahadevanImageInThatManipulatingImages2025, suChatAdpChatGPTpoweredAdaptation2024, pintoPredictiveTurntakingLeveraging2024, bastinGPTAllySafetyorientedSystem2025, yanoUnifiedUnderstandingEnvironment2024, shiradoRealismDrivesInterpersonal2025, zuLanguageSketchingLLMdriven2024, loLLMbasedRobotPersonality2025a, nardelliIntuitiveInteractionCognitive2025, bassiounyUJIButlerSymbolicNonsymbolic2025, starkDobbyConversationalService2024} \\

\bottomrule
\end{tabular}
\end{table}

\begin{table}[h!]
\centering
\small
\caption{References of Section 4.3. Iterative Optimization and Alignment}
\Description{References for the iterative optimization and alignment dimension in LLM-driven HRI (corresponding to Section 5.3). The table comprises three columns: Alignment (hierarchical classification of iterative optimization and alignment dimensions), Number (count of relevant supporting literature), and Papers (reference numbers of included literature). The core dimensions are divided into two major categories: 1) Longitudinal Personalization and Memory, including two subcategories—Sustained Personalization (15 papers, referenced as [9, 12, 14, 15, 56, 81, 91, 92, 96, 133, 143, 145, 148, 153, 162]) and Episodic Memory Integration (22 papers, referenced as [14, 17, 48, 53, 56, 67, 77, 91, 92, 96, 103, 107, 109, 114, 124, 133, 140, 141, 145, 153, 178, 184]); 2) Multi-Level Repair, including three subcategories—Behavioral Repair in Task Execution (13 papers, referenced as [9, 14, 27, 35, 36, 65, 75–77, 81, 93, 121, 182]), Emotional Repair in Social Interaction (10 papers, referenced as [4, 11, 12, 56, 57, 73, 93, 96, 103, 175]), and Repair in Ethical and Normative Alignment (10 papers, referenced as [32, 83, 87, 93, 96, 112, 143, 144, 155, 188]). The table systematically collates literature supporting each sub-dimension of iterative optimization and alignment, providing a foundational reference for research on long-term adaptation and repair mechanisms in LLM-driven HRI.}
\begin{tabular}{p{0.30\linewidth} p{0.10\linewidth} p{0.55\linewidth}}
    \toprule
    \textbf{Alignment} & \textbf{Number} & \textbf{Papers} \\
    \midrule 

\multicolumn{3}{l}{\textit{\textbf{Longitudinal Personalization and Memory}}} \\[0.3em]
Sustained Personalization      & 15 & \cite{spitaleVITAMultiModalLLMBased2025a, taoLAMSLLMDrivenAutomatic2025a, bannaWordsIntegratingPersonality2025, loMemoryRobotDesign2025, leusmannInvestigatingLLMDrivenCuriosity2025, hsuResearchCareReflection2025, axelssonYouFollow2023, mannavaExploringSuitabilityConversational2024, suChatAdpChatGPTpoweredAdaptation2024, bastinGPTAllySafetyorientedSystem2025, wangChallengesAdoptingCompanion2025, shenSocialRobotsSocial2025, loLLMbasedRobotPersonality2025a, bassiounyUJIButlerSymbolicNonsymbolic2025, starkDobbyConversationalService2024} \\

Episodic Memory Integration  & 22 & \cite{taoLAMSLLMDrivenAutomatic2025a, loMemoryRobotDesign2025, pinedaSeeYouLater2025a, kamelabadComparingMonolingualBilingual2025, panACKnowledgeComputationalFramework2025, hsuResearchCareReflection2025, mannavaExploringSuitabilityConversational2024, sieversIntroducingNoteLevity2024, sieversInteractingSentimentalRobot2024, yeImprovedTrustHumanRobot2023, zhangLargeLanguageModels2023b, latifPhysicsAssistantLLMpoweredInteractive2024, padmanabhaVoicePilotHarnessingLlms2024, blancoAIenhancedSocialRobots2024, shenSocialRobotsSocial2025, loLLMbasedRobotPersonality2025a, herathFirstImpressionsHumanoid2025, nardelliIntuitiveInteractionCognitive2025, bassiounyUJIButlerSymbolicNonsymbolic2025, rosenPreviousExperienceMatters2024, starkDobbyConversationalService2024, grassiEnhancingLLMBasedHumanRobot2024} \\[0.6em]

\multicolumn{3}{l}{\textit{\textbf{Multi-Level Repair}}} \\[0.3em]
Behavioral Repair in Task Execution   & 13 & \cite{zhangExploringRobotPersonality2025a, reimannWhatCanYou2025a, leusmannInvestigatingLLMDrivenCuriosity2025, axelssonYouFollow2023, kontogiorgosQuestioningRobotUsing2025, mahadevanGenerativeExpressiveRobot2024, cuiNoRightOnline2023a, latifPhysicsAssistantLLMpoweredInteractive2024, elgarfCreativeBotCreativeStoryteller2022, laiNaturalMultimodalFusionBased2025, farooqDAIMHRINewHumanRobot2024, bassiounyUJIButlerSymbolicNonsymbolic2025, jinRobotGPTRobotManipulation2024} \\

Emotional Repair in Social Interaction          & 10 & \cite{hsuBittersweetSnapshotsLife2025a, bannaWordsIntegratingPersonality2025, aliInclusiveCocreativeChildrobot2025a, xuExploringUseRobots2025a, axelssonOhSorryThink2024, hsuResearchCareReflection2025, mahadevanGenerativeExpressiveRobot2024, mannavaExploringSuitabilityConversational2024, nardelliIntuitiveInteractionCognitive2025, kodurExploringDynamicsHumanRobot2025} \\

Repair in Ethical and Normative Alignment         & 10 & \cite{limaPromotingCognitiveHealth2025, spitaleVITAMultiModalLLMBased2025a, stampfExploringPassengerAutomatedVehicle2024, pereiraMultimodalUserEnjoyment2024a, mahadevanGenerativeExpressiveRobot2024, mannavaExploringSuitabilityConversational2024, zhangPromptingEmbodiedAI2025, liStargazerInteractiveCamera2023, tsushimaTaskPlanningFactory2025, dellannaSONARAdaptiveControl2024} \\

\bottomrule
\end{tabular}
\end{table}

\begin{table}[h!]
\centering
\small
\caption{References of Section 5. Design Components}
\Description{References for design components in LLM-driven HRI (corresponding to Section 6.1). The table is structured with three columns: Components (hierarchical classification of core design components), Number (count of relevant supporting literature), and Papers (reference numbers of included literature). The design components are divided into three major categories: 1) Modality and Interaction Channels, including seven subcategories—Text (53 papers), Voice (71 papers), Visuals (56 papers), Motion (52 papers), Hybrid (53 papers), Tangible and Haptic Interaction (9 papers), and Proximity (13 papers), each listing corresponding literature references; 2) Robot Morphology (noting the typo "Mophology" in the original table), including five subcategories—Humanoid (39 papers), Functional (31 papers), Zoomorphic (2 papers), Desktop Companions (9 papers), and VR/AR-based (5 papers); 3) Levels of Autonomy, including three subcategories—Full Autonomy (46 papers), Semi-Autonomy (37 papers), and Teleoperation (3 papers). The table systematically collates literature supporting each design component and its subcategories, providing a comprehensive reference for the selection and design of interaction modalities, robot forms, and autonomy levels in LLM-driven HRI research.}
\begin{tabular}{p{0.15\linewidth} p{0.10\linewidth} p{0.70\linewidth}}
    \toprule
    \textbf{Components} & \textbf{Number} & \textbf{Papers} \\
    \midrule 

\multicolumn{3}{l}{\textit{\textbf{Modality and Interaction Channels}}} \\[0.3em]
Text      & 53 & \cite{limaPromotingCognitiveHealth2025, loMemoryRobotDesign2025, westerFacingLLMsRobot2024a, itoRobotDynamicallyAsking2025a, karliAlchemistLLMAidedEndUser2024a, wangChildRobotRelationalNorm2025a, ikedaMARCERMultimodalAugmented2025a, zhangBalancingUserControl2025a, antonyXpressSystemDynamic2025a, panACKnowledgeComputationalFramework2025, arjmandEmpathicGroundingExplorations2024, pereiraMultimodalUserEnjoyment2024a, huDesigningTelepresenceRobots2025, goubardCognitiveModellingVisual, axelssonYouFollow2023, mahadevanImageInThatManipulatingImages2025, kontogiorgosQuestioningRobotUsing2025, bellucciImmersiveTailoringEmbodied2025, vermaTheoryMindAbilities2024, mahadevanGenerativeExpressiveRobot2024, mannavaExploringSuitabilityConversational2024, sieversInteractingSentimentalRobot2024, cuiNoRightOnline2023a, yeImprovedTrustHumanRobot2023, suChatAdpChatGPTpoweredAdaptation2024, zhangLargeLanguageModels2023b, latifPhysicsAssistantLLMpoweredInteractive2024, yuImprovingPerceivedEmotional2024, pintoPredictiveTurntakingLeveraging2024, bastinGPTAllySafetyorientedSystem2025, yanoUnifiedUnderstandingEnvironment2024, shiradoRealismDrivesInterpersonal2025, kimUnderstandingLargelanguageModel2024d, choLivingAlongsideAreca2025, padmanabhaVoicePilotHarnessingLlms2024, wangCrowdBotOpenenvironmentRobot2024, hoSETPAiREdDesigningParental2025, zuLanguageSketchingLLMdriven2024, farooqDAIMHRINewHumanRobot2024, grassiGroundingConversationalRobots2024, tsushimaTaskPlanningFactory2025, shenSocialRobotsSocial2025, loLLMbasedRobotPersonality2025a, herathFirstImpressionsHumanoid2025, dellannaSONARAdaptiveControl2024, nardelliIntuitiveInteractionCognitive2025, sakamotoEffectivenessConversationalRobots2025, rosenPreviousExperienceMatters2024, kodurExploringDynamicsHumanRobot2025, starkDobbyConversationalService2024, malnatskyFittingHumorAgeBased2025, jinRobotGPTRobotManipulation2024, wilcockErrRoboticEarn2023a} \\

Voice   & 71 & \cite{limaPromotingCognitiveHealth2025, zhangExploringRobotPersonality2025a, spitaleVITAMultiModalLLMBased2025a, hsuBittersweetSnapshotsLife2025a, bannaWordsIntegratingPersonality2025, loMemoryRobotDesign2025, westerFacingLLMsRobot2024a, pinedaSeeYouLater2025a, reimannWhatCanYou2025a, kamelabadComparingMonolingualBilingual2025,aliInclusiveCocreativeChildrobot2025a,zhangWalkExperimentControlling2025, itoRobotDynamicallyAsking2025a, karliAlchemistLLMAidedEndUser2024a, wangChildRobotRelationalNorm2025a, ikedaMARCERMultimodalAugmented2025a, xuExploringUseRobots2025a, leusmannInvestigatingLLMDrivenCuriosity2025, zhangBalancingUserControl2025a, antonyXpressSystemDynamic2025a, stampfExploringPassengerAutomatedVehicle2024, arjmandEmpathicGroundingExplorations2024, pereiraMultimodalUserEnjoyment2024a, huDesigningTelepresenceRobots2025, axelssonOhSorryThink2024, hsuResearchCareReflection2025, axelssonYouFollow2023, kontogiorgosQuestioningRobotUsing2025, bellucciImmersiveTailoringEmbodied2025, mahadevanGenerativeExpressiveRobot2024, skantzeApplyingGeneralTurntaking2025, mannavaExploringSuitabilityConversational2024, sieversInteractingSentimentalRobot2024, yeImprovedTrustHumanRobot2023, suChatAdpChatGPTpoweredAdaptation2024, latifPhysicsAssistantLLMpoweredInteractive2024, yuImprovingPerceivedEmotional2024, pintoPredictiveTurntakingLeveraging2024, bastinGPTAllySafetyorientedSystem2025, yanoUnifiedUnderstandingEnvironment2024, zhangPromptingEmbodiedAI2025, elfleetInvestigatingImpactMultimodal2024, perella-holfeldParentEducatorConcerns2024a, elgarfCreativeBotCreativeStoryteller2022, kimUnderstandingLargelanguageModel2024d, wangChallengesAdoptingCompanion2025, padmanabhaVoicePilotHarnessingLlms2024, blancoAIenhancedSocialRobots2024, salemComparativeHumanrobotInteraction2024, liStargazerInteractiveCamera2023, wangCrowdBotOpenenvironmentRobot2024, geGenComUIExploringGenerative2025, hoSETPAiREdDesigningParental2025, laiNaturalMultimodalFusionBased2025, zuLanguageSketchingLLMdriven2024, farooqDAIMHRINewHumanRobot2024, grassiGroundingConversationalRobots2024, shenSocialRobotsSocial2025, loLLMbasedRobotPersonality2025a, grassiStrategiesControllingConversation2025, herathFirstImpressionsHumanoid2025, dellannaSONARAdaptiveControl2024, nardelliIntuitiveInteractionCognitive2025, bassiounyUJIButlerSymbolicNonsymbolic2025, sakamotoEffectivenessConversationalRobots2025, rosenPreviousExperienceMatters2024, kodurExploringDynamicsHumanRobot2025, starkDobbyConversationalService2024, malnatskyFittingHumorAgeBased2025, grassiEnhancingLLMBasedHumanRobot2024, wilcockErrRoboticEarn2023a} \\

Visuals   & 56 & \cite{limaPromotingCognitiveHealth2025, zhangExploringRobotPersonality2025a, spitaleVITAMultiModalLLMBased2025a, taoLAMSLLMDrivenAutomatic2025a, zhangWalkExperimentControlling2025, itoRobotDynamicallyAsking2025a, karliAlchemistLLMAidedEndUser2024a, ikedaMARCERMultimodalAugmented2025a, xuExploringUseRobots2025a, antonyXpressSystemDynamic2025a, stampfExploringPassengerAutomatedVehicle2024, panACKnowledgeComputationalFramework2025, arjmandEmpathicGroundingExplorations2024, pereiraMultimodalUserEnjoyment2024a, goubardCognitiveModellingVisual, hsuResearchCareReflection2025, choARECADesignSpeculation2023, axelssonYouFollow2023, ferriniPerceptsSemanticsMultimodala, mahadevanImageInThatManipulatingImages2025, kontogiorgosQuestioningRobotUsing2025, bellucciImmersiveTailoringEmbodied2025, vermaTheoryMindAbilities2024, mahadevanGenerativeExpressiveRobot2024, skantzeApplyingGeneralTurntaking2025, mannavaExploringSuitabilityConversational2024, sieversIntroducingNoteLevity2024, sieversInteractingSentimentalRobot2024, cuiNoRightOnline2023a, suChatAdpChatGPTpoweredAdaptation2024, yuImprovingPerceivedEmotional2024, yanoUnifiedUnderstandingEnvironment2024, zhangPromptingEmbodiedAI2025, shiradoRealismDrivesInterpersonal2025, elfleetInvestigatingImpactMultimodal2024, perella-holfeldParentEducatorConcerns2024a, elgarfCreativeBotCreativeStoryteller2022, blancoAIenhancedSocialRobots2024, salemComparativeHumanrobotInteraction2024, liStargazerInteractiveCamera2023, wangCrowdBotOpenenvironmentRobot2024, wangPepperPoseFullbodyPose2024, geGenComUIExploringGenerative2025, hoSETPAiREdDesigningParental2025, zuLanguageSketchingLLMdriven2024, grassiGroundingConversationalRobots2024, tsushimaTaskPlanningFactory2025, shenSocialRobotsSocial2025, loLLMbasedRobotPersonality2025a, dellannaSONARAdaptiveControl2024, nardelliIntuitiveInteractionCognitive2025, bassiounyUJIButlerSymbolicNonsymbolic2025, starkDobbyConversationalService2024, grassiEnhancingLLMBasedHumanRobot2024, jinRobotGPTRobotManipulation2024, wilcockErrRoboticEarn2023a} \\

Motion   & 52 & \cite{limaPromotingCognitiveHealth2025, hsuBittersweetSnapshotsLife2025a, taoLAMSLLMDrivenAutomatic2025a, loMemoryRobotDesign2025, pinedaSeeYouLater2025a, reimannWhatCanYou2025a, aliInclusiveCocreativeChildrobot2025a, ikedaMARCERMultimodalAugmented2025a, xuExploringUseRobots2025a, leusmannInvestigatingLLMDrivenCuriosity2025, panACKnowledgeComputationalFramework2025, arjmandEmpathicGroundingExplorations2024, huDesigningTelepresenceRobots2025, goubardCognitiveModellingVisual, axelssonOhSorryThink2024, axelssonYouFollow2023, ferriniPerceptsSemanticsMultimodala, mahadevanImageInThatManipulatingImages2025, kontogiorgosQuestioningRobotUsing2025, bellucciImmersiveTailoringEmbodied2025, vermaTheoryMindAbilities2024, mahadevanGenerativeExpressiveRobot2024, sieversIntroducingNoteLevity2024, sieversInteractingSentimentalRobot2024, cuiNoRightOnline2023a, yeImprovedTrustHumanRobot2023, yuImprovingPerceivedEmotional2024, zhangPromptingEmbodiedAI2025, shiradoRealismDrivesInterpersonal2025, elfleetInvestigatingImpactMultimodal2024, kimUnderstandingLargelanguageModel2024d, choLivingAlongsideAreca2025, padmanabhaVoicePilotHarnessingLlms2024, blancoAIenhancedSocialRobots2024, liStargazerInteractiveCamera2023, wangCrowdBotOpenenvironmentRobot2024, wangPepperPoseFullbodyPose2024, hoSETPAiREdDesigningParental2025, laiNaturalMultimodalFusionBased2025, tsushimaTaskPlanningFactory2025, shenSocialRobotsSocial2025, loLLMbasedRobotPersonality2025a, herathFirstImpressionsHumanoid2025, dellannaSONARAdaptiveControl2024, nardelliIntuitiveInteractionCognitive2025, bassiounyUJIButlerSymbolicNonsymbolic2025, sakamotoEffectivenessConversationalRobots2025, rosenPreviousExperienceMatters2024, kodurExploringDynamicsHumanRobot2025, starkDobbyConversationalService2024, malnatskyFittingHumorAgeBased2025, jinRobotGPTRobotManipulation2024} \\

Hybrid   & 53 & \cite{limaPromotingCognitiveHealth2025, loMemoryRobotDesign2025, pinedaSeeYouLater2025a, kamelabadComparingMonolingualBilingual2025, aliInclusiveCocreativeChildrobot2025a, zhangWalkExperimentControlling2025, itoRobotDynamicallyAsking2025a, karliAlchemistLLMAidedEndUser2024a, wangChildRobotRelationalNorm2025a, ikedaMARCERMultimodalAugmented2025a, leusmannInvestigatingLLMDrivenCuriosity2025, antonyXpressSystemDynamic2025a, arjmandEmpathicGroundingExplorations2024, pereiraMultimodalUserEnjoyment2024a, hsuResearchCareReflection2025, axelssonYouFollow2023, ferriniPerceptsSemanticsMultimodala, mahadevanImageInThatManipulatingImages2025, kontogiorgosQuestioningRobotUsing2025, bellucciImmersiveTailoringEmbodied2025, sieversInteractingSentimentalRobot2024, yeImprovedTrustHumanRobot2023, suChatAdpChatGPTpoweredAdaptation2024, latifPhysicsAssistantLLMpoweredInteractive2024, pintoPredictiveTurntakingLeveraging2024, bastinGPTAllySafetyorientedSystem2025, zhangPromptingEmbodiedAI2025, elfleetInvestigatingImpactMultimodal2024, elgarfCreativeBotCreativeStoryteller2022, kimUnderstandingLargelanguageModel2024d, wangChallengesAdoptingCompanion2025, blancoAIenhancedSocialRobots2024, salemComparativeHumanrobotInteraction2024, liStargazerInteractiveCamera2023, wangCrowdBotOpenenvironmentRobot2024, wangPepperPoseFullbodyPose2024, geGenComUIExploringGenerative2025, hoSETPAiREdDesigningParental2025, laiNaturalMultimodalFusionBased2025, zuLanguageSketchingLLMdriven2024, farooqDAIMHRINewHumanRobot2024, grassiGroundingConversationalRobots2024, tsushimaTaskPlanningFactory2025, shenSocialRobotsSocial2025, loLLMbasedRobotPersonality2025a, herathFirstImpressionsHumanoid2025, dellannaSONARAdaptiveControl2024, nardelliIntuitiveInteractionCognitive2025, rosenPreviousExperienceMatters2024, starkDobbyConversationalService2024, malnatskyFittingHumorAgeBased2025, grassiEnhancingLLMBasedHumanRobot2024, jinRobotGPTRobotManipulation2024} \\

Tangible and Haptic Interaction   & 9 & \cite{taoLAMSLLMDrivenAutomatic2025a, zhangBalancingUserControl2025a, choARECADesignSpeculation2023, sieversInteractingSentimentalRobot2024, elgarfCreativeBotCreativeStoryteller2022, choLivingAlongsideAreca2025, wangChallengesAdoptingCompanion2025, hoSETPAiREdDesigningParental2025, bassiounyUJIButlerSymbolicNonsymbolic2025} \\

Proximity   & 13 & \cite{pinedaSeeYouLater2025a, kontogiorgosQuestioningRobotUsing2025, sieversIntroducingNoteLevity2024, zhangLargeLanguageModels2023b, zhangPromptingEmbodiedAI2025, shiradoRealismDrivesInterpersonal2025, wangPepperPoseFullbodyPose2024, laiNaturalMultimodalFusionBased2025, farooqDAIMHRINewHumanRobot2024, grassiStrategiesControllingConversation2025, dellannaSONARAdaptiveControl2024, nardelliIntuitiveInteractionCognitive2025, starkDobbyConversationalService2024} \\[0.6em]

\multicolumn{3}{l}{\textit{\textbf{Robot Mophology}}} \\[0.3em]
Humanoid   & 39 & \cite{zhangExploringRobotPersonality2025a, spitaleVITAMultiModalLLMBased2025a, hsuBittersweetSnapshotsLife2025a, bannaWordsIntegratingPersonality2025, loMemoryRobotDesign2025, westerFacingLLMsRobot2024a, reimannWhatCanYou2025a,kamelabadComparingMonolingualBilingual2025, itoRobotDynamicallyAsking2025a, wangChildRobotRelationalNorm2025a, leusmannInvestigatingLLMDrivenCuriosity2025, antonyXpressSystemDynamic2025a, arjmandEmpathicGroundingExplorations2024, pereiraMultimodalUserEnjoyment2024a, axelssonOhSorryThink2024, hsuResearchCareReflection2025, axelssonYouFollow2023, ferriniPerceptsSemanticsMultimodala, skantzeApplyingGeneralTurntaking2025, mannavaExploringSuitabilityConversational2024, sieversIntroducingNoteLevity2024, sieversInteractingSentimentalRobot2024, latifPhysicsAssistantLLMpoweredInteractive2024, pintoPredictiveTurntakingLeveraging2024, elgarfCreativeBotCreativeStoryteller2022, kimUnderstandingLargelanguageModel2024d, salemComparativeHumanrobotInteraction2024, wangPepperPoseFullbodyPose2024, grassiGroundingConversationalRobots2024, loLLMbasedRobotPersonality2025a, grassiStrategiesControllingConversation2025, herathFirstImpressionsHumanoid2025, dellannaSONARAdaptiveControl2024, nardelliIntuitiveInteractionCognitive2025, sakamotoEffectivenessConversationalRobots2025, rosenPreviousExperienceMatters2024, malnatskyFittingHumorAgeBased2025, grassiEnhancingLLMBasedHumanRobot2024, wilcockErrRoboticEarn2023a} \\

Functional         & 31 & \cite{limaPromotingCognitiveHealth2025, taoLAMSLLMDrivenAutomatic2025a, pinedaSeeYouLater2025a, karliAlchemistLLMAidedEndUser2024a, ikedaMARCERMultimodalAugmented2025a, zhangBalancingUserControl2025a, huDesigningTelepresenceRobots2025, goubardCognitiveModellingVisual, choARECADesignSpeculation2023, mahadevanImageInThatManipulatingImages2025, kontogiorgosQuestioningRobotUsing2025, vermaTheoryMindAbilities2024, mahadevanGenerativeExpressiveRobot2024, cuiNoRightOnline2023a, yeImprovedTrustHumanRobot2023, suChatAdpChatGPTpoweredAdaptation2024, zhangLargeLanguageModels2023b, bastinGPTAllySafetyorientedSystem2025, yanoUnifiedUnderstandingEnvironment2024, choLivingAlongsideAreca2025, padmanabhaVoicePilotHarnessingLlms2024, liStargazerInteractiveCamera2023, geGenComUIExploringGenerative2025, laiNaturalMultimodalFusionBased2025, zuLanguageSketchingLLMdriven2024, farooqDAIMHRINewHumanRobot2024, tsushimaTaskPlanningFactory2025, bassiounyUJIButlerSymbolicNonsymbolic2025, kodurExploringDynamicsHumanRobot2025, starkDobbyConversationalService2024, jinRobotGPTRobotManipulation2024} \\

Zoomorphic          & 2 & \cite{zhangWalkExperimentControlling2025, wangCrowdBotOpenenvironmentRobot2024} \\

Desktop Companions           & 9 & \cite{aliInclusiveCocreativeChildrobot2025a, xuExploringUseRobots2025a, panACKnowledgeComputationalFramework2025, yuImprovingPerceivedEmotional2024, perella-holfeldParentEducatorConcerns2024a, wangChallengesAdoptingCompanion2025, blancoAIenhancedSocialRobots2024, hoSETPAiREdDesigningParental2025, shenSocialRobotsSocial2025} \\

VR/AR-based          & 5 & \cite{stampfExploringPassengerAutomatedVehicle2024, bellucciImmersiveTailoringEmbodied2025, zhangPromptingEmbodiedAI2025, shiradoRealismDrivesInterpersonal2025, elfleetInvestigatingImpactMultimodal2024} \\[0.6em]

\multicolumn{3}{l}{\textit{\textbf{Levels of Autonomy}}} \\[0.3em]
Full Autonomy   & 46 & \cite{spitaleVITAMultiModalLLMBased2025a, bannaWordsIntegratingPersonality2025, loMemoryRobotDesign2025, westerFacingLLMsRobot2024a, pinedaSeeYouLater2025a, reimannWhatCanYou2025a, kamelabadComparingMonolingualBilingual2025, wangChildRobotRelationalNorm2025a, leusmannInvestigatingLLMDrivenCuriosity2025, antonyXpressSystemDynamic2025a, panACKnowledgeComputationalFramework2025, pereiraMultimodalUserEnjoyment2024a, axelssonOhSorryThink2024, hsuResearchCareReflection2025, choARECADesignSpeculation2023, axelssonYouFollow2023, kontogiorgosQuestioningRobotUsing2025, skantzeApplyingGeneralTurntaking2025, sieversIntroducingNoteLevity2024, sieversInteractingSentimentalRobot2024, suChatAdpChatGPTpoweredAdaptation2024, zhangLargeLanguageModels2023b, latifPhysicsAssistantLLMpoweredInteractive2024, yuImprovingPerceivedEmotional2024, bastinGPTAllySafetyorientedSystem2025, yanoUnifiedUnderstandingEnvironment2024, shiradoRealismDrivesInterpersonal2025, elfleetInvestigatingImpactMultimodal2024, elgarfCreativeBotCreativeStoryteller2022, kimUnderstandingLargelanguageModel2024d, wangChallengesAdoptingCompanion2025, salemComparativeHumanrobotInteraction2024, wangCrowdBotOpenenvironmentRobot2024, wangPepperPoseFullbodyPose2024, zuLanguageSketchingLLMdriven2024, grassiGroundingConversationalRobots2024, shenSocialRobotsSocial2025, loLLMbasedRobotPersonality2025a, dellannaSONARAdaptiveControl2024, nardelliIntuitiveInteractionCognitive2025, rosenPreviousExperienceMatters2024, starkDobbyConversationalService2024, malnatskyFittingHumorAgeBased2025, grassiEnhancingLLMBasedHumanRobot2024, jinRobotGPTRobotManipulation2024, wilcockErrRoboticEarn2023a} \\

Semi-Autonomy        & 37 & \cite{limaPromotingCognitiveHealth2025, zhangExploringRobotPersonality2025a, hsuBittersweetSnapshotsLife2025a, aliInclusiveCocreativeChildrobot2025a, itoRobotDynamicallyAsking2025a, karliAlchemistLLMAidedEndUser2024a, ikedaMARCERMultimodalAugmented2025a, xuExploringUseRobots2025a, zhangBalancingUserControl2025a, stampfExploringPassengerAutomatedVehicle2024, arjmandEmpathicGroundingExplorations2024, goubardCognitiveModellingVisual, ferriniPerceptsSemanticsMultimodala, mahadevanImageInThatManipulatingImages2025, bellucciImmersiveTailoringEmbodied2025, vermaTheoryMindAbilities2024, mahadevanGenerativeExpressiveRobot2024, mannavaExploringSuitabilityConversational2024, cuiNoRightOnline2023a, yeImprovedTrustHumanRobot2023, pintoPredictiveTurntakingLeveraging2024, zhangPromptingEmbodiedAI2025, perella-holfeldParentEducatorConcerns2024a, choLivingAlongsideAreca2025, padmanabhaVoicePilotHarnessingLlms2024, blancoAIenhancedSocialRobots2024, liStargazerInteractiveCamera2023, geGenComUIExploringGenerative2025, hoSETPAiREdDesigningParental2025, laiNaturalMultimodalFusionBased2025, farooqDAIMHRINewHumanRobot2024, tsushimaTaskPlanningFactory2025, grassiStrategiesControllingConversation2025, herathFirstImpressionsHumanoid2025, bassiounyUJIButlerSymbolicNonsymbolic2025, sakamotoEffectivenessConversationalRobots2025, kodurExploringDynamicsHumanRobot2025} \\

Teleoperation         & 3 & \cite{taoLAMSLLMDrivenAutomatic2025a, zhangWalkExperimentControlling2025, huDesigningTelepresenceRobots2025} \\

\bottomrule
\end{tabular}
\end{table}

\begin{table}[h!]
\centering
\small
\caption{References of Section 6.1. Methodology}
\Description{References for research methodologies in LLM-driven HRI (corresponding to Section 6.2). The table consists of three columns: Methods (classification of research methodologies adopted in LLM-driven HRI studies), Number (count of relevant supporting literature for each method), and Papers (reference numbers of included literature). The methodologies are categorized into 11 types, with specific details as follows: Laboratory Experiment (58 papers), Field Deployments (17 papers), Interviews (29 papers), Questionnaires (70 papers, the most widely used method), Technical Evaluation (52 papers), Wizard-of-Oz (8 papers), Case studies (11 papers), Simulations (14 papers), Co-design workshops (6 papers), BodyStorming (2 papers), and Think-aloud protocols (2 papers). Each category lists the reference numbers of the literature that apply the corresponding methodology, systematically collating the methodological support for LLM-driven HRI research and providing a comprehensive reference for researchers in selecting appropriate research methods.}
\begin{tabular}{p{0.15\linewidth} p{0.10\linewidth} p{0.70\linewidth}}
    \toprule
    \textbf{Methods} & \textbf{Number} & \textbf{Papers} \\
    \midrule 

Laboratory Experiment      & 58 & \cite{spitaleVITAMultiModalLLMBased2025a, bannaWordsIntegratingPersonality2025, loMemoryRobotDesign2025, pinedaSeeYouLater2025a, reimannWhatCanYou2025a, kamelabadComparingMonolingualBilingual2025, aliInclusiveCocreativeChildrobot2025a,zhangWalkExperimentControlling2025, karliAlchemistLLMAidedEndUser2024a, ikedaMARCERMultimodalAugmented2025a, leusmannInvestigatingLLMDrivenCuriosity2025, zhangBalancingUserControl2025a, antonyXpressSystemDynamic2025a, stampfExploringPassengerAutomatedVehicle2024, arjmandEmpathicGroundingExplorations2024, pereiraMultimodalUserEnjoyment2024a, goubardCognitiveModellingVisual, axelssonYouFollow2023, ferriniPerceptsSemanticsMultimodala, mahadevanImageInThatManipulatingImages2025, kontogiorgosQuestioningRobotUsing2025, bellucciImmersiveTailoringEmbodied2025, vermaTheoryMindAbilities2024, skantzeApplyingGeneralTurntaking2025, mannavaExploringSuitabilityConversational2024, sieversIntroducingNoteLevity2024, sieversInteractingSentimentalRobot2024, cuiNoRightOnline2023a, yeImprovedTrustHumanRobot2023, suChatAdpChatGPTpoweredAdaptation2024, zhangLargeLanguageModels2023b, latifPhysicsAssistantLLMpoweredInteractive2024, yuImprovingPerceivedEmotional2024, bastinGPTAllySafetyorientedSystem2025, zhangPromptingEmbodiedAI2025, shiradoRealismDrivesInterpersonal2025, elfleetInvestigatingImpactMultimodal2024, elgarfCreativeBotCreativeStoryteller2022, kimUnderstandingLargelanguageModel2024d, padmanabhaVoicePilotHarnessingLlms2024, salemComparativeHumanrobotInteraction2024, liStargazerInteractiveCamera2023, geGenComUIExploringGenerative2025, laiNaturalMultimodalFusionBased2025, zuLanguageSketchingLLMdriven2024, farooqDAIMHRINewHumanRobot2024, grassiGroundingConversationalRobots2024, tsushimaTaskPlanningFactory2025, grassiStrategiesControllingConversation2025, dellannaSONARAdaptiveControl2024, nardelliIntuitiveInteractionCognitive2025, bassiounyUJIButlerSymbolicNonsymbolic2025, rosenPreviousExperienceMatters2024, kodurExploringDynamicsHumanRobot2025, starkDobbyConversationalService2024, grassiEnhancingLLMBasedHumanRobot2024, jinRobotGPTRobotManipulation2024, wilcockErrRoboticEarn2023a} \\

Field Deployments   & 17 & \cite{spitaleVITAMultiModalLLMBased2025a, itoRobotDynamicallyAsking2025a, wangChildRobotRelationalNorm2025a, xuExploringUseRobots2025a, huDesigningTelepresenceRobots2025, hsuResearchCareReflection2025, choARECADesignSpeculation2023, pintoPredictiveTurntakingLeveraging2024, yanoUnifiedUnderstandingEnvironment2024, choLivingAlongsideAreca2025, blancoAIenhancedSocialRobots2024, wangCrowdBotOpenenvironmentRobot2024, laiNaturalMultimodalFusionBased2025, tsushimaTaskPlanningFactory2025, shenSocialRobotsSocial2025, herathFirstImpressionsHumanoid2025, grassiEnhancingLLMBasedHumanRobot2024} \\

Interviews  & 29 & \cite{limaPromotingCognitiveHealth2025, spitaleVITAMultiModalLLMBased2025a, hsuBittersweetSnapshotsLife2025a, taoLAMSLLMDrivenAutomatic2025a, westerFacingLLMsRobot2024a, pinedaSeeYouLater2025a, zhangWalkExperimentControlling2025, itoRobotDynamicallyAsking2025a, karliAlchemistLLMAidedEndUser2024a, ikedaMARCERMultimodalAugmented2025a, xuExploringUseRobots2025a, leusmannInvestigatingLLMDrivenCuriosity2025, antonyXpressSystemDynamic2025a, panACKnowledgeComputationalFramework2025, huDesigningTelepresenceRobots2025, axelssonOhSorryThink2024, hsuResearchCareReflection2025, zhangPromptingEmbodiedAI2025, elfleetInvestigatingImpactMultimodal2024, kimUnderstandingLargelanguageModel2024d, choLivingAlongsideAreca2025, wangChallengesAdoptingCompanion2025, liStargazerInteractiveCamera2023, wangPepperPoseFullbodyPose2024, geGenComUIExploringGenerative2025, hoSETPAiREdDesigningParental2025, shenSocialRobotsSocial2025, rosenPreviousExperienceMatters2024, starkDobbyConversationalService2024} \\

Questionnaires   & 70 & \cite{limaPromotingCognitiveHealth2025, zhangExploringRobotPersonality2025a, spitaleVITAMultiModalLLMBased2025a, taoLAMSLLMDrivenAutomatic2025a, bannaWordsIntegratingPersonality2025, westerFacingLLMsRobot2024a, pinedaSeeYouLater2025a, reimannWhatCanYou2025a, kamelabadComparingMonolingualBilingual2025, aliInclusiveCocreativeChildrobot2025a,zhangWalkExperimentControlling2025, karliAlchemistLLMAidedEndUser2024a, wangChildRobotRelationalNorm2025a, ikedaMARCERMultimodalAugmented2025a, xuExploringUseRobots2025a, leusmannInvestigatingLLMDrivenCuriosity2025, zhangBalancingUserControl2025a, antonyXpressSystemDynamic2025a, stampfExploringPassengerAutomatedVehicle2024, panACKnowledgeComputationalFramework2025, arjmandEmpathicGroundingExplorations2024, pereiraMultimodalUserEnjoyment2024a, goubardCognitiveModellingVisual, axelssonOhSorryThink2024, axelssonYouFollow2023, mahadevanImageInThatManipulatingImages2025, kontogiorgosQuestioningRobotUsing2025, vermaTheoryMindAbilities2024, mahadevanGenerativeExpressiveRobot2024, skantzeApplyingGeneralTurntaking2025, mannavaExploringSuitabilityConversational2024, sieversIntroducingNoteLevity2024, sieversInteractingSentimentalRobot2024, cuiNoRightOnline2023a, yeImprovedTrustHumanRobot2023, suChatAdpChatGPTpoweredAdaptation2024, zhangLargeLanguageModels2023b, latifPhysicsAssistantLLMpoweredInteractive2024, yuImprovingPerceivedEmotional2024, bastinGPTAllySafetyorientedSystem2025, yanoUnifiedUnderstandingEnvironment2024, zhangPromptingEmbodiedAI2025, shiradoRealismDrivesInterpersonal2025, elfleetInvestigatingImpactMultimodal2024, perella-holfeldParentEducatorConcerns2024a, elgarfCreativeBotCreativeStoryteller2022, kimUnderstandingLargelanguageModel2024d, wangChallengesAdoptingCompanion2025, padmanabhaVoicePilotHarnessingLlms2024, blancoAIenhancedSocialRobots2024, salemComparativeHumanrobotInteraction2024, liStargazerInteractiveCamera2023, wangCrowdBotOpenenvironmentRobot2024, wangPepperPoseFullbodyPose2024, geGenComUIExploringGenerative2025, hoSETPAiREdDesigningParental2025, zuLanguageSketchingLLMdriven2024, grassiGroundingConversationalRobots2024, tsushimaTaskPlanningFactory2025, shenSocialRobotsSocial2025, loLLMbasedRobotPersonality2025a, grassiStrategiesControllingConversation2025, herathFirstImpressionsHumanoid2025, dellannaSONARAdaptiveControl2024, nardelliIntuitiveInteractionCognitive2025, rosenPreviousExperienceMatters2024, kodurExploringDynamicsHumanRobot2025, starkDobbyConversationalService2024, malnatskyFittingHumorAgeBased2025, jinRobotGPTRobotManipulation2024} \\

Technical Evaluation   & 52 & \cite{spitaleVITAMultiModalLLMBased2025a, taoLAMSLLMDrivenAutomatic2025a, reimannWhatCanYou2025a, kamelabadComparingMonolingualBilingual2025, zhangWalkExperimentControlling2025, itoRobotDynamicallyAsking2025a, karliAlchemistLLMAidedEndUser2024a, ikedaMARCERMultimodalAugmented2025a, zhangBalancingUserControl2025a, antonyXpressSystemDynamic2025a, panACKnowledgeComputationalFramework2025, pereiraMultimodalUserEnjoyment2024a, huDesigningTelepresenceRobots2025, goubardCognitiveModellingVisual, axelssonYouFollow2023, ferriniPerceptsSemanticsMultimodala, mahadevanImageInThatManipulatingImages2025, kontogiorgosQuestioningRobotUsing2025, bellucciImmersiveTailoringEmbodied2025, mahadevanGenerativeExpressiveRobot2024, mannavaExploringSuitabilityConversational2024, sieversInteractingSentimentalRobot2024, cuiNoRightOnline2023a, yeImprovedTrustHumanRobot2023, suChatAdpChatGPTpoweredAdaptation2024, latifPhysicsAssistantLLMpoweredInteractive2024, yuImprovingPerceivedEmotional2024, pintoPredictiveTurntakingLeveraging2024, bastinGPTAllySafetyorientedSystem2025, yanoUnifiedUnderstandingEnvironment2024, shiradoRealismDrivesInterpersonal2025, elgarfCreativeBotCreativeStoryteller2022, kimUnderstandingLargelanguageModel2024d, blancoAIenhancedSocialRobots2024, liStargazerInteractiveCamera2023, wangCrowdBotOpenenvironmentRobot2024, wangPepperPoseFullbodyPose2024, geGenComUIExploringGenerative2025, laiNaturalMultimodalFusionBased2025, zuLanguageSketchingLLMdriven2024, farooqDAIMHRINewHumanRobot2024, grassiGroundingConversationalRobots2024, tsushimaTaskPlanningFactory2025, loLLMbasedRobotPersonality2025a, herathFirstImpressionsHumanoid2025, dellannaSONARAdaptiveControl2024, nardelliIntuitiveInteractionCognitive2025, bassiounyUJIButlerSymbolicNonsymbolic2025, sakamotoEffectivenessConversationalRobots2025, grassiEnhancingLLMBasedHumanRobot2024, jinRobotGPTRobotManipulation2024, wilcockErrRoboticEarn2023a} \\

Wizard-of-Oz    & 8 & \cite{stampfExploringPassengerAutomatedVehicle2024, panACKnowledgeComputationalFramework2025, arjmandEmpathicGroundingExplorations2024, huDesigningTelepresenceRobots2025, hsuResearchCareReflection2025, zhangPromptingEmbodiedAI2025, perella-holfeldParentEducatorConcerns2024a, sakamotoEffectivenessConversationalRobots2025} \\

Case studies   & 11 & \cite{itoRobotDynamicallyAsking2025a, antonyXpressSystemDynamic2025a, panACKnowledgeComputationalFramework2025, mahadevanImageInThatManipulatingImages2025, vermaTheoryMindAbilities2024, suChatAdpChatGPTpoweredAdaptation2024, zhangLargeLanguageModels2023b, latifPhysicsAssistantLLMpoweredInteractive2024, tsushimaTaskPlanningFactory2025, dellannaSONARAdaptiveControl2024, nardelliIntuitiveInteractionCognitive2025} \\

Simulations   & 14 & \cite{karliAlchemistLLMAidedEndUser2024a, panACKnowledgeComputationalFramework2025, mahadevanImageInThatManipulatingImages2025, mahadevanGenerativeExpressiveRobot2024, yeImprovedTrustHumanRobot2023, zhangPromptingEmbodiedAI2025, wangCrowdBotOpenenvironmentRobot2024, wangPepperPoseFullbodyPose2024, zuLanguageSketchingLLMdriven2024, tsushimaTaskPlanningFactory2025, loLLMbasedRobotPersonality2025a, dellannaSONARAdaptiveControl2024, nardelliIntuitiveInteractionCognitive2025, jinRobotGPTRobotManipulation2024} \\

Co-design workshops   & 6 & \cite{hsuBittersweetSnapshotsLife2025a, wangChildRobotRelationalNorm2025a, axelssonOhSorryThink2024, hsuResearchCareReflection2025, hoSETPAiREdDesigningParental2025, malnatskyFittingHumorAgeBased2025} \\

BodyStorming    & 2 & \cite{axelssonOhSorryThink2024, malnatskyFittingHumorAgeBased2025} \\
Think-aloud protocols    & 2 & \cite{liStargazerInteractiveCamera2023, hoSETPAiREdDesigningParental2025} \\

\bottomrule
\end{tabular}
\end{table}

\begin{table}[h!]
\centering
\small
\caption{References of Section 6.2. Evaluation Strategies}
\Description{References for evaluation strategies in LLM-driven HRI (corresponding to Section 7). The table is structured with three columns: Strategies (hierarchical classification of evaluation metrics), Number (count of relevant supporting literature), and Papers (reference numbers of included literature). The evaluation strategies are divided into two major categories: 1) Objective evaluation metrics, including three subcategories—Task Efficiency and Timing (46 papers), Task Accuracy and Performance (42 papers), and LLM-Specific Performance (30 papers); 2) Subjective evaluation metrics, including six subcategories—User’s Perceptual and Relational Experience (65 papers, the most widely cited), Perceived Intelligence (30 papers), Anthropomorphism (19 papers), Usability (31 papers), Safety (24 papers), and Cognitive Load and Workload (13 papers). Each subcategory lists the reference numbers of literature that adopt the corresponding evaluation metrics, systematically collating the literature support for objective and subjective evaluation in LLM-driven HRI research and providing a comprehensive reference for researchers in selecting appropriate evaluation indicators.}
\begin{tabular}{p{0.3\linewidth} p{0.10\linewidth} p{0.55\linewidth}}
    \toprule
    \textbf{Strategies} & \textbf{Number} & \textbf{Papers} \\
    \midrule

\multicolumn{3}{l}{\textit{\textbf{Objective}}} \\[0.3em]
Task Efficiency and Timing      & 46 & \cite{limaPromotingCognitiveHealth2025, taoLAMSLLMDrivenAutomatic2025a, bannaWordsIntegratingPersonality2025, reimannWhatCanYou2025a,aliInclusiveCocreativeChildrobot2025a, zhangWalkExperimentControlling2025, karliAlchemistLLMAidedEndUser2024a, ikedaMARCERMultimodalAugmented2025a, xuExploringUseRobots2025a, leusmannInvestigatingLLMDrivenCuriosity2025, hsuResearchCareReflection2025, axelssonYouFollow2023, mahadevanImageInThatManipulatingImages2025, bellucciImmersiveTailoringEmbodied2025, skantzeApplyingGeneralTurntaking2025, sieversInteractingSentimentalRobot2024, yeImprovedTrustHumanRobot2023, latifPhysicsAssistantLLMpoweredInteractive2024, yuImprovingPerceivedEmotional2024, pintoPredictiveTurntakingLeveraging2024, yanoUnifiedUnderstandingEnvironment2024, zhangPromptingEmbodiedAI2025, shiradoRealismDrivesInterpersonal2025, elfleetInvestigatingImpactMultimodal2024, elgarfCreativeBotCreativeStoryteller2022, kimUnderstandingLargelanguageModel2024d, choLivingAlongsideAreca2025, padmanabhaVoicePilotHarnessingLlms2024, liStargazerInteractiveCamera2023, wangCrowdBotOpenenvironmentRobot2024, wangPepperPoseFullbodyPose2024, geGenComUIExploringGenerative2025, laiNaturalMultimodalFusionBased2025, zuLanguageSketchingLLMdriven2024, grassiGroundingConversationalRobots2024, tsushimaTaskPlanningFactory2025, shenSocialRobotsSocial2025, loLLMbasedRobotPersonality2025a, grassiStrategiesControllingConversation2025, dellannaSONARAdaptiveControl2024, nardelliIntuitiveInteractionCognitive2025, sakamotoEffectivenessConversationalRobots2025, rosenPreviousExperienceMatters2024, kodurExploringDynamicsHumanRobot2025, grassiEnhancingLLMBasedHumanRobot2024, jinRobotGPTRobotManipulation2024} \\

Task Accuracy and Performance   & 42 & \cite{limaPromotingCognitiveHealth2025, loMemoryRobotDesign2025, pinedaSeeYouLater2025a, zhangWalkExperimentControlling2025, itoRobotDynamicallyAsking2025a, karliAlchemistLLMAidedEndUser2024a, wangChildRobotRelationalNorm2025a, ikedaMARCERMultimodalAugmented2025a, panACKnowledgeComputationalFramework2025, pereiraMultimodalUserEnjoyment2024a, ferriniPerceptsSemanticsMultimodala, mahadevanImageInThatManipulatingImages2025, kontogiorgosQuestioningRobotUsing2025, bellucciImmersiveTailoringEmbodied2025, vermaTheoryMindAbilities2024, mannavaExploringSuitabilityConversational2024, sieversInteractingSentimentalRobot2024, cuiNoRightOnline2023a, yeImprovedTrustHumanRobot2023, suChatAdpChatGPTpoweredAdaptation2024, zhangLargeLanguageModels2023b, latifPhysicsAssistantLLMpoweredInteractive2024, bastinGPTAllySafetyorientedSystem2025, yanoUnifiedUnderstandingEnvironment2024, shiradoRealismDrivesInterpersonal2025, liStargazerInteractiveCamera2023, wangCrowdBotOpenenvironmentRobot2024, wangPepperPoseFullbodyPose2024, geGenComUIExploringGenerative2025, hoSETPAiREdDesigningParental2025, farooqDAIMHRINewHumanRobot2024, grassiGroundingConversationalRobots2024, tsushimaTaskPlanningFactory2025, shenSocialRobotsSocial2025, loLLMbasedRobotPersonality2025a, dellannaSONARAdaptiveControl2024, nardelliIntuitiveInteractionCognitive2025, bassiounyUJIButlerSymbolicNonsymbolic2025, sakamotoEffectivenessConversationalRobots2025, kodurExploringDynamicsHumanRobot2025, starkDobbyConversationalService2024, jinRobotGPTRobotManipulation2024} \\

LLM-Specific Performance   & 30 & \cite{loMemoryRobotDesign2025, westerFacingLLMsRobot2024a, karliAlchemistLLMAidedEndUser2024a, ikedaMARCERMultimodalAugmented2025a, pereiraMultimodalUserEnjoyment2024a, goubardCognitiveModellingVisual, bellucciImmersiveTailoringEmbodied2025, vermaTheoryMindAbilities2024, mahadevanGenerativeExpressiveRobot2024, yeImprovedTrustHumanRobot2023, zhangLargeLanguageModels2023b, yuImprovingPerceivedEmotional2024, pintoPredictiveTurntakingLeveraging2024, bastinGPTAllySafetyorientedSystem2025, perella-holfeldParentEducatorConcerns2024a, kimUnderstandingLargelanguageModel2024d, blancoAIenhancedSocialRobots2024, hoSETPAiREdDesigningParental2025, farooqDAIMHRINewHumanRobot2024, grassiGroundingConversationalRobots2024, tsushimaTaskPlanningFactory2025, loLLMbasedRobotPersonality2025a, herathFirstImpressionsHumanoid2025, dellannaSONARAdaptiveControl2024, nardelliIntuitiveInteractionCognitive2025, sakamotoEffectivenessConversationalRobots2025, malnatskyFittingHumorAgeBased2025, grassiEnhancingLLMBasedHumanRobot2024, jinRobotGPTRobotManipulation2024, wilcockErrRoboticEarn2023a} \\[0.6em]

\multicolumn{3}{l}{\textit{\textbf{Subjective}}} \\[0.3em]

User's Perceptual and Relational Experience   & 65 & \cite{limaPromotingCognitiveHealth2025, zhangExploringRobotPersonality2025a, spitaleVITAMultiModalLLMBased2025a, hsuBittersweetSnapshotsLife2025a, taoLAMSLLMDrivenAutomatic2025a, bannaWordsIntegratingPersonality2025, westerFacingLLMsRobot2024a, pinedaSeeYouLater2025a, reimannWhatCanYou2025a, kamelabadComparingMonolingualBilingual2025,aliInclusiveCocreativeChildrobot2025a, itoRobotDynamicallyAsking2025a, wangChildRobotRelationalNorm2025a, ikedaMARCERMultimodalAugmented2025a, xuExploringUseRobots2025a, leusmannInvestigatingLLMDrivenCuriosity2025, zhangBalancingUserControl2025a, antonyXpressSystemDynamic2025a, stampfExploringPassengerAutomatedVehicle2024, panACKnowledgeComputationalFramework2025, arjmandEmpathicGroundingExplorations2024, pereiraMultimodalUserEnjoyment2024a, huDesigningTelepresenceRobots2025, axelssonOhSorryThink2024, hsuResearchCareReflection2025, mahadevanImageInThatManipulatingImages2025, kontogiorgosQuestioningRobotUsing2025, vermaTheoryMindAbilities2024, mahadevanGenerativeExpressiveRobot2024, mannavaExploringSuitabilityConversational2024, sieversIntroducingNoteLevity2024, sieversInteractingSentimentalRobot2024, yeImprovedTrustHumanRobot2023, suChatAdpChatGPTpoweredAdaptation2024, zhangLargeLanguageModels2023b, latifPhysicsAssistantLLMpoweredInteractive2024, yuImprovingPerceivedEmotional2024, bastinGPTAllySafetyorientedSystem2025, yanoUnifiedUnderstandingEnvironment2024, zhangPromptingEmbodiedAI2025, shiradoRealismDrivesInterpersonal2025, elfleetInvestigatingImpactMultimodal2024, perella-holfeldParentEducatorConcerns2024a, elgarfCreativeBotCreativeStoryteller2022, kimUnderstandingLargelanguageModel2024d, choLivingAlongsideAreca2025, wangChallengesAdoptingCompanion2025, blancoAIenhancedSocialRobots2024, salemComparativeHumanrobotInteraction2024, liStargazerInteractiveCamera2023, wangPepperPoseFullbodyPose2024, hoSETPAiREdDesigningParental2025, zuLanguageSketchingLLMdriven2024, farooqDAIMHRINewHumanRobot2024, grassiGroundingConversationalRobots2024, tsushimaTaskPlanningFactory2025, shenSocialRobotsSocial2025, grassiStrategiesControllingConversation2025, herathFirstImpressionsHumanoid2025, dellannaSONARAdaptiveControl2024, nardelliIntuitiveInteractionCognitive2025, sakamotoEffectivenessConversationalRobots2025, rosenPreviousExperienceMatters2024, starkDobbyConversationalService2024, malnatskyFittingHumorAgeBased2025} \\

Perceived Intelligence        & 30 & \cite{spitaleVITAMultiModalLLMBased2025a, westerFacingLLMsRobot2024a, kamelabadComparingMonolingualBilingual2025, itoRobotDynamicallyAsking2025a, leusmannInvestigatingLLMDrivenCuriosity2025, zhangBalancingUserControl2025a, huDesigningTelepresenceRobots2025, hsuResearchCareReflection2025, choARECADesignSpeculation2023, sieversInteractingSentimentalRobot2024, yuImprovingPerceivedEmotional2024, bastinGPTAllySafetyorientedSystem2025, elfleetInvestigatingImpactMultimodal2024, perella-holfeldParentEducatorConcerns2024a, elgarfCreativeBotCreativeStoryteller2022, kimUnderstandingLargelanguageModel2024d, salemComparativeHumanrobotInteraction2024, liStargazerInteractiveCamera2023, geGenComUIExploringGenerative2025, hoSETPAiREdDesigningParental2025, zuLanguageSketchingLLMdriven2024, grassiGroundingConversationalRobots2024, shenSocialRobotsSocial2025, loLLMbasedRobotPersonality2025a, dellannaSONARAdaptiveControl2024, nardelliIntuitiveInteractionCognitive2025, bassiounyUJIButlerSymbolicNonsymbolic2025, sakamotoEffectivenessConversationalRobots2025, rosenPreviousExperienceMatters2024, starkDobbyConversationalService2024} \\

Anthropomorphism          & 19 & \cite{spitaleVITAMultiModalLLMBased2025a, bannaWordsIntegratingPersonality2025, kamelabadComparingMonolingualBilingual2025, aliInclusiveCocreativeChildrobot2025a, antonyXpressSystemDynamic2025a, choARECADesignSpeculation2023, axelssonYouFollow2023, sieversInteractingSentimentalRobot2024, yuImprovingPerceivedEmotional2024, elfleetInvestigatingImpactMultimodal2024, kimUnderstandingLargelanguageModel2024d, salemComparativeHumanrobotInteraction2024, geGenComUIExploringGenerative2025, shenSocialRobotsSocial2025, loLLMbasedRobotPersonality2025a, herathFirstImpressionsHumanoid2025, dellannaSONARAdaptiveControl2024, sakamotoEffectivenessConversationalRobots2025, starkDobbyConversationalService2024} \\

Usability           & 31 & \cite{limaPromotingCognitiveHealth2025, spitaleVITAMultiModalLLMBased2025a, taoLAMSLLMDrivenAutomatic2025a, reimannWhatCanYou2025a, zhangWalkExperimentControlling2025, karliAlchemistLLMAidedEndUser2024a, wangChildRobotRelationalNorm2025a, ikedaMARCERMultimodalAugmented2025a, xuExploringUseRobots2025a, leusmannInvestigatingLLMDrivenCuriosity2025, zhangBalancingUserControl2025a, stampfExploringPassengerAutomatedVehicle2024, panACKnowledgeComputationalFramework2025, huDesigningTelepresenceRobots2025, choARECADesignSpeculation2023, mahadevanImageInThatManipulatingImages2025, sieversIntroducingNoteLevity2024, cuiNoRightOnline2023a, perella-holfeldParentEducatorConcerns2024a, padmanabhaVoicePilotHarnessingLlms2024, liStargazerInteractiveCamera2023, wangCrowdBotOpenenvironmentRobot2024, geGenComUIExploringGenerative2025, hoSETPAiREdDesigningParental2025, laiNaturalMultimodalFusionBased2025, tsushimaTaskPlanningFactory2025, dellannaSONARAdaptiveControl2024, nardelliIntuitiveInteractionCognitive2025, bassiounyUJIButlerSymbolicNonsymbolic2025, kodurExploringDynamicsHumanRobot2025, starkDobbyConversationalService2024} \\

Safety          & 24 & \cite{spitaleVITAMultiModalLLMBased2025a, hsuBittersweetSnapshotsLife2025a, loMemoryRobotDesign2025, kamelabadComparingMonolingualBilingual2025, stampfExploringPassengerAutomatedVehicle2024, mannavaExploringSuitabilityConversational2024, suChatAdpChatGPTpoweredAdaptation2024, zhangLargeLanguageModels2023b, bastinGPTAllySafetyorientedSystem2025, shiradoRealismDrivesInterpersonal2025, elfleetInvestigatingImpactMultimodal2024, perella-holfeldParentEducatorConcerns2024a, kimUnderstandingLargelanguageModel2024d, wangChallengesAdoptingCompanion2025, salemComparativeHumanrobotInteraction2024, geGenComUIExploringGenerative2025, hoSETPAiREdDesigningParental2025, laiNaturalMultimodalFusionBased2025, farooqDAIMHRINewHumanRobot2024, grassiGroundingConversationalRobots2024, tsushimaTaskPlanningFactory2025, shenSocialRobotsSocial2025, bassiounyUJIButlerSymbolicNonsymbolic2025, kodurExploringDynamicsHumanRobot2025} \\

Cognitive Load and Workload         & 13 & \cite{spitaleVITAMultiModalLLMBased2025a, taoLAMSLLMDrivenAutomatic2025a, zhangWalkExperimentControlling2025, wangChildRobotRelationalNorm2025a, leusmannInvestigatingLLMDrivenCuriosity2025, mahadevanImageInThatManipulatingImages2025, yeImprovedTrustHumanRobot2023, liStargazerInteractiveCamera2023, geGenComUIExploringGenerative2025, hoSETPAiREdDesigningParental2025, grassiStrategiesControllingConversation2025, dellannaSONARAdaptiveControl2024, rosenPreviousExperienceMatters2024} \\

\bottomrule
\end{tabular}
\end{table}


\begin{table}[ht!]
\centering
\small
\caption{References of Section 7. Applications}
\Description{References for application domains in LLM-driven HRI (corresponding to Section 7). The table is structured with three columns: Applications (classification of core application domains of LLM-driven HRI), Number (count of relevant supporting literature for each domain), and Papers (reference numbers of included literature). The application domains are divided into eight categories with specific details as follows: 1) Social and Conversational Systems (18 papers, referenced as [5, 32, 37, 48, 92, 93, 103, 112, 115, 126, 127, 133, 140–142, 157, 180, 184]); 2) Healthcare and Wellbeing (12 papers, referenced as [11, 17, 56, 57, 76, 87, 91, 107, 143, 148, 168, 182]); 3) Domestic and Everyday Use (17 papers, referenced as [6, 14, 24, 25, 46, 59, 65, 73, 81, 94, 109, 124, 159, 162, 175, 183, 197]); 4) Teaching and Education (13 papers, referenced as [4, 9, 35, 49, 54, 63, 67, 77, 83, 95, 96, 113, 158]); 5) Industrial Manufacturing (7 papers, referenced as [15, 36, 68, 75, 114, 155, 178]); 6) AR/VR-enabled Interactions (6 papers, referenced as [16, 34, 136, 144, 187, 188]); 7) Public Space Service (9 papers, referenced as [42, 47, 53, 58, 121, 145, 164, 170, 176]); 8) Other (4 papers, referenced as [12, 27, 69, 153]). The table systematically collates literature supporting each application domain, reflecting the practical deployment scope of LLM-driven HRI and providing a comprehensive reference for researchers in selecting application-oriented research directions.}
\begin{tabular}{p{0.30\linewidth} p{0.10\linewidth} p{0.55\linewidth}}
    \toprule
    \textbf{Applications} & \textbf{Number} & \textbf{Papers} \\
    \midrule

Social and Conversational Systems     & 18 & \cite{antonyXpressSystemDynamic2025a, pereiraMultimodalUserEnjoyment2024a, ferriniPerceptsSemanticsMultimodala, vermaTheoryMindAbilities2024, mahadevanGenerativeExpressiveRobot2024, skantzeApplyingGeneralTurntaking2025, sieversIntroducingNoteLevity2024, sieversInteractingSentimentalRobot2024, zhangLargeLanguageModels2023b, yuImprovingPerceivedEmotional2024, pintoPredictiveTurntakingLeveraging2024, salemComparativeHumanrobotInteraction2024, shenSocialRobotsSocial2025, loLLMbasedRobotPersonality2025a, dellannaSONARAdaptiveControl2024, nardelliIntuitiveInteractionCognitive2025, sakamotoEffectivenessConversationalRobots2025, grassiEnhancingLLMBasedHumanRobot2024} \\

Healthcare and Wellbeing  & 12 & \cite{limaPromotingCognitiveHealth2025, zhangExploringRobotPersonality2025a, spitaleVITAMultiModalLLMBased2025a, hsuBittersweetSnapshotsLife2025a, loMemoryRobotDesign2025, westerFacingLLMsRobot2024a, axelssonOhSorryThink2024, hsuResearchCareReflection2025, suChatAdpChatGPTpoweredAdaptation2024, padmanabhaVoicePilotHarnessingLlms2024, blancoAIenhancedSocialRobots2024, laiNaturalMultimodalFusionBased2025} \\

Domestic and Everyday Use       & 17 & \cite{ikedaMARCERMultimodalAugmented2025a, xuExploringUseRobots2025a, leusmannInvestigatingLLMDrivenCuriosity2025, zhangBalancingUserControl2025a, panACKnowledgeComputationalFramework2025, arjmandEmpathicGroundingExplorations2024, goubardCognitiveModellingVisual, choARECADesignSpeculation2023, mahadevanImageInThatManipulatingImages2025, choLivingAlongsideAreca2025, wangChallengesAdoptingCompanion2025, wangPepperPoseFullbodyPose2024, zuLanguageSketchingLLMdriven2024, bassiounyUJIButlerSymbolicNonsymbolic2025, rosenPreviousExperienceMatters2024, kodurExploringDynamicsHumanRobot2025, jinRobotGPTRobotManipulation2024} \\

Teaching and Education         & 13 & \cite{kamelabadComparingMonolingualBilingual2025, aliInclusiveCocreativeChildrobot2025a, itoRobotDynamicallyAsking2025a, wangChildRobotRelationalNorm2025a, axelssonYouFollow2023, mannavaExploringSuitabilityConversational2024, latifPhysicsAssistantLLMpoweredInteractive2024, perella-holfeldParentEducatorConcerns2024a, elgarfCreativeBotCreativeStoryteller2022, liStargazerInteractiveCamera2023, hoSETPAiREdDesigningParental2025, grassiStrategiesControllingConversation2025, malnatskyFittingHumorAgeBased2025} \\

Industrial Manufacturing         & 7 & \cite{pinedaSeeYouLater2025a, karliAlchemistLLMAidedEndUser2024a, kontogiorgosQuestioningRobotUsing2025, yeImprovedTrustHumanRobot2023, bastinGPTAllySafetyorientedSystem2025, farooqDAIMHRINewHumanRobot2024, tsushimaTaskPlanningFactory2025} \\

AR/VR-enabled Interactions         & 6 & \cite{zhangWalkExperimentControlling2025, stampfExploringPassengerAutomatedVehicle2024, bellucciImmersiveTailoringEmbodied2025, zhangPromptingEmbodiedAI2025, shiradoRealismDrivesInterpersonal2025, elfleetInvestigatingImpactMultimodal2024} \\

Public Space Service       & 9 & \cite{reimannWhatCanYou2025a, huDesigningTelepresenceRobots2025, yanoUnifiedUnderstandingEnvironment2024, wangCrowdBotOpenenvironmentRobot2024, geGenComUIExploringGenerative2025, grassiGroundingConversationalRobots2024, herathFirstImpressionsHumanoid2025, starkDobbyConversationalService2024, wilcockErrRoboticEarn2023a} \\

Other      & 4 & \cite{taoLAMSLLMDrivenAutomatic2025a, kimUnderstandingLargelanguageModel2024d, cuiNoRightOnline2023a, bannaWordsIntegratingPersonality2025} \\

\bottomrule
\end{tabular}
\end{table}


\end{document}